# A Convergence Theory for Deep Learning via Over-Parameterization

Zeyuan Allen-Zhu
zeyuan@csail.mit.edu
Microsoft Research AI

Yuanzhi Li
yuanzhil@stanford.edu
Stanford University
Princeton University

Zhao Song
zhaos@utexas.edu
UT-Austin
University of Washington
Harvard University

**Abstract**

Deep neural networks (DNNs) have demonstrated dominating performance in many fields; since AlexNet, networks used in practice are going wider and deeper. On the theoretical side, a long line of works has been focusing on training neural networks with one hidden layer. The theory of multi-layer networks remains largely unsettled.

In this work, we prove why stochastic gradient descent (SGD) can find *global minima* on the training objective of DNNs in *polynomial time*. We only make two assumptions: the inputs are non-degenerate and the network is over-parameterized. The latter means the network width is sufficiently large: *polynomial* in $L$, the number of layers and in $n$, the number of samples.

Our key technique is to derive that, in a sufficiently large neighborhood of the random initialization, the optimization landscape is almost-convex and semi-smooth even with ReLU activations. This implies an equivalence between over-parameterized neural networks and neural tangent kernel (NTK) in the finite (and polynomial) width setting.

As concrete examples, starting from randomly initialized weights, we prove that SGD can attain 100% training accuracy in classification tasks, or minimize regression loss in linear convergence speed, with running time polynomial in $n, L$. Our theory applies to the widely-used but non-smooth ReLU activation, and to any smooth and possibly non-convex loss functions. In terms of network architectures, our theory at least applies to fully-connected neural networks, convolutional neural networks (CNN), and residual neural networks (ResNet).

*This paper was presented at ICML 2019, and is a simplified version of our prior work for recurrent neural networks (RNNs) [6]. V1 appears on arXiv on this date and no new result is added since then. V2 adds citations, V3/V4/V5 polish writing. This work was done when Yuanzhi Li and Zhao Song were 2018 summer interns at Microsoft Research Redmond. When this work was performed, Yuanzhi Li was also affiliated with Princeton University, and Zhao Song was also affiliated with UW and Harvard. We would like to specially thank Greg Yang for many enlightening discussions, thank Ofer Dekel, Sebastien Bubeck, and Harry Shum for very helpful conversations, and thank Jincheng Mei for carefully checking the proofs of this paper.

# 1 Introduction

Neural networks have demonstrated a great success in numerous machine-learning tasks [7, 26, 31, 35, 38, 48, 49]. One of the empirical findings is that neural networks, trained by first-order methods from random initialization, have a remarkable ability to fit training data [60].

From an *expressibility* perspective, this may not be surprising since modern neural networks are often over-parameterized: they have much more parameters than the number of training samples. There certainly *exist* parameter choices with zero training error as long as data is non-degenerate.

**Yet**, from an optimization perspective, the fact that randomly-initialized first-order methods can find global minima on the training data is *quite non-trivial*: neural networks are often equipped with the ReLU activation, making the training objective not only non-convex, but even non-smooth. Even the general convergence for finding approximate critical points of a non-convex, non-smooth function is not fully-understood [13] and appears to be a challenging question on its own. This is in direct contrast to practice, in which ReLU networks trained by stochastic gradient descent (SGD) from random initialization *almost never* suffer from non-smoothness or non-convexity, and can avoid local minima for a variety of network architectures (see Goodfellow et al. [25]). A theoretical justification was missing to explain this phenomenon.

There are quite a few papers trying to understand the success of neural networks from optimization perspective. Many of them focus on the case when the inputs are random Gaussian, and work only for two-layer neural networks [12, 20, 22, 37, 45, 52, 57, 62, 63]. Li and Liang [36] show that for a two-layer network with ReLU activation, SGD finds nearly-global optimal (say, 99% classification accuracy) solutions on the training data, as long as the network is *over-parameterized*, meaning that the number of neurons is polynomially large comparing to the input size. Moreover, if the data is sufficiently structured (say, coming from mixtures of separable distributions), this accuracy extends also to *test data*. As a separate note, over-parameterization is suggested as the possible key to avoid bad local minima by Safran and Shamir [46] even for two-layer networks.

There are also results that go beyond two-layer networks with limitations. Some consider deep *linear* neural networks without any activation functions [8, 10, 27, 33]. Daniely [15] studies multi-layer neural networks but essentially only with respect to the *convex* task of training the last layer.[1] Soudry and Carmon [54] show that under over-parameterization and under random input perturbation, there is bad local minima for multi-layer neural networks. Jacot et al. [32] derive global convergence using neural tangent kernel for *infinite-width* neural networks.

In this paper, we study the following fundamental questions

*Can DNN be trained close to zero training error efficiently under mild assumptions?*

*If so, can the running time depend only* polynomially *in the network depth and input size?*

**Motivation.** In 2012, AlexNet was born with 5 convolutional layers [35]. The later VGG network uses 19 layers [51], and GoogleNet uses 22 layers [56]. In practice, we cannot go deeper by naively stacking layers together, due to the so-called vanishing/exploding gradient problem. To deal with this issue, networks with residual links (ResNet) were proposed with the capability of handling at least 152 layers [31]. Compared with practical networks that go much deeper, existing theory has been mostly around two-layer (thus one-hidden-layer) neural networks, even just for the training process alone. Thus,

*Can we theoretically justify how the training process has worked for multi-layer neural networks?*

In this paper, we extend the over-parameterization theory to *multi-layer* neural networks.

[1]Daniely [15] works in a parameter regime where the weight changes of all layers except the last one make negligible contribution to the final output.

## 1.1 Our Result

We show that over-parameterized neural networks can be trained by vanilla first-order methods such as gradient descent (GD) or stochastic gradient descent (SGD) to global minima (e.g. *zero* training error), as long as the data is non-degenerate.

We say that the data is non-degenerate if every pairs of samples are distinct. This is a minimal requirement since a dataset with two identical data points of different labels cannot be trained to zero error. We denote by $\delta$ the minimum (relative) distance between two data points, and by $n$ the number of training samples. Now, consider an $L$-layer fully-connected feedforward neural network, each hidden layer consisting of $m$ neurons equipped with ReLU activation. We show that,

- As long as $m \geq \mathsf{poly}(n, L, \delta^{-1})$, starting from random Gaussian initialization, GD/SGD finds an $\varepsilon$-error global minimum in $\ell_2$ regression using at most $T = \mathsf{poly}(n, L, \delta^{-1}) \log \frac{1}{\varepsilon}$ iterations.
- If the task is multi-label classification, then GD/SGD finds an 100% accuracy classifier on the training set in $T = \mathsf{poly}(n, L, \delta^{-1})$ iterations.
- Our result also applies to other Lipschitz-smooth loss functions, and some other network architectures including convolutional neural networks (CNNs) and residual networks (ResNet).

In contrast, prior work on this task either requires $m$ and $T$ to grow in $e^{O(L)}$ (and essentially only the last layer is trained) [15]; or requires $m = \infty$ [32].

**Our Contributions.** We summarize our technical contributions below.

- For a sufficiently large neighborhood of the random initialization, we prove that the training landscape is almost convex and semi-smooth. This somewhat explains the empirical finding by Goodfellow et al. [25] that GD/SGD will not be trapped in local minima. (See Section 4.1.)
- For a sufficiently large neighborhood of the random initialization, we derive an equivalence between neural networks and the neural tangent kernel (NTK) introduced by Jacot et al. [32]. Unlike the prior work in which they show the equivalence only for infinite-width networks (i.e., $m = \infty$), here we only need $m = \mathsf{poly}(L)$ for such an equivalence to hold. (See Section 4.2.)
- We show that equipped with ReLU activation, neural networks do not suffer from exponential gradient explosion or vanishing. This is the key reason we can avoid exponential dependency on $L$. If one is okay with $e^{O(L)}$ dependency, many proofs shall become trivial. (See Section 5.)
- We derive a stability theory of neural networks against small but adversarial perturbations that may be of independent interests. Previous results on this topic either have exponential blowup in $L$ [15] or requires the width to go to infinity [32]. (See Section 5.)
- We derive our results by training only hidden layers. This can be more meaningful than training all the layers together, in which if one is not careful with parameter choices, the training process can degenerate as if only the last layer is trained [15]. That is a convex task and may not reflect the true power of deep learning. (Of course, as a simple corollary, our results also apply to training all the layers together.)

Finally, we emphasize that this present paper as a deeply-simplified version of the recurrent neural network (RNN) paper [6] by the same set of authors. To some extent, DNN is a "special case" of RNN,[2] thus most of the technical tools were already developed in [6]. We write this DNN result as a separate paper because: (1) not all the readers can easily derive the DNN result from [6]; (2) the convergence of DNN can be important on its own; (3) the proof in this paper is much simpler (30 vs 80 pages) and could reach out to a wider audience; (4) the simplicity of this paper allows us to tighten parameters in some non-trivial ways; and (5) the simplicity of this paper allows us to also study convolutional networks, residual networks, as well as different loss functions (all of them were missing from [6]). We also note that the techniques of this paper can be combined with [6] to show the global convergence of training over-parameterized *deep* RNN. We ignore the details so as not to complicate this paper.

[2]A recurrent neural network executed on input sequences with time horizon $L$ is very similar to a feedforward neural network with $L$ layers. The main difference is that in a feedforward network, weight matrices are different across layers, and thus independently randomly initialized; in contrast, in an RNN, the same weight matrix is applied across the entire time horizon, so we do not have fresh new randomness for proofs that involve in induction. In other words, the over-parameterized convergence theory of DNN is *much simpler* than that of RNN.

**Towards Generalization.** In practice, deeper and wider neural networks generalize better [55, 59], so what can we say in theory? Although this paper does not explicitly cover generalization to test data, since a neural network in our parameter regime simulates its neural tangent kernel (NTK), it is clear that neural networks provide generalization at least *as good as its NTK*.

In the PAC-learning language, one may study generalization with respect to *concept classes*. Follow-up work [5] shows that three-layer over-parameterized ReLU networks can efficiently (in polynomial time and sample complexity) learn the concept class of three-layer neural networks with smooth activations [5], and the follow-up work [4] shows stronger results for three-layer ResNet.

It is worth pointing out that the three-layer result [5] goes beyond the almost-convex regime and thus is not captured by its NTK; more interestingly, the three-layer ResNet result [4] is not achievable (in a provable sense) by any kernel method including any NTK.

**A concurrent but different result.** We acknowledge a concurrent work [19] that has a similar abstract but is different from us in many aspects. Since we noticed many readers cannot tell the two results apart, we compare them carefully below. Du et al. [19] has two main results:

- they show time complexity $\mathsf{poly}(n, 2^{O(L)}, 1/\lambda_{\min})$ for fully-connected networks; and
- they show time complexity $\mathsf{poly}(n, L, 1/\lambda_{\min})$ for ResNets.

Here, the *data-dependent* parameter $\lambda_{\min}$ is the minimal eigenvalue of a complicated, $L$-times recursively-defined $n \times n$ kernel matrix. They only proved $\lambda_{\min} > 0$ from $\delta > 0$. It is not clear whether $\frac{1}{\lambda_{\min}}$ is small or even polynomial from their writing. What is clear is that $\lambda_{\min}$ depends both on $n$ and $L$ (and even on $2^{O(L)}$ for fully-connected networks).

Interestingly, using an argument that $\lambda_{\min}$ only depends on $\mathsf{poly}(L)$ for ResNet, they argued that ResNet has "exponential improvement over fully-connected networks." According to our paper, such improvement does not hold for the ReLU activation since both complexities (for fully-connected and residual networks) can be polynomially bounded by $L$.

Their result is also different from us in many other aspects. Their result only applies to the constant-smooth activation functions (with a final bound that *exponentially* depends on this constant) and thus *cannot* apply to the state-of-the-art ReLU activation.[3] Their result only applies to GD but not to SGD. Their result only applies to $\ell_2$ loss but not others.

Finally, we acknowledge that the polynomials in our paper is quite large and might not be directly applicable to practical regime. However, most of the polynomial factors come from a *worst-case* analysis to handle the *non-smoothness* of ReLU. If instead smooth activations are considered, our bounds can be significantly improved.

## 1.2 Other Related Works

Li and Liang [36] originally prove their result for the cross-entropy loss, together with some test accuracy guarantee. (If data is "well-structured", they prove two-layer over-parameterized neural networks can learn it using SGD with polynomially many samples [36].) Later, the "training accuracy" (not the testing accuracy) part of [36] was extended to the $\ell_2$ loss [21]. The result of [21] claims to have adopted a learning rate $m$ times larger than [36], but that is unfair because they have re-scaled the network by a factor of $\sqrt{m}$.[4]

[3]For instance, we have to establish a semi-smoothness theorem for deep ReLU networks (see Theorem 4). If instead the activation function is Lipscthiz smooth, and if one does not care about exponential blow up in the number of layers $L$, then the network is automatically $2^{O(L)}$-Lipschitz smooth.

Linear networks without activation functions are important subjects on its own. Besides the already cited references [8, 10, 27, 33], there are a number of works that study *linear dynamical systems*, which can be viewed as the linear version of recurrent neural networks or reinforcement learning. Recent works in this line of research include [1, 9, 17, 18, 28–30, 41, 44, 50].

There is sequence of work about one-hidden-layer (multiple neurons) CNN [12, 20, 24, 43, 62]. Whether the patches overlap or not plays a crucial role in analyzing algorithms for such CNN. One category of the results have required the patches to be disjoint [12, 20, 62]. The other category [24, 43] have figured out a weaker assumption or even removed that patch-disjoint assumption. On input data distribution, most relied on inputs being Gaussian [12, 20, 43, 62], and some assumed inputs to be symmetrically distributed with identity covariance and boundedness [24].

As for ResNet, Li and Yuan [37] proved that SGD learns one-hidden-layer residual neural networks under Gaussian input assumption. The techniques in [62, 63] can also be generalized to one-hidden-layer ResNet under the Gaussian input assumption; they can show that GD starting from good initialization point (via tensor initialization) learns ResNet. Hardt and Ma [27] deep *linear* residual networks have no spurious local optima.

If no assumption is allowed, neural networks have been shown hard in several different perspectives. Thirty years ago, Blum and Rivest [11] first proved that learning the neural network is NP-complete. Stronger hardness results have been proved over the last decade [14, 16, 23, 34, 39, 40, 53].

# 2 Preliminaries

We use $\mathcal{N}(\mu, \sigma)$ to denote the Gaussian distribution of mean $\mu$ and variance $\sigma$; and $\mathcal{B}(m, \frac{1}{2})$ to denote the binomial distribution with $m$ trials and 1/2 success rate. We use $\|v\|_2$ or $\|v\|$ to denote Euclidean norms of vectors $v$, and $\|\mathbf{M}\|_2, \|\mathbf{M}\|_F$ to denote spectral and Frobenius norms of matrices $\mathbf{M}$. For a tuple $\overrightarrow{\mathbf{W}} = (\mathbf{W}_1, \dots, \mathbf{W}_L)$ of matrices, we let $\|\overrightarrow{\mathbf{W}}\|_2 = \max_{\ell \in [L]} \|\mathbf{W}_\ell\|_2$ and $\|\overrightarrow{\mathbf{W}}\|_F = (\sum_{\ell=1}^L \|\mathbf{W}_\ell\|_F^2)^{1/2}$.

We use $\phi(x) = \max\{0, x\}$ to denote the ReLU function, and extend it to vectors $v \in \mathbb{R}^m$ by letting $\phi(v) = (\phi(v_1), \dots, \phi(v_m))$. We use $\mathbb{1}_{event}$ to denote the indicator function for *event*.

The training data consist of vector pairs $\{(x_i, y_i^*)\}_{i \in [n]}$, where each $x_i \in \mathbb{R}^{\mathfrak{d}}$ is the feature vector and $y_i^*$ is the label of the $i$-th training sample. We assume *without loss of generality* that data are normalized so that $\|x_i\| = 1$ and its last coordinate $(x_i)_{\mathfrak{d}} = \frac{1}{\sqrt{2}}$.[5] We also assume $\|y_i^*\| \le O(1)$ for notation simplicity.[6]

We make the following separable assumption on the training data (motivated by [36]):

**Assumption 2.1.** *For every pair $i, j \in [n]$, we have $\|x_i - x_j\| \ge \delta$.*

---

[4] If one replaces any function $f(x)$ with $f\big(\frac{x}{\sqrt{m}}\big)$ then the gradient decreases by a factor of $\sqrt{m}$ and the needed movement in $x$ increases by a factor of $\sqrt{m}$. Thus, you can equivalently increase the learning rate by a factor of $m$.

[5] Without loss of generality, one can re-scale and assume $\|x_i\| \le 1/\sqrt{2}$ for every $i \in [n]$. Again, without loss of generality, one can pad each $x_i$ by an additional coordinate to ensure $\|x_i\| = 1/\sqrt{2}$. Finally, without loss of generality, one can pad each $x_i$ by an additional coordinate $\frac{1}{\sqrt{2}}$ to ensure $\|x_i\| = 1$. This last coordinate $\frac{1}{\sqrt{2}}$ is equivalent to introducing a (random) bias term, because $\mathbf{A}(\frac{y}{\sqrt{2}}, \frac{1}{\sqrt{2}}) = \frac{\mathbf{A}}{\sqrt{2}}(y, 0) + b$ where $b \sim \mathcal{N}(0, \frac{1}{m}\mathbf{I})$. In our proofs, the specific constant $\frac{1}{\sqrt{2}}$ does not matter.

[6] If $\|y_i^*\| \le \Omega$ for some parameter $\Omega$, our complexities shall also grow in $\mathsf{poly}(\Omega)$.

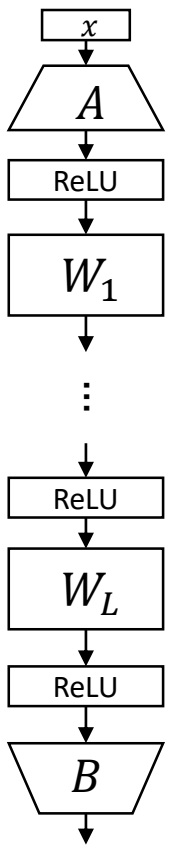

To present the simplest possible proof, the main body of this paper only focuses on depth-$L$ feedforward fully-connected neural networks with an $\ell_2$-regression task. Therefore, each $y_i^* \in \mathbb{R}^d$ is a target vector for the regression task. We explain how to extend it to more general settings in Section 6 and the Appendix. For notational simplicity, we assume all the hidden layers have the same number of neurons, and our results trivially generalize to each layer having different number of neurons. Specifically, we focus on the following network

$$\begin{aligned} g_{i,0} &= \mathbf{A}x_i & h_{i,0} &= \phi(\mathbf{A}x_i) & &\text{for } i \in [n] \\ g_{i,\ell} &= \mathbf{W}_\ell h_{i,\ell-1} & h_{i,\ell} &= \phi(\mathbf{W}_\ell h_{i,\ell-1}) & &\text{for } i \in [n], \ell \in [L] \\ y_i &= \mathbf{B}h_{i,L} & & & &\text{for } i \in [n] \end{aligned}$$

where $\mathbf{A} \in \mathbb{R}^{m\times\mathfrak{d}}$ is the weight matrix for the input layer, $\mathbf{W}_\ell \in \mathbb{R}^{m\times m}$ is the weight matrix for the $\ell$-th hidden layer, and $\mathbf{B} \in \mathbb{R}^{d\times m}$ is the weight matrix for the output layer. For notational convenience in the proofs, we may also use $h_{i,-1}$ to denote $x_i$ and $\mathbf{W}_0$ to denote $\mathbf{A}$.

**Definition 2.2** (diagonal sign matrix)**.** *For each $i \in [n]$ and $\ell \in \{0, 1, \dots, L\}$, we denote by $\mathbf{D}_{i,\ell}$ the diagonal sign matrix where $(\mathbf{D}_{i,\ell})_{k,k} = \mathbb{1}_{(\mathbf{W}_\ell h_{i,\ell-1})_k \geq 0}$ for each $k \in [m]$.*

As a result, we have $h_{i,\ell} = \mathbf{D}_{i,\ell}\mathbf{W}_\ell h_{i,\ell-1} = \mathbf{D}_{i,\ell} g_{i,\ell}$ and $(\mathbf{D}_{i,\ell})_{k,k} = \mathbb{1}_{(g_{i,\ell})_k \geq 0}$. We make the following standard choices of random initialization:

**Definition 2.3.** *We say that $\overrightarrow{\mathbf{W}} = (\mathbf{W}_1, \dots, \mathbf{W}_L)$, $\mathbf{A}$ and $\mathbf{B}$ are at random initialization if*

- $[\mathbf{W}_\ell]_{i,j} \sim \mathcal{N}(0, \frac{2}{m})$ *for every $i, j \in [m]$ and $\ell \in [L]$;*
- $\mathbf{A}_{i,j} \sim \mathcal{N}(0, \frac{2}{m})$ *for every $(i,j) \in [m] \times [\mathfrak{d}]$; and*
- $\mathbf{B}_{i,j} \sim \mathcal{N}(0, \frac{1}{d})$ *for every $(i,j) \in [d] \times [m]$.*

**Assumption 2.4.** *Throughout this paper we assume $m \geq \Omega\big(\mathsf{poly}(n, L, \delta^{-1}) \cdot d\big)$ for some sufficiently large polynomial. To present the simplest proof, we did not try to improve such polynomial factors. We will also assume $\delta \leq O(\frac{1}{L})$ for notation simplicity.*

## 2.1 Objective and Gradient

Our regression objective is

$$F(\overrightarrow{\mathbf{W}}) \stackrel{\text{def}}{=} \sum_{i=1}^{n} F_i(\overrightarrow{\mathbf{W}}) \quad \text{where} \quad F_i(\overrightarrow{\mathbf{W}}) \stackrel{\text{def}}{=} \frac{1}{2}\|\mathbf{B}h_{i,L} - y_i^*\|^2 \quad \text{for each } i \in [n]$$

We also denote by $\mathsf{loss}_i \stackrel{\text{def}}{=} \mathbf{B}h_{i,L} - y_i^*$ the *loss vector* for sample $i$. For simplicity, we focus on training only hidden weights $\overrightarrow{\mathbf{W}}$ in this paper and leave $\mathbf{A}$ and $\mathbf{B}$ at random initialization. Our result naturally extends to the case when $\mathbf{A}$, $\mathbf{B}$ and $\overrightarrow{\mathbf{W}}$ are jointly trained.[7]

**Definition 2.5.** *For each $\ell \in \{1, 2, \cdots, L\}$, we define $\mathsf{Back}_{i,\ell} \stackrel{\text{def}}{=} \mathbf{B}\mathbf{D}_{i,L}\mathbf{W}_L \cdots \mathbf{D}_{i,\ell}\mathbf{W}_\ell \in \mathbb{R}^{d\times m}$ and for $\ell = L+1$, we define $\mathsf{Back}_{i,\ell} = \mathbf{B} \in \mathbb{R}^{d\times m}$.*

Using this notation, one can calculate the gradient of $F(\overrightarrow{\mathbf{W}})$ as follows.

[7] We note that if one jointly trains all the layers, in certain parameter regimes, it may be equivalent to as if only the last layer is trained [15]. We therefore choose to fix the last layer $\mathbf{B}$ to avoid such confusion.

**Fact 2.6.** *The gradient with respect to the k-th row of $\mathbf{W}_\ell \in \mathbb{R}^{m\times m}$ is*

$$\nabla_{[\mathbf{W}_\ell]_k} F(\overrightarrow{\mathbf{W}}) = \sum_{i=1}^n (\mathsf{Back}_{i,\ell+1}^\top \mathsf{loss}_i)_k \cdot h_{i,\ell-1} \cdot \mathbb{1}_{\langle [\mathbf{W}_\ell]_k, h_{i,\ell-1}\rangle \geq 0}$$

*The gradient with respect to $\mathbf{W}_\ell$ is*

$$\nabla_{\mathbf{W}_\ell} F(\overrightarrow{\mathbf{W}}) = \sum_{i=1}^n \mathbf{D}_{i,\ell} (\mathsf{Back}_{i,\ell+1}^\top \mathsf{loss}_i) h_{i,\ell-1}^\top$$

*We denote by $\nabla F(\overrightarrow{\mathbf{W}}) = \big(\nabla_{\mathbf{W}_1} F(\overrightarrow{\mathbf{W}}), \dots, \nabla_{\mathbf{W}_L} F(\overrightarrow{\mathbf{W}})\big)$.*

# 3 Our Results and Techniques

To present our result in the simplest possible way, we choose to mainly focus on fully-connected $L$-layer neural networks with the $\ell_2$ regression loss. We shall extend it to more general settings (such as convolutional and residual networks and other losses) in Section 6. Our main results can be stated as follows:

**Theorem 1** (gradient descent). *Suppose $m \geq \widetilde{\Omega}\big(\mathsf{poly}(n, L, \delta^{-1}) \cdot d\big)$. Starting from random initialization, with probability at least $1 - e^{-\Omega(\log^2 m)}$, gradient descent with learning rate $\eta = \Theta\big(\frac{d\delta}{\mathsf{poly}(n,L)\cdot m}\big)$ finds a point $F(\overrightarrow{\mathbf{W}}) \leq \varepsilon$ in $T = \Theta\big(\frac{\mathsf{poly}(n,L)}{\delta^2} \cdot \log \varepsilon^{-1}\big)$ iterations.*

This is known as the linear convergence rate because $\varepsilon$ drops exponentially fast in $T$. We have not tried to improve the polynomial factors in $m$ and $T$, and are aware of several ways to improve these factors (but at the expense of complicating the proof). We note that $\mathfrak{d}$ is the data input dimension and our result is independent of $\mathfrak{d}$.

*Remark.* In our version 1, for simplicity, we also put a $\log^2(1/\varepsilon)$ factor in the amount of overparameterization $m$ in Theorem 1. Since some readers have raised concerns regarding this [19], we have removed it at the expense of changing half a line of the proof.

**Theorem 2** (SGD). *Suppose $b \in [n]$ and $m \geq \widetilde{\Omega}\big(\frac{\mathsf{poly}(n,L,\delta^{-1})\cdot d}{b}\big)$. Starting from random initialization, with probability at least $1 - e^{-\Omega(\log^2 m)}$, SGD with learning rate $\eta = \Theta(\frac{b\delta d}{\mathsf{poly}(n,L) m \log^2 m})$ and mini-batch size $b$ finds $F(\overrightarrow{\mathbf{W}}) \leq \varepsilon$ in $T = \Theta\big(\frac{\mathsf{poly}(n,L)\cdot \log^2 m}{\delta^2 b} \cdot \log \varepsilon^{-1}\big)$ iterations.*

This is again a linear convergence rate because $T \propto \log \frac{1}{\varepsilon}$. The reason for the additional $\log^2 m$ factor comparing to Theorem 1 is because we have a $1 - e^{-\Omega(\log^2 m)}$ high confidence bound.

*Remark.* For experts in optimization theory, one may immediately question the accuracy of Theorem 2, because SGD is known to converge at a slower rate $T \propto \frac{1}{\mathsf{poly}(\varepsilon)}$ even for convex functions. There is no contradiction here. Imaging a strongly convex function $f(x) = \sum_{i=1}^n f_i(x)$ that has a common minimizer $x^* \in \arg\min_x \{f_i(x)\}$ for every $i \in [n]$, then SGD is known to converge in a linear convergence rate.

# 4 Conceptual Messages and Technical Theorems

We highlight two conceptual messages that arise from the proofs of Theorem 1 and 2.

## 4.1 Objective is Almost Convex and Semi-Smooth

The first message is about the optimization landscape for points that are sufficiently close to the random initialization. It consists of two theorems, Theorem 3 says that the objective is "almost convex" and Theorem 4 says that the objective is "semi-smooth."

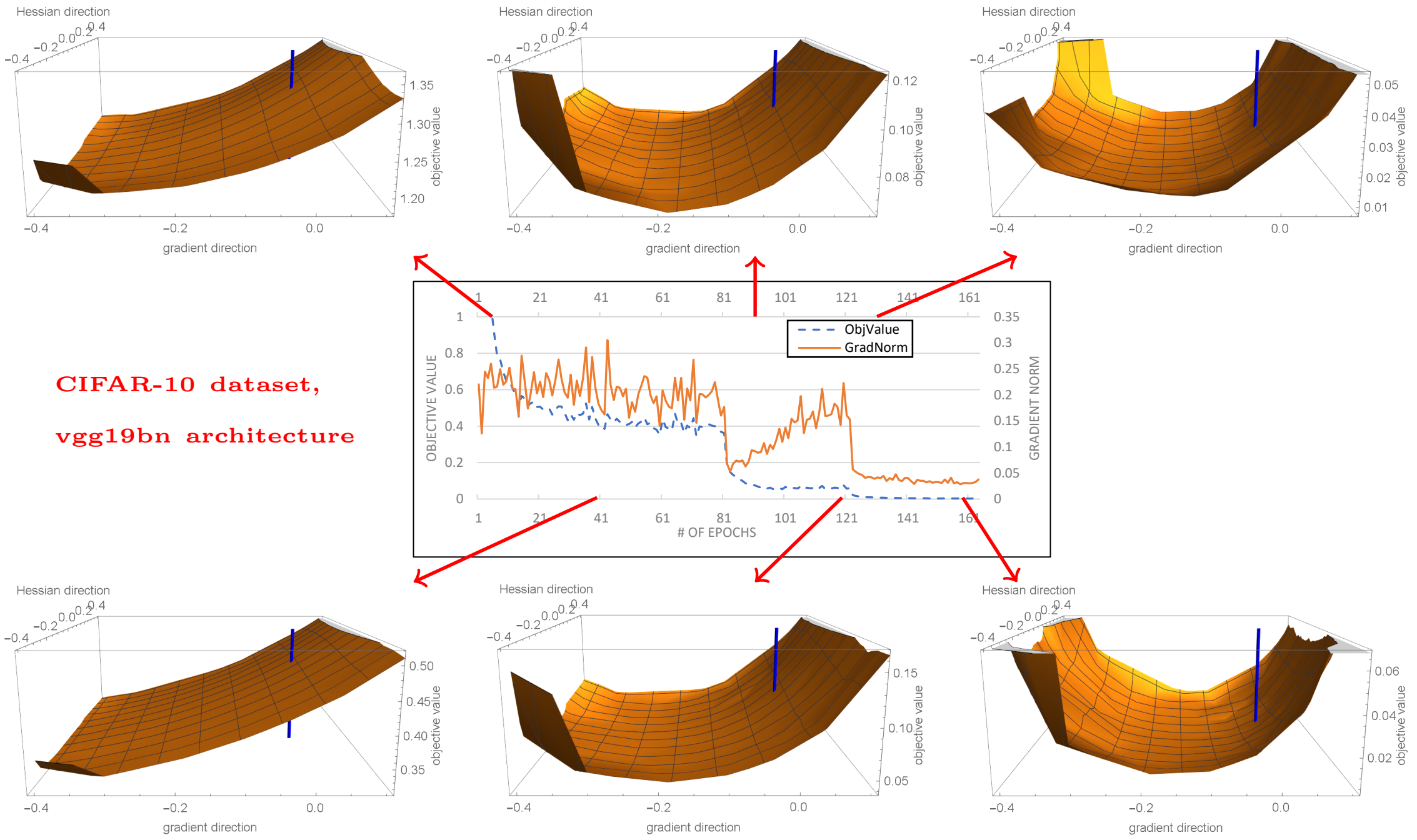

Figure 1: Landscapes of the CIFAR10 image-classification training objective $F(W)$ near the SGD training trajectory. The blue vertical stick marks the current point $W = W_t$ at the current iteration $t$. The $x$ and $y$ axes represent the gradient direction $\nabla F(W_t)$ and the most negatively curved direction of the Hessian after smoothing (approximately found by Oja's method [2, 3]). The $z$ axis represents the objective value.

---

**Observation.** As far as minimizing objective is concerned, the (negative) gradient direction sufficiently decreases the training objective, and it is not needed to use second-order method to find negative curvature. This is consistent with our findings Theorem 3 and 4.

**Remark 1.** Gradient norm does not tend to zero because cross-entropy loss is not strongly convex (see Section 6).
**Remark 2.** The task is CIFAR10 (for CIFAR100 or CIFAR10 with noisy label, see Figure 2 through 7 in appendix).
**Remark 3.** Architecture is VGG19 (for Resnet-32 or ResNet-110, see Figure 2 through 7 in appendix).
**Remark 4.** The six plots correspond to epochs 5, 40, 90, 120, 130 and 160. We start with learning rate 0.1, and decrease it to 0.01 at epoch 81, and to 0.001 at epoch 122. SGD with momentum 0.9 is used. The training code is unchanged from [58] and we only write new code for plotting such landscapes.

**Theorem 3** (no critical point)**.** *With probability* $\geq 1 - e^{-\Omega(m/\mathsf{poly}(n,L,\delta^{-1}))}$ *over randomness* $\overrightarrow{\mathbf{W}}^{(0)}, \mathbf{A}, \mathbf{B}$, *it satisfies for every* $\ell \in [L]$, *every* $i \in [n]$, *and every* $\overrightarrow{\mathbf{W}}$ *with* $\|\overrightarrow{\mathbf{W}} - \overrightarrow{\mathbf{W}}^{(0)}\|_2 \leq \frac{1}{\mathsf{poly}(n,L,\delta^{-1})}$,

$$\|\nabla F(\overrightarrow{\mathbf{W}})\|_F^2 \leq O\Big(F(\overrightarrow{\mathbf{W}}) \times \frac{Lnm}{d}\Big) \quad \textit{and} \quad \|\nabla F(\overrightarrow{\mathbf{W}})\|_F^2 \geq \Omega\Big(F(\overrightarrow{\mathbf{W}}) \times \frac{\delta m}{dn^2}\Big) \ .$$

The first property above is easy to prove, while the second property above says that as long as the objective is large, the gradient norm is also large. (See also Figure 1.) This means, when we are sufficiently close to the random initialization, there is no saddle point or critical point of any order.

Theorem 3 gives us hope to find *global minima* of the objective $F(\overrightarrow{\mathbf{W}})$, but is not enough. If we follow the negative gradient direction of $F(\overrightarrow{\mathbf{W}})$, how can we guarantee that the objective truly decreases? Classical optimization theory usually relies on objective's (Lipscthiz) smoothness [42] to derive an objective-decrease guarantee. Unfortunately, smoothness property at least requires the objective to be twice differentiable, but ReLU activation is not. To deal with this issue, we prove the following.

**Theorem 4** (semi-smoothness)**.** *With probability at least* $1 - e^{-\Omega(m/\mathsf{poly}(L,\log m))}$ *over the randomness of* $\overrightarrow{\mathbf{W}}^{(0)}, \mathbf{A}, \mathbf{B}$, *we have: for every* $\breve{\overrightarrow{\mathbf{W}}} \in (\mathbb{R}^{m\times m})^L$ *with* $\|\breve{\overrightarrow{\mathbf{W}}} - \overrightarrow{\mathbf{W}}^{(0)}\|_2 \leq \frac{1}{\mathsf{poly}(L,\log m)}$, *and for* *every* $\overrightarrow{\mathbf{W}}' \in (\mathbb{R}^{m\times m})^L$ *with* $\|\overrightarrow{\mathbf{W}}'\|_2 \le \frac{1}{\mathsf{poly}(L,\log m)}$, *the following inequality holds*

$$F(\breve{\overrightarrow{\mathbf{W}}} + \overrightarrow{\mathbf{W}}') \le F(\breve{\overrightarrow{\mathbf{W}}}) + \langle \nabla F(\breve{\overrightarrow{\mathbf{W}}}), \overrightarrow{\mathbf{W}}' \rangle + \frac{\mathsf{poly}(L)\sqrt{nm\log m}}{\sqrt{d}} \cdot \|\overrightarrow{\mathbf{W}}'\|_2 \big(F(\breve{\overrightarrow{\mathbf{W}}})\big)^{1/2} + O\big(\frac{nL^2 m}{d}\big)\|\overrightarrow{\mathbf{W}}'\|_2^2$$

Different from classical smoothness, we still have a first-order term $\|\overrightarrow{\mathbf{W}}'\|_2$ on the right hand side, while classical smoothness only has a second-order term $\|\overrightarrow{\mathbf{W}}'\|_2^2$. As one can see in our final proofs, as $m$ goes larger, the effect of the first-order term becomes smaller comparing to the second-order term. This brings Theorem 4 closer, but still not identical, to the classical Lipschitz smoothness.

**Back to Theorem 1 and 2.** The derivation of Theorem 1+2 from Theorem 3+4 is quite straightforward, and can be found in Section 12 and 13. At a high level, we show that GD/SGD can converge fast enough so that the weights stay close to random initialization by spectral norm bound $\frac{1}{\mathsf{poly}(n,L,\delta^{-1})}$. This ensures Theorem 3 and 4 both apply.[8]

In practice, one often goes beyond this theory-predicted spectral-norm boundary. However, quite interestingly, we still observe Theorem 3 and 4 hold in practice (see Figure 1). The gradient is sufficiently large and going in its negative direction can indeed decrease the objective.

## 4.2 Equivalence to Neural Tangent Kernel

Recall on input $x \in \mathbb{R}^{\mathfrak{d}}$, the network output $y(\overrightarrow{\mathbf{W}}; x) \stackrel{\text{def}}{=} y = \mathbf{B}h_L \in \mathbb{R}^d$ is a function of the weights $\overrightarrow{\mathbf{W}}$. Let us here focus on $d = 1$ for notational simplicity and leave $d > 1$ to the appendix. The *neural tangent kernel (NTK)* [32] is usually referred to as the feature space defined by the network gradient at random initialization. In other words,

- Given two inputs $x, \widetilde{x} \in \mathbb{R}^{\mathfrak{d}}$, the NTK kernel function is given as
$$K^{\mathsf{ntk}}(x, \widetilde{x}) \stackrel{\text{def}}{=} \langle \nabla y(\overrightarrow{\mathbf{W}}^{(0)}; x), \nabla y(\overrightarrow{\mathbf{W}}^{(0)}; \widetilde{x}) \rangle$$
- Given weight matrix tuple $\overrightarrow{\mathbf{W}}'$, the NTK model computes (we call the NTK objective)
$$y^{\mathsf{ntk}}(\overrightarrow{\mathbf{W}}'; x) \stackrel{\text{def}}{=} \langle \nabla y(\overrightarrow{\mathbf{W}}^{(0)}; x), \overrightarrow{\mathbf{W}}' \rangle = \textstyle\sum_{\ell=1}^L \langle \nabla_{\mathbf{W}_\ell} y(\overrightarrow{\mathbf{W}}^{(0)}; x), \mathbf{W}'_\ell \rangle \ .$$

In contrast, the dynamic NTK is given by arbitrary weight tuple $\overrightarrow{\mathbf{W}} = \overrightarrow{\mathbf{W}}^{(0)} + \overrightarrow{\mathbf{W}}'$ that may not be at random initialization. Jacot et al. [32] proved in their original paper that, when $m$ is infinite, dynamic NTK and NTK are identical because during the training process $\lim_{m\to\infty} \|\overrightarrow{\mathbf{W}}'\|_2 = 0$ if $\overrightarrow{\mathbf{W}} = \overrightarrow{\mathbf{W}}^{(0)} + \overrightarrow{\mathbf{W}}'$ is output of gradient descent.

In this paper, we complement Jacot et al. [32] by showing a *polynomial* bound on this equivalence, for *any* point that is within a certain ball of $\overrightarrow{\mathbf{W}}^{(0)}$. It is a simple corollary of Theorem 3 and Theorem 4, but we state it independently here since it may be of additional interest.[9]

**Theorem 5.** *Let* $\overrightarrow{\mathbf{W}}^{(0)}, \mathbf{A}, \mathbf{B}$ *be at random initialization. For fixed unit vectors* $x, \widetilde{x} \in \mathbb{R}^{\mathfrak{d}}$, *every (small) parameter* $\omega \le \frac{1}{\mathsf{poly}(L,\log m)}$, *with probability at least* $1 - e^{-\Omega(m\omega^{2/3}L)}$ *over* $\overrightarrow{\mathbf{W}}^{(0)}, \mathbf{A}, \mathbf{B}$, *we have for all* $\overrightarrow{\mathbf{W}}'$ *with* $\|\overrightarrow{\mathbf{W}}'\|_2 \le \omega$,

*(a)* $\|\nabla y(\overrightarrow{\mathbf{W}}^{(0)} + \overrightarrow{\mathbf{W}}'; x) - \nabla y^{\mathsf{ntk}}(\overrightarrow{\mathbf{W}}'; x)\|_F \le \widetilde{O}\big(\omega^{1/3}L^3\big) \cdot \|\nabla y^{\mathsf{ntk}}(\overrightarrow{\mathbf{W}}'; x)\|_F$;

*(b)* $y(\overrightarrow{\mathbf{W}}^{(0)} + \overrightarrow{\mathbf{W}}'; x) = y(\overrightarrow{\mathbf{W}}^{(0)}; x) + y^{\mathsf{ntk}}(\overrightarrow{\mathbf{W}}'; x) \pm \widetilde{O}\big(L^3\omega^{4/3}\sqrt{m}\big)$; *and*

[8] This spectral norm bound seems small, but is in fact quite large: it can totally change the outputs and fit the training data, because weights are randomly initialized (per entry) at around $\frac{1}{\sqrt{m}}$ for $m$ being large.

[9] Although Theorem 5 was not explicitly stated until version 5 of this paper, its proof was fully contained in the proofs of Theorem 1 and 2. Since some readers cannot find it, we state it here as a separate theorem.

*(c)* $\langle \nabla y(\overrightarrow{\mathbf{W}}^{(0)} + \overrightarrow{\mathbf{W}}'; x), \nabla y(\overrightarrow{\mathbf{W}}^{(0)} + \overrightarrow{\mathbf{W}}'; \widetilde{x}) \rangle = K^{\mathsf{ntk}}(x, \widetilde{x}) \pm \widetilde{O}(\omega^{1/3} L^3) \cdot \sqrt{K^{\mathsf{ntk}}(x,x) K^{\mathsf{ntk}}(\widetilde{x}, \widetilde{x})}$ .

*Theorem 5a and 5c says that dynamic NTK and NTK are almost equivalent up to a small multiplicative factor as long as* $\omega < \frac{1}{\mathsf{poly}(L, \log m)}$*; while Theorem 5b says that the NTK objective is almost exactly the first-order approximation of the neural network output as long as* $\omega < \frac{1}{m^{3/8}\mathsf{poly}(L)}$*.*

In comparison, in Theorem 1 and 2, GD/SGD outputs $\overrightarrow{\mathbf{W}}'$ satisfying $\omega \leq \frac{\mathsf{poly}(n, \delta^{-1})}{\sqrt{m}} \ll \frac{1}{m^{3/8}\mathsf{poly}(L)}$ under the assumption $m \geq \mathsf{poly}(n, L, \delta^{-1})$.[10] Thus, Theorem 5b implies $\overrightarrow{\mathbf{W}}'$ is also a solution to the NTK regression objective.

*Remark* 4.1. If one wishes to have $y(\overrightarrow{\mathbf{W}}^{(0)} + \overrightarrow{\mathbf{W}}'; x) \approx y^{\mathsf{ntk}}(\overrightarrow{\mathbf{W}}'; x)$ without the zero-order term $y(\overrightarrow{\mathbf{W}}^{(0)}; x)$, this can be achieved by properly scaling down the random initialization by a factor of the target error $\varepsilon < 1$. This was used in the follow-up [5] to achieve small generalization error on over-parameterized neural networks.

# 5 Proof Overview

Our proof to the Theorem 3 and 4 mostly consist of the following steps.

**Step 1: properties at random initialization.** Let $\overrightarrow{\mathbf{W}} = \overrightarrow{\mathbf{W}}^{(0)}$ be at random initialization and $h_{i,\ell}$ and $\mathbf{D}_{i,\ell}$ be defined with respect to $\overrightarrow{\mathbf{W}}$. We first show that forward propagation neither explode or vanish. That is,

$$\|h_{i,\ell}\| \approx 1 \text{ for all } i \in [n] \text{ and } \ell \in [L].$$

This is basically because for a fixed $y$, we have $\|\mathbf{W}y\|^2$ is around 2, and if its signs are sufficiently random, then ReLU activation kills half of the norm, that is $\|\phi(\mathbf{W}y)\| \approx 1$. Then applying induction finishes the proof.

Analyzing forward propagation is not enough. We also need spectral norm bounds on the backward matrix and on the intermediate matrix

$$\|\mathbf{B}\mathbf{D}_{i,L}\mathbf{W}_L \cdots \mathbf{D}_{i,a}\mathbf{W}_a\|_2 \leq O(\sqrt{m/d}) \quad \text{and} \quad \|\mathbf{D}_{i,a}\mathbf{W}_a \cdots \mathbf{D}_{i,b}\mathbf{W}_b\|_2 \leq O(\sqrt{L}) \tag{5.1}$$

for every $a, b \in [L]$. Note that if one naively bounds the spectral norm by induction, then $\|\mathbf{D}_{i,a}\mathbf{W}_a\|_2 \approx 2$ and it will *exponentially blow up!* Our careful analysis ensures that even when $L$ layers are stacked together, there is no exponential blow up in $L$.

The final lemma in this step proves that, as long as $\|x_i - x_j\| \geq \delta$, then

$$\|h_{i,\ell} - h_{j,\ell}\| \geq \Omega(\delta) \text{ for each layer } \ell \in [L].$$

Again, if one is willing to sacrifice an exponential factor and prove a lower bound $\delta \cdot 2^{-\Omega(L)}$, this will be easy. What is hard is to derive such lower bound without sacrificing more than a constant factor, but under the condition of $\delta \leq \frac{1}{CL}$. Details are in Section 7.

**Step 2: stability after adversarial perturbation.** We show that for every $\overrightarrow{\mathbf{W}}$ that is "close" to initialization, meaning $\|\overrightarrow{\mathbf{W}} - \overrightarrow{\mathbf{W}}^{(0)}\|_2 \leq \omega$ for some $\omega \leq \frac{1}{\mathsf{poly}(L)}$, then

(a) the number of sign changes $\|\mathbf{D}_{i,\ell} - \mathbf{D}^{(0)}_{i,\ell}\|_0$ is at most $O(m\omega^{2/3}L) \ll m$, and

(b) the perturbation amount $\|h_{i,\ell} - h^{(0)}_{i,\ell}\| \leq O(\omega L^{5/2}) \ll 1$.

[10] See (12.1) and (13.1) in the proofs.

Since $\omega \leq \frac{1}{\mathsf{poly}(L)}$, both changes above become negligible. We call this "forward stability", and it is the most technical proof of this paper. Intuitively, both "(a) implies (b)" and "(b) implies (a)" are trivial to prove by matrix concentration.[11] Unfortunately, one cannot apply such derivation by induction, because constants will blow up exponentially in the number of layers. We need some careful double induction introduced by [6], and details in Section 8.1. Another main result in this step is to derive stability for the backward matrix and the intermediate matrix. We show that when $w \leq \mathsf{poly}(L)$, (5.1) remains to hold. Details are in Section 8.2 and 8.3.

*Remark.* In the final proof, $\overrightarrow{\mathbf{W}}$ is a point obtained by GD/SGD starting from $\overrightarrow{\mathbf{W}}^{(0)}$, and thus $\overrightarrow{\mathbf{W}}$ may depend on the randomness of $\overrightarrow{\mathbf{W}}^{(0)}$. Since we cannot control how such randomness correlates, we argue for the above stability properties against *all possible* $\overrightarrow{\mathbf{W}}$. This is why we call it "stability against adversarial perturbation."

**Step 3: gradient bound.** The hard part of Theorem 3 is to show gradient lower bound. For this purpose, recall from Fact 2.6 that each sample $i \in [n]$ contributes to the full gradient matrix by $\mathbf{D}_{i,\ell}(\mathsf{Back}_{i,\ell+1}^\top \mathsf{loss}_i) h_{i,\ell-1}^\top$, where the backward matrix is applied to a loss vector $\mathsf{loss}_i$. To show this is large, intuitively, one wishes to show $(\mathsf{Back}_{i,\ell+1}^\top \mathsf{loss}_i)$ and $h_{i,\ell-1}$ are both vectors with large Euclidean norm.

Thanks to Step 1 and 2, this is not hard for a *single* sample $i \in [n]$. For instance, $\|h_{i,\ell-1}^{(0)}\| \approx 1$ by Step 1 and we know $\|h_{i,\ell-1} - h_{i,\ell-1}^{(0)}\| \leq o(1)$ from Step 2. One can also argue for $\mathsf{Back}_{i,\ell+1}^\top \mathsf{loss}_i$ but this is a bit harder. Indeed, when moving from random initialization $\overrightarrow{\mathbf{W}}^{(0)}$ to $\overrightarrow{\mathbf{W}}$, the loss vector $\mathsf{loss}_i$ can change completely. Fortunately, $\mathsf{loss}_i \in \mathbb{R}^d$ is a low-dimensional vector, so one can calculate $\|\mathsf{Back}_{i,\ell+1}^\top u\|$ for every fixed $u$ and then apply $\varepsilon$-net.

Finally, how to combine the above argument with multiple samples $i \in [n]$? These matrices are clearly not independent and may (in principle) sum up to zero. To deal with this, we use $\|h_{i,\ell} - h_{j,\ell}\| \geq \Omega(\delta)$ from Step 1. In other words, even if the contribution matrix $\mathbf{D}_{i,\ell}(\mathsf{Back}_{i,\ell+1}^\top \mathsf{loss}_i) h_{i,\ell-1}^\top$ with respect to one sample $i$ is fixed, the contribution matrix with respect to other samples $j \in [n] \setminus \{i\}$ are still sufficiently random. Thus, the final gradient matrix will still be large. This idea comes from the prior work [36], [12] and helps us prove Theorem 3. Details in Appendix 9 and 10.

**Step 4: semi-smoothness.** In order to prove Theorem 4, one needs to argue, if we are currently at $\breve{\overrightarrow{\mathbf{W}}}$ and perturb it by $\overrightarrow{\mathbf{W}}'$, then how much does the objective change in second and higher order terms. This is different from our stability theory in Step 2, because Step 2 is regarding having a perturbation on $\overrightarrow{\mathbf{W}}^{(0)}$; in contrast, in Theorem 4 we need a (small) perturbation $\overrightarrow{\mathbf{W}}'$ on top of $\breve{\overrightarrow{\mathbf{W}}}$, which may already be a point perturbed from $\overrightarrow{\mathbf{W}}^{(0)}$. Nevertheless, we still manage to show that, if $\breve{h}_{i,\ell}$ is calculated on $\breve{\overrightarrow{\mathbf{W}}}$ and $h_{i,\ell}$ is calculated on $\breve{\overrightarrow{\mathbf{W}}} + \overrightarrow{\mathbf{W}}'$, then $\|h_{i,\ell} - \breve{h}_{i,\ell}\| \leq O(L^{1.5})\|\mathbf{W}'\|_2$. This, along with other properties to prove, ensures semi-smoothness. This explains Theorem 4 and details are in Section 11.

*Remark.* In other words, the amount of changes to each hidden layer (i.e., $h_{i,\ell} - \breve{h}_{i,\ell}$) is proportional to the amount of perturbation $\|\mathbf{W}'\|_2$. This may sound familiar to some readers: a ReLU function is Lipschitz continuous $|\phi(a) - \phi(b)| \leq |a - b|$, and composing Lipschitz functions still yield Lipschitz functions. What is perhaps surprising here is that this "composition" does not create exponential blow-up in the Lipschitz continuity parameter, as long as the amount of over-parameterization is sufficient and $\overrightarrow{\breve{\mathbf{W}}}$ is close to initialization.

[11] Namely, if the number of sign changes is bounded in all layers, then $h_{i,\ell}$ and $h_{i,\ell}^{(0)}$ cannot be too far away by applying matrix concentration; and reversely, if $h_{i,\ell}$ is not far from $h_{i,\ell}^{(0)}$ in all layers, then the number of sign changes per layer must be small.

[12] This is the only technical idea that we borrowed from Li and Liang [36], which is the over-parameterization theory for 2-layer neural networks.

# 6 Notable Extensions

Our Step 1 through Step 4 in Section 5 in fact give rise to a general plan for proving the training convergence of any neural network (at least with respect to the ReLU activation). Thus, it is expected that it can be generalized to many other settings. Not only we can have different number of neurons each layer, our theorems can be extended at least in the following three major directions.[13]

**Different loss functions.** There is absolutely no need to restrict only to $\ell_2$ regression loss. We prove in Appendix A that, for any Lipschitz-smooth loss function $f$:

**Theorem 6** (arbitrary loss)**.** *From random initialization, with probability at least* $1 - e^{-\Omega(\log^2 m)}$*, gradient descent with appropriate learning rate satisfy the following.*

- *If $f$ is nonconvex but $\sigma$-gradient dominant (a.k.a. Polyak-Łojasiewicz), GD finds $\varepsilon$-error minimizer in*[14]
$$T = \widetilde{O}\big(\tfrac{\mathsf{poly}(n,L)}{\sigma\delta^2} \cdot \log \tfrac{1}{\varepsilon}\big) \text{ iterations}$$
*as long as* $m \geq \widetilde{\Omega}\big(\mathsf{poly}(n, L, \delta^{-1}) \cdot d\sigma^{-2}\big)$.
- *If $f$ is convex, then GD finds $\varepsilon$-error minimizer in*
$$T = \widetilde{O}\big(\tfrac{\mathsf{poly}(n,L)}{\delta^2} \cdot \tfrac{1}{\varepsilon}\big) \text{ iterations}$$
*as long as* $m \geq \widetilde{\Omega}\big(\mathsf{poly}(n, L, \delta^{-1}) \cdot d \log \varepsilon^{-1}\big)$.
- *If $f$ is non-convex, then SGD finds a point with $\|\nabla f\| \leq \varepsilon$ in at most*[15]
$$T = \widetilde{O}\big(\tfrac{\mathsf{poly}(n,L)}{\delta^2} \cdot \tfrac{1}{\varepsilon^2}\big) \text{ iterations}$$
*as long as* $m \geq \widetilde{\Omega}\big(\mathsf{poly}(n, L, \delta^{-1}) \cdot d\varepsilon^{-1}\big)$.
- *If $f$ is cross-entropy for multi-label classification, then GD attains* 100% *training accuracy in at most*[16]*.*
$$T = \widetilde{O}\big(\tfrac{\mathsf{poly}(n,L)}{\delta^2}\big) \text{ iterations}$$
*as long as* $m \geq \widetilde{\Omega}\big(\mathsf{poly}(n, L, \delta^{-1}) \cdot d\big)$.

We remark here that the $\ell_2$ loss is 1-gradient dominant so it falls into the above general Theorem 6. One can also derive similar bounds for (mini-batch) SGD so we do not repeat the statements here.

**Convolutional neural networks (CNN).** There are lots of different ways to design CNN and each of them may require somewhat different proofs. In Appendix B, we study the case when

[13] In principle, each such proof may require a careful rewriting of the main body of this paper. We choose to sketch only the proof difference (in the appendix) in order to keep this paper short. If there is sufficient interest from the readers, we can consider adding the full proofs in the future revision of this paper.

[14] Note that the loss function when combined with the neural network together $f(\mathbf{B}h_{i,L})$ is *not* gradient dominant. Therefore, one cannot apply classical theory on gradient dominant functions to derive our same result.

[15] Again, this cannot be derived from classical theory of finding approximate saddle points for non-convex functions, because weights $\overrightarrow{\mathbf{W}}$ with small $\|\nabla f(\mathbf{B}h_{i,L})\|$ is a very different (usually much harder) task comparing to having small gradient with respect to $\overrightarrow{\mathbf{W}}$ for the entire composite function $f(\mathbf{B}h_{i,L})$.

[16] This is because attaining constant objective error $\varepsilon = 1/4$ for the cross-entropy loss suffices to imply perfect training accuracy.

$\mathbf{A}, \mathbf{W}_1, \dots, \mathbf{W}_{L-1}$ are convolutional while $\mathbf{W}_L$ and $\mathbf{B}$ are fully connected. We assume for notational simplicity that each hidden layer has $\mathfrak{d}$ points each with $m$ channels. (In vision tasks, a point is a pixel). In the most general setting, these values $\mathfrak{d}$ and $m$ can vary across layers. We prove the following theorem:

**Theorem 7** (CNN). *As long as $m \geq \widetilde{\Omega}\big(\mathsf{poly}(n, L, \mathfrak{d}, \delta^{-1}) \cdot d\big)$, with high probability, GD and SGD find an $\varepsilon$-error solution for $\ell_2$ regression in $T = \widetilde{O}\big(\frac{\mathsf{poly}(n,L,\mathfrak{d})}{\delta^2} \cdot \log \varepsilon^{-1}\big)$ iterations for CNN.*

Of course, one can replace $\ell_2$ loss with other loss functions in Theorem 6 to get different types of convergence rates. We do not repeat them here.

**Residual neural networks (ResNet).** There are lots of different ways to design ResNet and each of them may require somewhat different proofs. In symbols, between two layers, one may study $h_\ell = \phi(h_{\ell-1} + \mathbf{W} h_{\ell-1})$, $h_\ell = \phi(h_{\ell-1} + \mathbf{W}_2 \phi(\mathbf{W}_1 h_{\ell-1}))$, or even $h_\ell = \phi(h_{\ell-1} + \mathbf{W}_3 \phi(\mathbf{W}_2 \phi(\mathbf{W}_1 h_{\ell-1})))$. Since the main purpose here is to illustrate the generality of our techniques but not to attack each specific setting, in Appendix C, we choose to consider the simplest residual setting $h_\ell = \phi(h_{\ell-1} + \mathbf{W} h_{\ell-1})$ (that was also studied for instance by theoretical work [27]). With appropriately chosen random initialization, we prove the following theorem:

**Theorem 8** (ResNet). *As long as $m \geq \widetilde{\Omega}\big(\mathsf{poly}(n, L, \delta^{-1}) \cdot d\big)$, with high probability, GD and SGD find an $\varepsilon$-error solution for $\ell_2$ regression in $T = \widetilde{O}\big(\frac{\mathsf{poly}(n,L)}{\delta^2} \cdot \log \varepsilon^{-1}\big)$ iterations for ResNet.*

Of course, one can replace $\ell_2$ loss with other loss functions in Theorem 6 to get different types of convergence rates. We do not repeat them here.

# Detailed Proofs

- In Section 7, we derive network properties at random initialization.
- In Section 8, we derive the stability theory against adversarial perturbation.
- In Section 9, we gradient upper and lower bounds at random initialization.
- In Section 10, we prove Theorem 3.
- In Section 11, we prove Theorem 4.
- In Section 12, we prove Theorem 1.
- In Section 13, we prove Theorem 2.
- In Section 14, we prove Theorem 5.

## 7 Properties at Random Initialization

Throughout this section we assume $\overrightarrow{\mathbf{W}}, \mathbf{A}$ and $\mathbf{B}$ are randomly generated according to Def. 2.3. The diagonal sign matrices $\mathbf{D}_{i,\ell}$ are also determined according to this random initialization.

### 7.1 Forward Propagation

**Lemma 7.1** (forward propagation). *If $\varepsilon \in (0, 1]$, with probability at least $1 - O(nL) \cdot e^{-\Omega(m\varepsilon^2/L)}$ over the randomness of $\mathbf{A} \in \mathbb{R}^{m \times \mathfrak{d}}$ and $\overrightarrow{\mathbf{W}} \in (\mathbb{R}^{m \times m})^L$, we have*

$$\forall i \in [n], \ell \in \{0, 1, \dots, L\} \quad : \quad \|h_{i,\ell}\| \in [1 - \varepsilon, 1 + \varepsilon] \ .$$

*Remark.* Lemma 7.1 is in fact trivial to prove if the allowed failure probability is instead $e^{-\Omega(m\varepsilon^2/L^2)}$ (by applying concentration inequality layer by layer).

Before proving Lemma 7.1 we note a simple mathematical fact:

**Fact 7.2.** *Let $h, q \in \mathbb{R}^p$ be fixed vectors and $h \neq 0$, $\mathbf{W} \in \mathbb{R}^{m\times p}$ be random matrix with i.i.d. entries $\mathbf{W}_{i,j} \sim \mathcal{N}(0, \frac{2}{m})$, and vector $v \in \mathbb{R}^m$ defined as $v_i = \phi((\mathbf{W}h)_i) = \mathbb{1}_{(\mathbf{W}(h+q))_i \geq 0}(\mathbf{W}h)_i$. Then,*

- *$|v_i|$ follows i.i.d. from the following distribution: with half probability $|v_i| = 0$, and with the other half probability $|v_i|$ follows from folded Gaussian distributions $|\mathcal{N}(0, \frac{2\|h\|^2}{m})|$.*
- *$\frac{m\|v\|^2}{2\|h\|^2}$ is in distribution identical to $\chi^2_\omega$ (chi-square distribution of order $\omega$) where $\omega$ follows from binomial distribution $\mathcal{B}(m, 1/2)$.*

*Proof of Fact 7.2.* We assume each vector $\mathbf{W}_i$ is generated by first generating a gaussian vector $g \sim \mathcal{N}(0, \frac{2\mathbf{I}}{m})$ and then setting $\mathbf{W}_i = \pm g$ where the sign is chosen with half-half probability. Now, $|\langle \mathbf{W}_i, h\rangle| = |\langle g, h\rangle|$ only depends on $g$, and is in distribution identical to $|\mathcal{N}(0, \frac{2\|h\|^2}{m})|$. Next, after the sign is determined, the indicator $\mathbb{1}_{\langle \mathbf{W}_i, h+q\rangle \geq 0}$ is 1 with half probability and 0 with another half. Therefore, $|v_i|$ satisfies the aforementioned distribution. As for $\|v\|^2$, letting $\omega \in \{0, 1, \dots, m\}$ be the variable indicator how many indicators are 1, then $\omega \sim \mathcal{B}(m, 1/2)$ and $\frac{m\|v\|^2}{2\|h\|^2} \sim \chi^2_\omega$. □

*Proof of Lemma 7.1.* We only prove Lemma 7.1 for a fixed $i \in [n]$ and $\ell \in \{0, 1, 2, \dots, L\}$ because we can apply union bound at the end. Below, we drop the subscript $i$ for notational convenience, and write $h_{i,\ell}$ and $x_i$ as $h_\ell$ and $x$ respectively.

Letting $\Delta_\ell \stackrel{\text{def}}{=} \frac{\|h_\ell\|^2}{\|h_{\ell-1}\|^2}$, we can write

$$\log \|h_{b-1}\|^2 = \log \|x\|^2 + \sum_{\ell=0}^{b-1} \log \Delta_\ell = \sum_{\ell=0}^{b-1} \log \Delta_\ell \ .$$

According to Fact 7.2, fixing any $h_{\ell-1} \neq 0$ and letting $\mathbf{W}_\ell$ be the only source of randomness, we have $\frac{m}{2}\Delta_\ell \sim \chi^2_\omega$ where $\omega \sim \mathcal{B}(m, 1/2)$. For such reason, for each $\Delta_\ell$, we can write $\Delta_\ell = \Delta_{\ell,\omega}$ where $\frac{m}{2}\Delta_{\ell,\omega} \sim \chi^2_\omega$ and $\omega \sim \mathcal{B}(m, 1/2)$. In the analysis below, we condition on the event that $\omega \in [0.4m, 0.6m]$; this happens with probability $\geq 1 - e^{-\Omega(m)}$ for each layer $\ell \in [L]$. To simplify our notations, if this event does not hold, we set $\Delta_\ell = 1$.

**Expectation.** One can verify that $\mathbb{E}[\log \Delta_{\ell,\omega} \mid \omega] = \log \frac{4}{m} + \psi(\frac{\omega}{2})$ where $\psi(h) = \frac{\Gamma'(h)}{\Gamma(h)}$ is the digamma function. Using the bound $\log h - \frac{1}{h} \leq \psi(h) \leq \log h - \frac{1}{2h}$ of digamma function, we have

$$\log \frac{2\omega}{m} - \frac{2}{\omega} \leq \mathbb{E}[\log \Delta_{\ell,\omega} \mid \omega] \leq \log \frac{2\omega}{m} - \frac{1}{\omega}.$$

Whenever $\omega \in [0.4m, 0.6m]$, we can write

$$\log \frac{2\omega}{m} = \log\left(1 + \frac{2\omega - m}{m}\right) \geq \frac{2\omega - m}{m} - \left(\frac{2\omega - m}{m}\right)^2$$

It is easy to verify $\mathbb{E}_\omega\left[\frac{2\omega-m}{m}\right] = 0$ and $\mathbb{E}_\omega\left[\left(\frac{2\omega-m}{m}\right)^2\right] = \frac{1}{m}$. Therefore,

$$\mathbb{E}_\omega\left[\log \frac{2\omega}{m}\right] \geq -\frac{1}{m} - \mathbf{Pr}\left[\omega \notin [0.4m, 0.6m]\right] \cdot \log \frac{2}{m} \geq -\frac{2}{m}$$

Combining everything together, along with the fact that $\mathbb{E}_\omega[\log \frac{2\omega}{m}] \leq \log \frac{\mathbb{E}[2\omega]}{m} = 0$, we have (when $m$ is sufficiently larger than a constant)

$$-\frac{4}{m} \leq \mathbb{E}[\log \Delta_\ell] \leq 0. \tag{7.1}$$

**Subgaussian Tail.** By standard tail bound for chi-square distribution, we know that

$$\forall t \in [0,\infty)\colon \quad \mathbf{Pr}\left[\left|\frac{m}{2}\Delta_{\ell,\omega} - \omega\right| \le t \,\middle|\, \omega\right] \ge 1 - 2e^{-\Omega(t^2/\omega)} - e^{-\Omega(t)} \ .$$

Since we only need to focus on $\omega \ge 0.4m$, this means

$$\forall t \in [0,m]\colon \quad \mathbf{Pr}\left[\left|\frac{m}{2}\Delta_{\ell,\omega} - \omega\right| \le t \,\middle|\, \omega \ge 0.4m\right] \ge 1 - O(e^{-\Omega(t^2/m)}) \ .$$

On the other hand, by Chernoff-Hoeffding bound, we also have

$$\mathbf{Pr}_{\omega}\left[\left|\omega - \frac{m}{2}\right| \le t\right] \ge 1 - O(e^{-\Omega(t^2/m)})$$

Together, using the definition $\Delta_\ell = \Delta_{\ell,\omega}$ (or $\Delta_\ell = 1$ if $\omega \notin [0.4m, 0.6m]$), we obtain

$$\forall t \in [0,m]\colon \quad \mathbf{Pr}\left[\left|\frac{m}{2}\Delta_\ell - \frac{m}{2}\right| \le t\right] \ge 1 - O(e^{-\Omega(t^2/m)}) \ .$$

This implies,

$$\forall t \in \left[0, \frac{m}{4}\right]\colon \quad \mathbf{Pr}\left[|\log \Delta_\ell| \le \frac{t}{m}\right] \ge 1 - O(e^{-\Omega(t^2/m)}) \ . \tag{7.2}$$

Now, let us make another simplification: define $\widehat{\Delta}_\ell = \Delta_\ell$ if $|\log \Delta_\ell| \le \frac{1}{4}$ and $\widehat{\Delta}_\ell = 1$ otherwise. In this way, (7.2) implies that $X = \log \widehat{\Delta}_\ell$ is an $O(m)$-subgaussian random variable.

**Concentration.** Using martingale concentration on subgaussian variables (see for instance [47]), we have for $\varepsilon \in (0,1]$,

$$\mathbf{Pr}\left[\left|\sum_{\ell=0}^{b-1} \log \widehat{\Delta}_\ell - \mathbb{E}[\log \widehat{\Delta}_\ell]\right| > \varepsilon\right] \le O\big(e^{-\Omega(\varepsilon^2 m/L)}\big).$$

Since with probability $\ge 1 - Le^{-\Omega(m)}$ it satisfies $\widehat{\Delta}_\ell = \Delta_\ell$ for all $\ell \in [L]$, combining this with (7.1), we have

$$\mathbf{Pr}\left[\left|\sum_{\ell=0}^{b-1} \log \Delta_\ell\right| > \varepsilon\right] \le O\big(e^{-\Omega(\varepsilon^2 m/L)}\big).$$

In other words, $\|h_{b-1}\|^2 \in \big[1-\varepsilon, 1+\varepsilon\big]$ with probability at least $1 - O\big(e^{-\Omega(\varepsilon^2 m/L)}\big)$. □

## 7.2 Intermediate Layers

**Lemma 7.3** (intermediate layers)**.** *Suppose $m \ge \Omega(nL\log(nL))$. With probability at least $\ge 1 - e^{-\Omega(m/L)}$ over the randomness of $\overrightarrow{\mathbf{W}} \in (\mathbb{R}^{m\times m})^L$, for all $i \in [n], 1 \le a \le b \le L$,*

*(a)* $\|\mathbf{W}_b \mathbf{D}_{i,b-1}\mathbf{W}_{b-1}\cdots \mathbf{D}_{i,a}\mathbf{W}_a\|_2 \le O(\sqrt{L})$.

*(b)* $\|\mathbf{W}_b \mathbf{D}_{i,b-1}\mathbf{W}_{b-1}\cdots \mathbf{D}_{i,a}\mathbf{W}_a v\| \le 2\|v\|$ *for all vectors $v$ with* $\|v\|_0 \le O\big(\frac{m}{L\log m}\big)$.

*(c)* $\|u^\top \mathbf{W}_b \mathbf{D}_{i,b-1}\mathbf{W}_{b-1}\cdots \mathbf{D}_{i,a}\mathbf{W}_a\| \le O(1)\|u\|$ *for all vectors $u$ with* $\|u\|_0 \le O\big(\frac{m}{L\log m}\big)$.

*For any integer $s$ with $1 \le s \le O\big(\frac{m}{L\log m}\big)$, with probability at least $1 - e^{-\Omega(s\log m)}$ over the randomness of $\overrightarrow{\mathbf{W}} \in (\mathbb{R}^{m\times m})^L$:*

*(d)* $|u^\top \mathbf{W}_b \mathbf{D}_{i,b-1}\mathbf{W}_{b-1}\cdots \mathbf{D}_{i,a}\mathbf{W}_a v| \le \|u\|\|v\| \cdot O\big(\frac{\sqrt{s\log m}}{\sqrt{m}}\big)$ *for all vectors $u,v$ with* $\|u\|_0, \|v\|_0 \le s$.

*Proof.* Again we prove the lemma for fixed $i, a$ and $b$ because we can take a union bound at the end. We drop the subscript $i$ for notational convenience.

(a) Let $z_{a-1}$ be any fixed unit vector, and define $z_\ell = \mathbf{D}_\ell \mathbf{W}_\ell \cdots \mathbf{D}_a \mathbf{W}_a z_{a-1}$. According to Fact 7.2 again, fixing any $z_{\ell-1}$ and letting $\mathbf{W}_\ell$ be the only source of randomness, defining $\Delta_\ell \stackrel{\text{def}}{=} \frac{\|z_\ell\|^2}{\|z_{\ell-1}\|^2}$, we have that $\frac{m}{2}\Delta_\ell$ is distributed according to a $\chi^2_\omega$ where $\omega \sim \mathcal{B}(m, \frac{1}{2})$. Therefore, we have

$$\log \|z_{b-1}\|^2 = \log \|z_{a-1}\|^2 + \sum_{\ell=a}^{b-1} \log \Delta_\ell = \sum_{\ell=a}^{b-1} \log \Delta_\ell \enspace .$$

Using exactly the same proof as Lemma 7.1, we have

$$\|z_{b-1}\|^2 = \|\mathbf{W}_b \mathbf{D}_{b-1} \mathbf{W}_{b-1} \cdots \mathbf{D}_a \mathbf{W}_a z_{a-1}\|^2 \in \big[1 - 1/3, 1 + 1/3\big]$$

with probability at least $1 - e^{-\Omega(m/L)}$. As a result, if we fix a subset $M \subseteq [m]$ of cardinality $|M| \leq O(m/L)$, taking $\varepsilon$-net, we know that with probability at least $e^{-\Omega(m/L)}$, it satisfies

$$\|\mathbf{W}_b \mathbf{D}_{b-1} \mathbf{W}_{b-1} \cdots \mathbf{D}_a \mathbf{W}_a u\| \leq 2\|u\| \tag{7.3}$$

for all vectors $u$ whose coordinates are zeros outside $M$. Now, for an arbitrary unit vector $v \in \mathbb{R}^m$, we can decompose it as $v = u_1 + \cdots + u_N$ where $N = O(L)$, each $u_j$ is non-zero only at $O(m/L)$ coordinates, and the vectors $u_1, \dots, u_N$ are non-zeros on different coordinates. We can apply (7.3) for each each such $u_j$ and triangle inequality. This gives

$$\|\mathbf{W}_b \mathbf{D}_{b-1} \mathbf{W}_{b-1} \cdots \mathbf{D}_a \mathbf{W}_a v\| \leq 2 \sum_{j=1}^N \|u_j\| \leq 2\sqrt{N} \Big(\sum_{j=1}^N \|u_j\|^2\Big)^{1/2} \leq O(\sqrt{L}) \cdot \|v\|.$$

(b) The proof of Lemma 7.3b is the same as Lemma 7.3a, except to take $\varepsilon$-net over all $O\big(\frac{m}{L \log m}\big)$-sparse vectors $u$ and then applying union bound.

(c) Similar to the proof of Lemma 7.3a, for any fixed vector $v$, we have that with probability at least $1 - e^{-\Omega(m/L)}$ (over the randomness of $\mathbf{W}_{b-1}, \dots, \mathbf{W}_1, \mathbf{A}$),

$$\|\mathbf{D}_{b-1} \mathbf{W}_{b-1} \cdots \mathbf{D}_a \mathbf{W}_a v\| \leq 2\|v\|.$$

Conditioning on this event happens, using the randomness of $\mathbf{W}_b$, we have for each fixed vector $u \in \mathbb{R}^m$, we have

$$\Pr_{\mathbf{W}_b}\left[\left|u^\top \mathbf{W}_b \big(\mathbf{D}_{b-1} \mathbf{W}_{b-1} \cdots \mathbf{D}_a \mathbf{W}_a v\big)\right| \geq \frac{4}{\sqrt{L}} \|u\| \|v\|\right] \leq e^{-\Omega(m/L)}.$$

Now consider the case that $v$ is a sparse vector that is only non-zero over some fixed index set $M \subseteq [m]$ (with $|M| \leq O(m/L)$), and that $u$ is of sparsity $s = O\big(\frac{m}{L \log m}\big)$. Taking $\varepsilon$-net over all such possible vectors $u$ and $v$, we have with probability at least $1 - e^{-\Omega(m/L)}$, for all vectors $u \in \mathbb{R}^m$ with $\|u\|_0 \leq s$ and all vectors $v \in \mathbb{R}^m$ that have non-zeros only in $M$,

$$\left|u^\top \mathbf{W}_b \big(\mathbf{D}_{b-1} \mathbf{W}_{b-1} \cdots \mathbf{D}_a \mathbf{W}_a v\big)\right| \leq \frac{8}{\sqrt{L}} \|u\| \|v\| \enspace . \tag{7.4}$$

Back to the case when $v$ is an arbitrary vector, we can partition $[m]$ into $N$ index sets $[m] = M_1 \cup M_2 \cup \cdots \cup M_N$ and write $v = v_1 + v_2 + \cdots + v_N$, where $N = O(L)$ and each $v_j$ is non-zero only in $M_j$. By applying (7.4) for $N$ times and using triangle inequality, we have

$$\begin{aligned}\left|u^\top \mathbf{W}_b \big(\mathbf{D}_{b-1} \mathbf{W}_{b-1} \cdots \mathbf{D}_a \mathbf{W}_a v\big)\right| &\leq \sum_{j=1}^N \left|u^\top \mathbf{W}_b \big(\mathbf{D}_{b-1} \mathbf{W}_{b-1} \cdots \mathbf{D}_a \mathbf{W}_a v_j\big)\right| \\ &\leq \frac{8}{\sqrt{L}} \|u\| \times \sum_{j=1}^N \|v_j\| \leq O(1) \times \|u\| \|v\| \enspace .\end{aligned}$$

(d) We apply the same proof as Lemma 7.3c with minor changes to the parameters. We can show with probability at least $1 - e^{-\Omega(m/L)}$ (over the randomness of $\mathbf{W}_{b-1}, \dots, \mathbf{W}_1, \mathbf{A}$), for a fixed vector $v \in \mathbb{R}^m$:

$$\|\mathbf{D}_{b-1}\mathbf{W}_{b-1} \cdots \mathbf{D}_a \mathbf{W}_a v\| \le 2\|v\| \ .$$

Further using the randomness of $\mathbf{W}_b$, we have that conditioning on the above event, fixing any $u \in \mathbb{R}^m$, with probability at least $1 - e^{-\Omega(s \log m)}$ over the randomness of $\mathbf{W}_b$:

$$\left|u\mathbf{W}_b\big(\mathbf{D}_{b-1}\mathbf{W}_{b-1} \cdots \mathbf{D}_a \mathbf{W}_a v\big)\right| \le \big(\frac{s \log m}{m}\big)^{1/2} \times O\big(\|v\|\|u\|\big) \ .$$

Finally, taking $\varepsilon$-net over all possible vectors $u, v$ that are $s$ sparse, we have the desired result.

□

## 7.3 Backward Propagation

**Lemma 7.4** (backward propagation). *Suppose $m \ge \Omega(nL\log(nL))$. If $s \ge \Omega\big(\frac{d}{\log m}\big)$ and $s \le O\big(\frac{m}{L \log m}\big)$, then with probability at least $1 - e^{-\Omega(s \log m)}$, for all $i \in [n]$, $a = 1, 2, \dots, L+1$,*

*(a) $|v^\top \mathbf{B}\mathbf{D}_{i,L}\mathbf{W}_L \cdots \mathbf{D}_{i,a}\mathbf{W}_a u| \le O\big(\frac{\sqrt{s \log m}}{\sqrt{d}}\big)\|v\|\|u\|$ for all $v \in \mathbb{R}^d$ and all $u \in \mathbb{R}^m$ with $\|u\|_0 \le s$.*

*With probability at least $\ge 1 - e^{-\Omega(m/L)}$, for all $i \in [n], 1 \le a \le L$,*

*(b) $\|v^\top \mathbf{B}\mathbf{D}_{i,L}\mathbf{W}_L \cdots \mathbf{D}_{i,a}\mathbf{W}_a\| \le O(\sqrt{m/d})\|v\|$ for all vectors $u \in \mathbb{R}^d$ if $d \le O\big(\frac{m}{L \log m}\big)$.*

*Proof.* (a) The proof follows the same idea of Lemma 7.3 (but choosing $b = L$). Given any fixed vector $u$, we have with probability at least $1 - e^{-\Omega(m/L)}$ (over the randomness of $\mathbf{W}_L, \dots, \mathbf{W}_1, \mathbf{A}$),

$$\|\mathbf{D}_L\mathbf{W}_L \cdots \mathbf{D}_a \mathbf{W}_a u\| \le 2\|u\| \ .$$

Conditioning on this event happens, using the randomness of $\mathbf{B}$ (recall each entry of $\mathbf{B}$ follows from $\mathcal{N}(0, \frac{1}{d})$), we have for each fixed vector $u \in \mathbb{R}^m$,

$$\Pr_{\mathbf{B}}\left[\left|v^\top \mathbf{B}\big(\mathbf{D}_L\mathbf{W}_L \cdots \mathbf{D}_a \mathbf{W}_a u\big)\right| \ge \frac{\sqrt{s \log m}}{\sqrt{d}} \cdot O(\|u\|\|v\|)\right] \le e^{-\Omega(s \log m)} \ .$$

Finally, one can take $\varepsilon$-net over all $s$-sparse vectors $u \in \mathbb{R}^m$ and all vectors $v \in \mathbb{R}^d$ and apply union bound.

(b) The proof is identical to Lemma 7.3c, except the fact that each entry of $\mathbf{B}$ follows from $\mathcal{N}(0, \frac{1}{d})$ instead of $\mathcal{N}(0, \frac{2}{m})$.

□

## 7.4 $\delta$-Separateness

**Lemma 7.5** ($\delta$-separateness). *Let $m \ge \Omega\big(\frac{L \log(nL)}{\delta^6}\big)$. There exists some constant $C > 1$ so that, if $\delta \le \frac{1}{CL}$, $\|x_1\| = \cdots = \|x_n\| = 1$ and $\|x_i - x_j\| \ge \delta$ for every pair $i, j \in [n]$, then with probability at least $1 - e^{-\Omega(\delta^6 m/L)}$, we have:*

$$\forall i \ne j \in [n], \quad \forall \ell \in \{0, 1, \dots, L\}\colon \|(\mathbf{I} - \frac{h_{i,\ell}h_{i,\ell}^\top}{\|h_{i,\ell}\|^2})h_{j,\ell}\| \ge \frac{\delta}{2}.$$

*Proof of Lemma 7.5.* We first apply Lemma 7.1 to show that $\|h_{i,\ell}\| \in [1 - \delta^3/10, 1 + \delta^3/10]$. Next we prove Lemma 7.5 by induction.

In the base case of $\ell = -1$, since $\|x_i - x_j\| \geq \delta$ by our Assumption 2.1 and without loss of generality $\|x_i\| = 1$ and $(x_i)_{\mathfrak{d}} = \frac{1}{\sqrt{2}}$, we already have

$$\|(\mathbf{I} - \frac{h_{i,\ell} h_{i,\ell}^\top}{\|h_{i,\ell}\|^2}) h_{j,\ell}\|^2 = \|(\mathbf{I} - \frac{x_i x_i^\top}{\|x_i\|^2}) x_j\|^2 = \|x_j - x_i \cdot \langle x_i, x_j \rangle\|^2 = 1 - \big(\langle x_i, x_j \rangle\big)^2 \geq \frac{3}{4}\delta^2 \ .$$

Suppose $h_{i,\ell-1}$ and $h_{j,\ell-1}$ are fixed and satisfies $\|(\mathbf{I} - \frac{h_{i,\ell-1} h_{i,\ell-1}^\top}{\|h_{i,\ell-1}\|^2}) h_{j,\ell-1}\|^2 \geq \delta_{\ell-1}^2$ for some $\delta_{\ell-1} \geq \delta/2$. We write $\mathbf{W}_\ell h_{i,\ell-1} = \vec{g}_1$ where $\vec{g}_1 \sim N(0, \frac{2\|h_{i,\ell-1}\|^2}{m}\mathbf{I})$.

Denoting by $\widehat{h} = h_{i,\ell-1}/\|h_{i,\ell-1}\|$, we can write $\mathbf{W}_\ell h_{j,\ell-1} = \mathbf{W}_\ell \widehat{h}\widehat{h}^\top h_{j,\ell-1} + \mathbf{W}_\ell\big(\mathbf{I} - \widehat{h}\widehat{h}^\top\big) h_{j,\ell-1}$ and the randomness of the two terms are independent. In particular, we can write

$$\mathbf{W}_\ell h_{j,\ell-1} = \frac{\langle h_{i,\ell-1}, h_{j,\ell-1} \rangle}{\|h_{i,\ell-1}\|^2} \cdot \vec{g}_1 + \|\big(\mathbf{I} - \widehat{h}\widehat{h}^\top\big) h_{j,\ell-1}\| \cdot \vec{g}_2 \tag{7.5}$$

where $\vec{g}_2 \sim \mathcal{N}(0, \frac{2}{m}\mathbf{I})$ is independent of $g_1$. Applying Claim 7.6 for each coordinate $k \in [m]$ (and re-scaling by $\frac{m}{\|h_{i,\ell-1}\|^2}$, we have

$$\mathbb{E}[(\phi(\mathbf{W}_\ell h_{i,\ell-1}) - \phi(\mathbf{W}_\ell h_{j,\ell-1}))_k^2] \geq \left(\frac{\delta_{\ell-1}}{\|h_{i,\ell-1}\|}\right)^2 \left(1 - \frac{\delta_{\ell-1}}{\|h_{i,\ell-1}\|}\right) \cdot \frac{\|h_{i,\ell-1}\|^2}{m} \geq \frac{\delta_{\ell-1}^2 (1 - O(\delta_{\ell-1}))}{m}$$

Applying Chernoff bound (on independent subgaussian random variables), we have with probability at least $1 - e^{-\Omega(\delta_{\ell-1}^4 m)}$,[17]

$$\|h_{i,\ell} - h_{j,\ell}\|^2 = \|\phi(\mathbf{W}_\ell h_{i,\ell-1}) - \phi(\mathbf{W}_\ell h_{j,\ell-1})\|^2 \geq \delta_{\ell-1}^2 (1 - O(\delta_{\ell-1})) \ .$$

Since $\|h_{i,\ell}\|$ and $\|h_{j,\ell}\|$ are close to 1, we have

$$\begin{aligned}\left\|\left(\mathbf{I} - \frac{h_{i,\ell} h_{i,\ell}^\top}{\|h_{i,\ell}\|^2}\right) h_{j,\ell}\right\|^2 &= \|h_{j,\ell}\|^2 - \frac{\langle h_{i,\ell}, h_{j,\ell} \rangle^2}{\|h_{i,\ell}\|^2} \\ &= \|h_{j,\ell}\|^2 + \frac{\|h_{i,\ell} - h_{j,\ell}\|^2 - \|h_{i,\ell}\|^2 - \|h_{j,\ell}\|^2}{2\|h_{i,\ell}\|^2} \geq \delta_{\ell-1}^2 (1 - O(\delta_{\ell-1})) \ . \qquad \square\end{aligned}$$

### 7.4.1 Auxiliary Claim

The following mathematical fact is needed in the proof of Lemma 7.5. Its proof is by carefully integrating the PDF of Gaussian distribution.

**Claim 7.6.** *Given $g_1, g_2 \sim \mathcal{N}(0, 2)$, constant $\alpha \in \mathbb{R}$ and $\delta \in [0, \frac{1}{6}]$, we have*

$$\mathop{\mathbb{E}}_{g_1, g_2}\left[(\phi(g_1) - \phi(\alpha g_1 + \delta g_2))^2\right] \geq \delta^2(1 - \delta) \ .$$

*Proof of Claim 7.6.* We first tackle two easy cases.

Suppose $a < \frac{3}{4}$. If so, then with probability at least 0.3 we have $g_1 > 1$. If this happens, then with probability at least $1/2$ we have $g_2 < 0$. If both happens, we have

$$\phi(g_1) - \phi(\alpha g_1 + \delta g_2) = g_1 - \phi(\alpha g_1 + \delta g_2) \geq g_1 - \alpha g_1 \geq \frac{1}{4} \ .$$

Therefore, we have if $a < \frac{3}{4}$ then the expectation is at least 0.03. For similar reason, if $a > \frac{5}{4}$ we also have the expectation is at least 0.03. In the remainder of the proof, we assume $\alpha \in \big[\frac{3}{4}, \frac{5}{4}\big]$.

[17] More specifically, we can let $X_k = m\,(\phi(\mathbf{W}_\ell h_{i,\ell-1}) - \phi(\mathbf{W}_\ell h_{j,\ell-1}))_k^2$ which is $O(1)$-subgaussian and let $X = X_1 + \cdots + X_m$. We have $\mathbf{Pr}[X \geq \mathbb{E}[X](1 - \delta_{\ell-1})] \geq 1 - e^{-\Omega(\delta_{\ell-1}^2 \mathbb{E}[X])}$.

If $g_1 \geq 0$, we have

$$\begin{aligned}
f(g_1) &\stackrel{\text{def}}{=} \mathop{\mathbb{E}}_{g_2}\left[(\phi(g_1) - \phi(\alpha g_1 + \delta g_2))^2 \,|\, g_1 \geq 0\right] \\
&= \int_0^\infty \frac{(x-g_1)^2 \exp\left(-\frac{(x-\alpha g_1)^2}{4\delta^2}\right)}{\sqrt{4\pi\delta^2}} \mathrm{d}x \\
&= \frac{(\alpha-2)\delta g_1 e^{-\frac{\alpha^2 g_1^2}{4\delta^2}}}{\sqrt{\pi}} + \frac{1}{2}\left((\alpha-1)^2 g_1^2 + 2\delta^2\right)\left(\operatorname{erf}\left(\frac{\alpha g_1}{2\delta}\right)+1\right).
\end{aligned}$$

If $g_1 < 0$, we have

$$\begin{aligned}
f(g_1) &\stackrel{\text{def}}{=} \mathop{\mathbb{E}}_{g_2}\left[(\phi(g_1) - \phi(\alpha g_1 + \delta g_2))^2 \,|\, g_1 < 0\right] \\
&= \int_0^\infty \frac{x^2 \exp\left(-\frac{(x-\alpha g_1)^2}{4\delta^2}\right)}{\sqrt{4\pi\delta^2}} \mathrm{d}x \\
&= \frac{1}{2}\left(\alpha^2 g_1^2 + 2\delta^2\right)\left(\operatorname{erf}\left(\frac{\alpha g_1}{2\delta}\right)+1\right) + \frac{\alpha\delta g_1 e^{-\frac{\alpha^2 g_1^2}{4\delta^2}}}{\sqrt{\pi}}.
\end{aligned}$$

Overall, we have

$$\begin{aligned}
&\mathop{\mathbb{E}}_{g_1,g_2}\left[(\phi(g_1) - \phi(\alpha g_1 + \delta g_2))^2\right] \\
&= \int_0^\infty \frac{f(g)\exp\left(-\frac{g^2}{4}\right)}{\sqrt{4\pi}} \mathrm{d}g + \int_{-\infty}^0 \frac{f(g)\exp\left(-\frac{g^2}{4}\right)}{\sqrt{4\pi}} \mathrm{d}g \\
&= \left(\frac{(\alpha-1)^2\alpha\delta}{\pi(\alpha^2+\delta^2)} + \frac{(\alpha-2)\delta^3}{\pi(\alpha^2+\delta^2)} + \frac{1}{2}\left((\alpha-1)^2+\delta^2\right) + \frac{1}{\pi}\left((\alpha-1)^2+\delta^2\right)\arctan\left(\frac{\alpha}{\delta}\right)\right) \\
&\quad + \frac{1}{2\pi}\left(\pi\left(\alpha^2+\delta^2\right) - 2\left(\alpha^2+\delta^2\right)\arctan\left(\frac{\alpha}{\delta}\right) - 2\alpha\delta\right) \\
&= \frac{\delta\left(-2\alpha^2+\alpha-2\delta^2\right)}{\pi(\alpha^2+\delta^2)} + \frac{(1-2\alpha)\arctan\left(\frac{\alpha}{\delta}\right)}{\pi} + (\alpha-1)\alpha + \delta^2 + \frac{1}{2} \\
&= \left(\alpha^2 - 2\alpha + 1\right) + \delta^2 + \frac{2}{\pi}\sum_{k=1}^{\infty}(-1)^k \frac{(\alpha+k)\delta^{2k+1}}{(2k+1)\alpha^{2k+1}}.
\end{aligned}$$

It is easy to see that, as long as $\delta \leq \alpha$, we always have $\frac{(\alpha+k)\delta^{2k+1}}{(2k+1)\alpha^{2k+1}} \geq \frac{(\alpha+k+1)\delta^{2k+3}}{(2k+3)\alpha^{2k+3}}$. Therefore

$$\mathop{\mathbb{E}}_{g_1,g_2}\left[(\phi(g_1) - \phi(\alpha g_1 + \delta g_2))^2\right] \geq \left(\alpha^2 - 2\alpha + 1\right) + \delta^2 - \frac{2}{\pi}\frac{(\alpha+1)\delta^3}{3\alpha^3} \geq \delta^2(1-\delta) \ . \qquad \square$$

# 8 Stability against Adversarial Weight Perturbations

Let $\mathbf{A}$, $\mathbf{B}$ and $\overrightarrow{\mathbf{W}}^{(0)} = (\mathbf{W}_1^{(0)}, \dots, \mathbf{W}_L^{(0)})$ be matrices at random initialization (see Def. 2.3), and throughout this section, we consider (adversarially) perturbing $\overrightarrow{\mathbf{W}}$ by $\overrightarrow{\mathbf{W}}' = (\mathbf{W}'_1, \dots, \mathbf{W}'_L)$ satisfying $\|\overrightarrow{\mathbf{W}}'\|_2 \leq \omega$ (meaning, $\|\mathbf{W}'_\ell\|_2 \leq \omega$ for every $\ell \in [L]$). We stick to the following notations in this section

**Definition 8.1.**

$$\begin{aligned}
g_{i,0}^{(0)} &= \mathbf{A}x_i & g_{i,0} &= \mathbf{A}x_i & &\textit{for } i \in [n] \\
h_{i,0}^{(0)} &= \phi(\mathbf{A}x_i) & h_{i,0} &= \phi(\mathbf{A}x_i) & &\textit{for } i \in [n] \\
g_{i,\ell}^{(0)} &= \mathbf{W}_\ell^{(0)} h_{i,\ell-1} & g_{i,\ell} &= (\mathbf{W}_\ell^{(0)} + \mathbf{W}'_\ell) h_{i,\ell-1} & &\textit{for } i \in [n] \textit{ and } \ell \in [L] \\
h_{i,\ell}^{(0)} &= \phi(\mathbf{W}_\ell^{(0)} h_{i,\ell-1}) & h_{i,\ell} &= \phi((\mathbf{W}_\ell^{(0)} + \mathbf{W}'_\ell) h_{i,\ell-1}) & &\textit{for } i \in [n] \textit{ and } \ell \in [L]
\end{aligned}$$

*Define diagonal matrices* $\mathbf{D}_{i,\ell}^{(0)} \in \mathbb{R}^{m\times m}$ *and* $\mathbf{D}_{i,\ell} \in \mathbb{R}^{m\times m}$ *by letting* $(\mathbf{D}_{i,\ell}^{(0)})_{k,k} = \mathbb{1}_{(g_{i,\ell}^{(0)})_k \geq 0}$ *and* $(\mathbf{D}_{i,\ell})_{k,k} = \mathbb{1}_{(g_{i,\ell})_k \geq 0}, \forall k \in [m]$*. Accordingly, we let* $g'_{i,\ell} = g_{i,\ell} - g_{i,\ell}^{(0)}$*,* $h'_{i,\ell} = h_{i,\ell} - h_{i,\ell}^{(0)}$*, and diagonal matrix* $\mathbf{D}'_{i,\ell} = \mathbf{D}_{i,\ell} - \mathbf{D}_{i,\ell}^{(0)}$*.*

## 8.1 Forward Perturbation

**Lemma 8.2** (forward perturbation)**.** *Suppose* $\omega \leq \frac{1}{CL^{9/2}\log^3 m}$ *for some sufficiently large constant* $C > 1$*. With probability at least* $1 - e^{-\Omega(m\omega^{2/3}L)}$*, for every* $\overrightarrow{\mathbf{W}}'$ *satisfying* $\|\overrightarrow{\mathbf{W}}'\|_2 \leq \omega$*,*

*(a)* $g'_{i,\ell}$ *can be written as* $g'_{i,\ell} = g'_{i,\ell,1} + g'_{i,\ell,2}$ *where* $\|g'_{i,\ell,1}\| \leq O(\omega L^{3/2})$ *and* $\|g'_{i,\ell,2}\|_\infty \leq O\left(\frac{\omega L^{5/2}\sqrt{\log m}}{\sqrt{m}}\right)$

*(b)* $\|\mathbf{D}'_{i,\ell}\|_0 \leq O(m\omega^{2/3}L)$ *and* $\|\mathbf{D}'_{i,\ell} g_{i,\ell}\| \leq O(\omega L^{3/2})$*.*

*(c)* $\|g'_{i,\ell}\|, \|h'_{i,\ell}\| \leq O(\omega L^{5/2}\sqrt{\log m})$*.*

*Proof of Lemma 8.2.* In our proof below, we drop the subscript with respect to $i$ for notational simplicity, and one can always take a union bound over all possible indices $i$ at the end.

Using Lemma 7.1, we can first assume that $\|h_\ell^{(0)}\|, \|g_\ell^{(0)}\| \in [\frac{2}{3}, \frac{4}{3}]$ for all $\ell$. This happens with probability at least $1 - e^{-\Omega(m/L)}$. We also assume $\left\| \prod_{b=\ell}^{a+1} \mathbf{W}_b^{(0)} \mathbf{D}_{b-1}^{(0)} \right\|_2 \leq c_1\sqrt{L}$ where $c_1 > 0$ is the hidden constant in Lemma 7.3a.

We shall inductively prove Lemma 8.2. In the base case $\ell = 0$, we have $g'_\ell = 0$ so all the statements holds. In the remainder of the proof, we assume that Lemma 8.2 holds for $\ell - 1$ and we shall prove the three statements for layer $\ell$. To help the readers understand how the constants propagate without blowing up, we shall prove $\|g'_{i,\ell,1}\| \leq 4c_1 L^{1.5}\omega$ in Lemma 8.2a without the big-$O$ notation, while for all other terms we use big-$O$ to hide polynomial dependency on $c_1$.[18]

We first carefully rewrite:

$$\begin{aligned}
g'_\ell &= (\mathbf{W}_\ell^{(0)} + \mathbf{W}'_\ell)(\mathbf{D}_{\ell-1}^{(0)} + \mathbf{D}'_{\ell-1})(g_{\ell-1}^{(0)} + g'_{\ell-1}) - \mathbf{W}_\ell^{(0)} \mathbf{D}_{\ell-1}^{(0)} g_{\ell-1}^{(0)} \\
&= \mathbf{W}'_\ell(\mathbf{D}_{\ell-1}^{(0)} + \mathbf{D}'_{\ell-1})(g_{\ell-1}^{(0)} + g'_{\ell-1}) + \mathbf{W}_\ell^{(0)} \mathbf{D}'_{\ell-1}(g_{\ell-1}^{(0)} + g'_{\ell-1}) + \mathbf{W}_\ell^{(0)} \mathbf{D}_{\ell-1}^{(0)} g'_{\ell-1} \\
&= \cdots \\
&= \sum_{a=1}^{\ell} \left( \prod_{b=\ell}^{a+1} \mathbf{W}_b^{(0)} \mathbf{D}_{b-1}^{(0)} \right) \Big( \underbrace{\mathbf{W}'_a(\mathbf{D}_{a-1}^{(0)} + \mathbf{D}'_{a-1})(g_{a-1}^{(0)} + g'_{a-1})}_{(\diamondsuit)} + \underbrace{\mathbf{W}_a^{(0)} \mathbf{D}'_{a-1}(g_{a-1}^{(0)} + g'_{a-1})}_{(\heartsuit)} \Big)
\end{aligned}$$

[18] Alternatively, one can fully specify all the constants without using the big-$O$ notation. This was done in our prior work [6] but is notation-heavy. We refrain from doing so in this simplified paper.

For each term in ($\diamondsuit$), we have

$$\begin{aligned}
&\Big\|\Big(\prod_{b=\ell}^{a+1}\mathbf{W}_b^{(0)}\mathbf{D}_{b-1}^{(0)}\Big)\Big(\mathbf{W}'_a(\mathbf{D}_{a-1}^{(0)}+\mathbf{D}'_{a-1})(g_{a-1}^{(0)}+g'_{a-1})\Big)\Big\| \\
&\leq \Big\|\prod_{b=\ell}^{a+1}\mathbf{W}_b^{(0)}\mathbf{D}_{b-1}^{(0)}\Big\|_2\cdot\Big\|\mathbf{W}'_a\Big\|_2\cdot\Big\|\mathbf{D}_{a-1}^{(0)}+\mathbf{D}'_{a-1}\Big\|_2\cdot\Big\|g_{a-1}^{(0)}+g'_{a-1}\Big\| \\
&\overset{①}{\leq} c_1\cdot\omega\cdot 1\cdot\Big\|g_{a-1}^{(0)}+g'_{a-1}\Big\| \overset{②}{\leq} 2c_1\sqrt{L}\omega+O\big(\omega^2L^3\sqrt{\log m}\big) \enspace .
\end{aligned}$$

Above, inequality ① uses Lemma 7.3a and $\|\mathbf{D}_{a-1}^{(0)}+\mathbf{D}'_{a-1}\|_2=\|\mathbf{D}_{a-1}\|_2\leq 1$; and inequality ② has used $\|g_\ell^{(0)}\|\leq 2$ and our inductive assumption Lemma 8.2c. By triangle inequality, we have

$$g'_\ell=\overrightarrow{\mathsf{err}}_1+\sum_{a=1}^{\ell}\Big(\prod_{b=\ell}^{a+1}\mathbf{W}_b^{(0)}\mathbf{D}_{b-1}^{(0)}\Big)\Big(\underbrace{\mathbf{W}_a^{(0)}\mathbf{D}'_{a-1}(g_{a-1}^{(0)}+g'_{a-1})}_{(\heartsuit)}\Big)$$

where $\|\overrightarrow{\mathsf{err}}_1\|\leq 2c_1L^{1.5}\omega+O\big(\omega^2L^4\sqrt{\log m}\big)$. We next look at each term in ($\heartsuit$). For each $a=2,3,\dots,\ell$, we let

$$x\overset{\text{def}}{=}\mathbf{D}'_{a-1}(g_{a-1}^{(0)}+g'_{a-1})=\mathbf{D}'_{a-1}(\mathbf{W}_{a-1}^{(0)}h_{a-1}^{(0)}+g'_{a-1}) \enspace .$$

If we re-scale $x$ by $\frac{1}{\|h_{a-1}^{(0)}\|}$ (which is a constant in $[0.75,1.5]$), we can apply Claim 8.3 (with parameter choices in Corollary 8.4) on $x$ and this tells us, with probability at least $1-e^{-\Omega(m\omega^{2/3}L)}$:

$$\|x\|_0\leq O(m\omega^{2/3}L) \quad\text{and}\quad \|x\|\leq O(\omega L^{3/2}). \tag{8.1}$$

Next, each term in ($\heartsuit$) contributes to $g'_\ell$ by

$$y=\Big(\prod_{b=\ell}^{a+1}\mathbf{W}_b^{(0)}\mathbf{D}_{b-1}^{(0)}\Big)\mathbf{W}_a^{(0)}\Big(\mathbf{D}'_{a-1}(g_{a-1}^{(0)}+g'_{a-1})\Big)$$

using (8.1) and Claim 8.5 (with $s=O(m\omega^{2/3}L)$), we have with probability at least $1-e^{-\Omega(s\log m)}$, one can write $y=y_1+y_2$ for

$$\|y_1\|\leq O\big(\omega L^{3/2}\cdot L^{1/2}\omega^{1/3}\log m\big) \quad\text{and}\quad \|y_2\|_\infty\leq O\Big(\omega L^{3/2}\cdot\frac{\sqrt{\log m}}{\sqrt{m}}\Big) \enspace .$$

And therefore by triangle inequality we can write

$$g'_\ell=\overrightarrow{\mathsf{err}}_1+\overrightarrow{\mathsf{err}}_2+\overrightarrow{\mathsf{err}}_3$$

where $\|\overrightarrow{\mathsf{err}}_2\|\leq O\big(L\cdot\omega L^{3/2}\cdot L^{1/2}\omega^{1/3}\log m\big)=O\big(\omega^{4/3}L^3\log m\big)$ and $\|\overrightarrow{\mathsf{err}}_3\|_\infty\leq O\Big(L\cdot\omega L^{3/2}\cdot\frac{\sqrt{\log m}}{\sqrt{m}}\Big)$. Together with the upper bound on $\overrightarrow{\mathsf{err}}_1$, we have

$$\|\overrightarrow{\mathsf{err}}_1+\overrightarrow{\mathsf{err}}_2\|\leq 2c_1L^{1.5}\omega+O\big(\omega^2L^4\sqrt{\log m}+\omega^{4/3}L^3\log m\big) \enspace .$$

We emphasize that the above big-$O$ notion can hide polynomial dependency on $c_1$. Nevertheless, when $\omega$ is sufficiently small, the above term is at most $4c_1L^{1.5}\omega$. This finishes the proof of Lemma 8.2a for layer $\ell$ without blowing up the constant. Finally,

- Lemma 8.2b is due to (8.1),
- $g'_\ell$ part of Lemma 8.2c is a simple corollary of Lemma 8.2a, and
- $h'_\ell$ part of Lemma 8.2c is due to $h'_\ell=\mathbf{D}_\ell g'_\ell+\mathbf{D}'_\ell g_\ell$ together with the bound on $\|g'_\ell\|$ and the bound on $\mathbf{D}'_\ell g_\ell$ from Lemma 8.2b. □

### 8.1.1 Auxiliary Claim

**Claim 8.3.** *Suppose $\delta_2 \in [0, O(1)]$ and $\delta_\infty \in [0, \frac{1}{4\sqrt{m}}]$. Suppose $\mathbf{W}^{(0)} \in \mathbb{R}^{m\times m}$ is a random matrix with entries drawn i.i.d. from $\mathcal{N}\big(0, \frac{2}{m}\big)$. With probability at least $1 - e^{-\Omega(m^{3/2}\delta_\infty)}$, the following holds. Fix any unit vector $h^{(0)} \in \mathbb{R}^m$, and for all $g' \in \mathbb{R}^m$ that can be written as*

$$g' = g'_1 + g'_2 \text{ where } \|g'_1\| \le \delta_2 \text{ and } \|g'_2\|_\infty \le \delta_\infty .$$

*Let $\mathbf{D}' \in \mathbb{R}^{m\times m}$ be the diagonal matrix where $(\mathbf{D}')_{k,k} = \mathbb{1}_{(\mathbf{W}^{(0)}h^{(0)}+g')_k \ge 0} - \mathbb{1}_{(\mathbf{W}^{(0)}h^{(0)})_k \ge 0}, \forall k \in [m]$. Then, letting $x = \mathbf{D}'(\mathbf{W}^{(0)}h^{(0)} + g') \in \mathbb{R}^m$, we have*

$$\|x\|_0 \le \|\mathbf{D}'\|_0 \le O(m(\delta_2)^{2/3} + \delta_\infty m^{3/2}) \quad \text{and} \quad \|x\| \le O(\delta_2 + (\delta_\infty)^{3/2} m^{3/4}) \ .$$

**Corollary 8.4.** *In particular, if $\omega L^{3/2} \le O(1)$, then with probability at least $1 - e^{-\Omega(m\omega^{2/3}L)}$, for every $g' = g'_1 + g'_2$ with $\|g'_1\| \le O(\omega L^{3/2})$ and $\|g'_2\|_\infty \le O\big(\frac{\omega^{2/3}L}{m^{1/2}}\big)$, it satisfies*

$$\|\mathbf{D}'\|_0 \le O(m\omega^{2/3}L) \quad \text{and} \quad \|x\| \le O(\omega L^{3/2}) \ .$$

*Proof of Claim 8.3.* We first observe $g^{(0)} = \mathbf{W}^{(0)}h^{(0)}$ follows from $\mathcal{N}\big(0, \frac{2\mathbf{I}}{m}\big)$ regardless of the choice of $h^{(0)}$. Therefore, in the remainder of the proof, we just focus on the randomness of $g^{(0)}$.

We also observe that $(\mathbf{D}')_{j,j}$ is non-zero for some diagonal $j \in [m]$ only if

$$|(g'_1 + g'_2)_j| > |(g^{(0)})_j| \ . \tag{8.2}$$

Let $\xi \le \frac{1}{2\sqrt{m}}$ be a parameter to be chosen later. We shall make sure that $\|g'_2\|_\infty \le \xi/2$.

- We denote by $S_1 \subseteq [m]$ the index sets where $j$ satisfies $|(g^{(0)})_j| \le \xi$. Since we know $(g^{(0)})_j \sim \mathcal{N}(0, 2/m)$, we have $\mathbf{Pr}[|(g^{(0)})_j| \le \xi] \le O(\xi\sqrt{m})$ for each $j \in [m]$. Using Chernoff bound for all $j \in [m]$, we have with probability at least $1 - e^{-\Omega(m^{3/2}\xi)}$,

  $$|S_1| = \left|\left\{i \in [m] \colon |(g^{(0)})_j| \le \xi\right\}\right| \le O(\xi m^{3/2}) \ .$$

  Now, for each $j \in S_1$ such that $x_j \neq 0$, we must have $|x_j| = |(g^{(0)} + g'_1 + g'_2)_j| \le |(g'_1)_j| + 2\xi$ so we can calculate the $\ell_2$ norm of $x$ on $S_1$:

  $$\sum_{i \in S_1} x_j^2 \le O(\|g'_1\|^2 + \xi^2|S_1|) \le O(\|g'_1\|^2 + \xi^3 m^{3/2}) \ .$$

- We denote by $S_2 \subseteq [m] \setminus S_1$ the index set of all $j \in [m] \setminus S_1$ where $x_j \neq 0$. Using (8.2), we have for each $j \in S_2$:

  $$|(g'_1)_j| \ge |(g^{(0)})_j| - |(g'_2)_j| \ge \xi - \|g'_2\|_\infty \ge \xi/2 \ .$$

  This means

  $$|S_2| \le \frac{4\|g'_1\|^2}{\xi^2} \ .$$

  Now, for each $j \in S_2$ where $x_j \neq 0$, we know that the signs of $(g^{(0)} + g'_1 + g'_2)_j$ and $(g^{(0)})_j$ are opposite. Therefore, we must have

  $$|x_j| = |(g^{(0)} + g'_1 + g'_2)_j| \le |(g'_1 + g'_2)_j| \le |(g'_1)_j| + \xi/2 \le 2|(g'_1)_j|$$

  and therefore

  $$\sum_{j \in S_2} x_j^2 \le 4 \sum_{j \in S_2} (g'_1)_j^2 \le 4\|g'_1\|^2 \ .$$

From above, we have $\|x\|_0 \leq |S_1|+|S_2| \leq O\big(\xi m^{3/2}+\frac{(\delta_2)^2}{\xi^2}\big)$ and $\|x\|^2 \leq O\big((\delta_2)^2+\xi^3 m^{3/2}\big)$. Choosing $\xi = \max\{2\delta_\infty, \Theta(\frac{(\delta_2)^{2/3}}{m^{1/2}})\}$ for the former, and choosing $\xi = 2\delta_\infty$ for the latter, we have the desired result. □

**Claim 8.5.** *For any $2 \leq a \leq b \leq L$ and any positive integer $s \leq O\big(\frac{m}{L\log m}\big)$, with probability at least $1-e^{-\Omega(s\log m)}$, for all $x \in \mathbb{R}^m$ with $\|x\| \leq 1$ and $\|x\|_0 \leq s$, letting $y = \mathbf{W}_b^{(0)}\mathbf{D}_{b-1}^{(0)}\mathbf{W}_{b-1}^{(0)}\cdots\mathbf{D}_a^{(0)}\mathbf{W}_a^{(0)}x$, we can write $y = y_1 + y_2$ with*

$$\|y_1\| \leq O\big(\sqrt{s/m}\log m\big) \quad \textit{and} \quad \|y_2\|_\infty \leq \frac{2\sqrt{\log m}}{\sqrt{m}} \enspace .$$

*Proof of Claim 8.5.* First of all, fix any $x$, we can let $u = \mathbf{D}_{b-1}^{(0)}\mathbf{W}_{b-1}^{(0)}\cdots\mathbf{D}_a^{(0)}\mathbf{W}_a^{(0)}x$ and the same proof of Lemma 7.3 implies that with probability at least $1-e^{-\Omega(m/L)}$ we have $\|u\| \leq O(\|x\|)$. We next condition on this event happens.

Let $\beta = \sqrt{\log m}/\sqrt{m}$. If $u$ is fixed and using only the randomness of $\mathbf{W}_b$, we have $y_i \sim \mathcal{N}\big(0, \frac{2\|u\|^2}{m}\big)$ so for every $p \geq 1$, by Gaussian tail bound

$$\mathbf{Pr}[|y_i| \geq \beta p] \leq e^{-\Omega(\beta^2 p^2 m/\|x\|^2)} \leq e^{-\Omega(\beta^2 p^2 m)} \enspace .$$

As long as $\beta^2 p^2 m \geq \beta^2 m \geq \Omega(\log m)$, we know that if $|y_i| \geq \beta p$ occurs for $q/p^2$ indices $i$ out of $[m]$, this cannot happen with probability more than

$$\binom{m}{q/p^2} \times \Big(e^{-\Omega(\beta^2 p^2 m)}\Big)^{q/p^2} \leq e^{\frac{q}{p^2}\big(O(\log m)-\Omega(\beta^2 p^2 m)\big)} \leq e^{-\Omega(\beta^2 q m)} \enspace .$$

In other words,

$$\mathbf{Pr}\left[|\{i \in [m] \colon |y_i| \geq \beta p\}| > q/p^2\right] \leq e^{-\Omega(\beta^2 q m)} \enspace .$$

Finally, by applying union bound over $p = 1, 2, 4, 8, 16, \dots$ we have with probability $\geq 1-e^{-\Omega(\beta^2 q m)} \cdot \log q$,

$$\sum_{i \colon |y_i| \geq \beta} y_i^2 \leq \sum_{k=0}^{\lceil \log q \rceil} (2^{k+1}\beta)^2 \left|\left\{i \in [m] \colon |y_i| \geq 2^k \beta\right\}\right| \leq \sum_{k=0}^{\lceil \log q \rceil} (2^{k+1}\beta)^2 \cdot \frac{q}{2^{2k}} \leq O(q\beta^2 \log q) \tag{8.3}$$

In other words, vector $y$ can be written as $y = y_1 + y_2$ where $\|y_2\|_\infty \leq \beta$ and $\|y_1\|^2 \leq O(q\beta^2 \log q)$.

Finally, we want to take $\varepsilon$-net over all $s$-sparse inputs $x$. This requires $\beta^2 q m \geq \Omega(s \log m)$, so we can choose $q = \Theta\big(\frac{s\log m}{m\beta^2}\big) = \Theta(s)$. □

## 8.2 Intermediate Layers

**Lemma 8.6** (intermediate perturbation)**.** *For any integer $s$ with $1 \leq s \leq O\big(\frac{m}{L^3\log m}\big)$, with probability at least $1-e^{-\Omega(s\log m)}$ over the randomness of $\overrightarrow{\mathbf{W}}^{(0)}, \mathbf{A}$,*

- *for every $i \in [n], 1 \leq a \leq b \leq L$,*
- *for every diagonal matrices $\mathbf{D}''_{i,0}, \dots, \mathbf{D}''_{i,L} \in [-3,3]^{m\times m}$ with at most $s$ non-zero entries.*
- *for every perturbation matrices $\mathbf{W}'_1, \dots, \mathbf{W}'_L \in \mathbb{R}^{m\times m}$ with $\|\overrightarrow{\mathbf{W}}'\|_2 \leq \omega \in [0,1]$.*

*we have*

*(a)* $\|\mathbf{W}_b^{(0)}(\mathbf{D}_{i,b-1}^{(0)} + \mathbf{D}''_{i,b-1})\cdots(\mathbf{D}_{i,a}^{(0)} + \mathbf{D}''_{i,a})\mathbf{W}_a^{(0)}\|_2 \leq O(\sqrt{L})$.

*(b)* $\|(\mathbf{W}_b^{(0)} + \mathbf{W}'_b)(\mathbf{D}_{i,b-1}^{(0)} + \mathbf{D}''_{i,b-1})\cdots(\mathbf{D}_{i,a}^{(0)} + \mathbf{D}''_{i,a})(\mathbf{W}_a^{(0)} + \mathbf{W}'_a)\|_2 \leq O(\sqrt{L})$ *if* $\omega \leq O(\frac{1}{L^{1.5}})$.

*Proof.* For notational simplicity we ignore subscripts in $i$ in the proofs.

(a) Note that each $\mathbf{D}''_\ell$ can be written as $\mathbf{D}''_\ell = \mathbf{D}^{0/1}_\ell \mathbf{D}''_\ell \mathbf{D}^{0/1}_\ell$, where each $\mathbf{D}^{0/1}_\ell$ is a diagonal matrix satisfying

$$(\mathbf{D}^{0/1}_\ell)_{k,k} = \begin{cases} 1, & (\mathbf{D}''_\ell)_{k,k} \neq 0; \\ 0, & (\mathbf{D}''_\ell)_{k,k} = 0. \end{cases} \quad \text{and} \quad \|\mathbf{D}^{0/1}_\ell\|_0 \leq s \enspace .$$

In order to bound the spectral norm of $\mathbf{W}^{(0)}_b(\mathbf{D}^{(0)}_{b-1}+\mathbf{D}''_{b-1})\mathbf{W}^{(0)}_{b-1}\cdots(\mathbf{D}^{(0)}_a+\mathbf{D}''_a)\mathbf{W}^{(0)}_a$, by triangle inequality, we can expend it into $2^{b-a}$ matrices and bound their spectral norms individually. Each such matrix can be written as (ignoring the subscripts)

$$(\mathbf{W}^{(0)}\mathbf{D}^{(0)}\cdots\mathbf{W}^{(0)}\mathbf{D}^{0/1})\mathbf{D}''(\mathbf{D}^{0/1}\mathbf{W}^{(0)}\mathbf{D}^{(0)}\cdots\mathbf{W}^{(0)}\mathbf{D}^{0/1})\mathbf{D}''\cdots\mathbf{D}''(\mathbf{D}^{0/1}\mathbf{W}^{(0)}\mathbf{D}^{(0)}\cdots\mathbf{W}^{(0)}) \tag{8.4}$$

Therefore, it suffices for us to bound the spectral norm of the following four types of matrices:

- $\mathbf{W}^{(0)}\mathbf{D}^{(0)}\cdots\mathbf{W}^{(0)}\mathbf{D}^{0/1}$, such matrix has spectral norm at most 2 owing to Lemma 7.3b;
- $\mathbf{D}^{0/1}\mathbf{W}^{(0)}\mathbf{D}^{(0)}\cdots\mathbf{W}^{(0)}$, such matrix has spectral norm at most $O(1)$ owing to Lemma 7.3c;
- $\mathbf{D}^{0/1}\mathbf{W}^{(0)}\mathbf{D}^{(0)}\cdots\mathbf{W}^{(0)}\mathbf{D}^{0/1}$, such matrix has spectral norm at most $\frac{1}{100L^{1.5}}$ owing to Lemma 7.3d and our choice $s \leq O(\frac{m}{L^3\log m})$;
- $\mathbf{D}''$, such matrix has spectral norm at most 3.

Together, we have

$$\begin{aligned}&\left\|\mathbf{W}^{(0)}_b(\mathbf{D}^{(0)}_{b-1}+\mathbf{D}''_{b-1})\mathbf{W}^{(0)}_{b-1}\cdots(\mathbf{D}^{(0)}_a+\mathbf{D}''_a)\mathbf{W}^{(0)}_a\right\| \\ &\leq O(\sqrt{L}) + \sum_{j=1}^{b-a}\binom{b-a}{j}\cdot O(1)\cdot\left(\frac{1}{100L^{1.5}}\right)^{j-1}\cdot 3^j\cdot O(1) \leq O(\sqrt{L}) \enspace .\end{aligned}$$

(b) In order to bound the spectral norm of $(\mathbf{W}^{(0)}_b+\mathbf{W}'_b)(\mathbf{D}^{(0)}_{b-1}+\mathbf{D}''_{b-1})\cdots(\mathbf{D}^{(0)}_a+\mathbf{D}''_a)(\mathbf{W}^{(0)}_a+\mathbf{W}'_a)$, by triangle inequality, we can expend it into $2^{b-a+1}$ matrices in terms of $\mathbf{W}'$ and bound their spectral norms individually. Each such matrix can be written as (ignoring the subscripts, and denoting $\breve{\mathbf{D}} = \mathbf{D}^{(0)} + \mathbf{D}'$)

$$(\mathbf{W}^{(0)}\breve{\mathbf{D}}\cdots\mathbf{W}^{(0)}\breve{\mathbf{D}})\mathbf{W}'(\breve{\mathbf{D}}\mathbf{W}^{(0)}\cdots\mathbf{W}^{(0)}\breve{\mathbf{D}})\cdots\mathbf{W}'(\breve{\mathbf{D}}\mathbf{W}^{(0)}\cdots\breve{\mathbf{D}}\mathbf{W}^{(0)})$$

Moreover, from Lemma 8.6a, we know the following three types of matrices

- $\mathbf{W}^{(0)}\breve{\mathbf{D}}\cdots\mathbf{W}^{(0)}\breve{\mathbf{D}}$,
- $\breve{\mathbf{D}}\mathbf{W}^{(0)}\cdots\mathbf{W}^{(0)}\breve{\mathbf{D}}$, and
- $\breve{\mathbf{D}}\mathbf{W}^{(0)}\cdots\breve{\mathbf{D}}\mathbf{W}^{(0)}$

all have spectral norm at most $O(\sqrt{L})$. Together, using $\|\mathbf{W}'_\ell\|_2 \leq O(\frac{1}{L^{1.5}})$, we have

$$\begin{aligned}&\left\|(\mathbf{W}^{(0)}_b+\mathbf{W}'_b)(\mathbf{D}^{(0)}_{b-1}+\mathbf{D}''_{b-1})(\mathbf{W}^{(0)}_{b-1}+\mathbf{W}'_{b-1})\cdots(\mathbf{D}^{(0)}_a+\mathbf{D}''_a)(\mathbf{W}^{(0)}_a+\mathbf{W}'_a)\right\| \\ &\leq \sum_{j=0}^{b-a+1}\binom{b-a+1}{j}\cdot\Big(O(\sqrt{L})\Big)^{j+1}\cdot\Big(O(\frac{1}{L^{1.5}})\Big)^{j} \leq O(\sqrt{L}) \enspace .\end{aligned}$$

□

### 8.3 Backward

**Lemma 8.7** (backward perturbation)**.** *For any integer* $s \in \big[\Omega(\frac{d}{\log m}), O(\frac{m}{L^3\log m})\big]$, *for* $d \leq O(\frac{m}{L\log m})$, *with probability at least* $1 - e^{-\Omega(s\log m)}$ *over the randomness of* $\overrightarrow{\mathbf{W}}^{(0)}, \mathbf{A}, \mathbf{B}$,

- *for all $i \in [n]$, $a = 1, 2, \dots, L+1$,*
- *for every diagonal matrices $\mathbf{D}''_{i,0}, \dots, \mathbf{D}''_{i,L} \in [-3,3]^{m \times m}$ with at most $s$ non-zero entries,*
- *for every perturbation matrices $\mathbf{W}'_{i,1}, \dots, \mathbf{W}'_{i,L} \in \mathbb{R}^{m \times m}$ with $\|\overrightarrow{\mathbf{W}}'\|_2 \le \omega = O(\frac{1}{L^{1.5}})$,*

*it satisfies $\|\mathbf{B}(\mathbf{D}^{(0)}_{i,L}+\mathbf{D}''_{i,L})(\mathbf{W}^{(0)}_L+\mathbf{W}'_L)\cdots(\mathbf{W}^{(0)}_{a+1}+\mathbf{W}'_{a+1})(\mathbf{D}^{(0)}_{i,a}+\mathbf{D}''_{i,a}) - \mathbf{B}\mathbf{D}^{(0)}_{i,L}\mathbf{W}^{(0)}_L \cdots \mathbf{W}^{(0)}_{a+1}\mathbf{D}^{(0)}_{i,a}\|_2 \le O\big(\frac{\sqrt{L^3 s \log m + \omega^2 L^3 m}}{\sqrt{d}}\big)$. Note that if $s = O(m\omega^{2/3}L)$, this upper bound becomes $O\big(\frac{\omega^{1/3}L^2\sqrt{m\log m}}{\sqrt{d}}\big)$.*

*Proof.* For notational simplicity we ignore subscripts in $i$ in the proofs.

Ignoring the subscripts for cleanness, we have

$$
\begin{aligned}
&\big\|\mathbf{B}(\mathbf{D}^{(0)}_{i,L}+\mathbf{D}''_{i,L})(\mathbf{W}^{(0)}_L+\mathbf{W}'_L)\cdots(\mathbf{W}^{(0)}_{a+1}+\mathbf{W}'_{a+1})(\mathbf{D}^{(0)}_{i,a}+\mathbf{D}''_{i,a}) - \mathbf{B}\mathbf{D}^{(0)}_{i,L}\mathbf{W}^{(0)}_L \cdots \mathbf{W}^{(0)}_{a+1}\mathbf{D}^{(0)}_{i,a}\big\|_2 \\
&\le \sum_{\ell=a}^{L} \underbrace{\big\|\mathbf{B}\mathbf{D}^{(0)}_{i,L}\mathbf{W}^{(0)}_L\cdots\mathbf{W}^{(0)}_{\ell+1}\mathbf{D}^{0/1}_\ell\big\|_2}_{\text{Lemma 7.4a}} \|\mathbf{D}''_\ell\|_2 \underbrace{\big\|\mathbf{D}^{0/1}_\ell(\mathbf{W}^{(0)}_\ell+\mathbf{W}'_\ell)\cdots(\mathbf{D}^{(0)}_{i,a}+\mathbf{D}''_{i,a})\big\|_2}_{\text{Lemma 8.6b}} \\
&\quad + \sum_{\ell=a+1}^{L} \underbrace{\big\|\mathbf{B}\mathbf{D}^{(0)}_{i,L}\mathbf{W}^{(0)}_L\cdots\mathbf{W}^{(0)}_{\ell+1}\mathbf{D}^{(0)}_\ell\big\|_2}_{\text{Lemma 7.4b}} \|\mathbf{W}'_\ell\|_2 \underbrace{\big\|(\mathbf{D}^{(0)}_{\ell-1}+\mathbf{D}''_{\ell-1})(\mathbf{W}^{(0)}_{\ell-1}+\mathbf{W}'_{\ell-1})\cdots(\mathbf{D}^{(0)}_{i,a}+\mathbf{D}''_{i,a})\big\|_2}_{\text{Lemma 8.6b}} \\
&\le L \cdot O\left(\frac{\sqrt{s\log m}}{\sqrt{d}}\right)\cdot O(\sqrt{L}) + L\cdot O(\sqrt{m/d})\cdot\omega\cdot O(\sqrt{L})
\end{aligned}
$$

□

# 9 Gradient Bound at Random Initialization

Throughout this section we assume $\overrightarrow{\mathbf{W}}, \mathbf{A}$ and $\mathbf{B}$ are randomly generated according to Def. 2.3. The diagonal sign matrices $\mathbf{D}_{i,\ell}$ are also determined according to this random initialization.

Recall we have defined $\mathsf{Back}_{i,\ell} \stackrel{\text{def}}{=} \mathbf{B}\mathbf{D}_{i,L}\mathbf{W}_L\cdots\mathbf{D}_{i,\ell}\mathbf{W}_\ell \in \mathbb{R}^{d\times m}$. In this section, we introduce the following notion

**Definition 9.1.** *For any vector tuple $\vec{\mathsf{v}} = (\mathsf{v}_1, \dots, \mathsf{v}_n) \in (\mathbb{R}^d)^n$ (viewed as a fake loss vector), for each $\ell \in [L]$, we define*

$$
\widehat{\nabla}^{\vec{\mathsf{v}}}_{[\mathbf{W}_\ell]_k} F(\overrightarrow{\mathbf{W}}) \stackrel{\text{def}}{=} \sum_{i=1}^n (\mathsf{Back}^\top_{i,\ell+1}\mathsf{v}_i)_k \cdot h_{i,\ell-1}\cdot \mathbb{1}_{\langle[\mathbf{W}_\ell]_k, h_{i,\ell-1}\rangle\ge 0}, \forall k\in[m]
$$

$$
\widehat{\nabla}^{\vec{\mathsf{v}}}_{\mathbf{W}_\ell} F(\overrightarrow{\mathbf{W}}) \stackrel{\text{def}}{=} \sum_{i=1}^n \widehat{\nabla}^{\vec{\mathsf{v}}}_{\mathbf{W}_\ell} F_i(\overrightarrow{\mathbf{W}}) \quad \textit{where} \quad \widehat{\nabla}^{\vec{\mathsf{v}}}_{\mathbf{W}_\ell} F_i(\overrightarrow{\mathbf{W}}) \stackrel{\text{def}}{=} \mathbf{D}_{i,\ell}(\mathsf{Back}^\top_{i,\ell+1}\mathsf{v}_i)h^\top_{i,\ell-1}
$$

*Remark* 9.2. It is an easy exercise to check that, if letting $\vec{\mathsf{v}} = (\mathsf{v}_1, \dots, \mathsf{v}_n)$ where $\mathsf{v}_i = \mathbf{B}h_{i,L} - y^*_i$, then $\widehat{\nabla}^{\vec{\mathsf{v}}}_{[\mathbf{W}_\ell]_k} F(\overrightarrow{\mathbf{W}}) = \nabla_{[\mathbf{W}_\ell]_k} F(\overrightarrow{\mathbf{W}})$ and $\widehat{\nabla}^{\vec{\mathsf{v}}}_{\mathbf{W}_\ell} F_i(\overrightarrow{\mathbf{W}}) = \nabla_{\mathbf{W}_\ell} F_i(\overrightarrow{\mathbf{W}})$.

Our main lemma of this section is the following.

**Lemma 9.3** (gradient bound at random initialization)**.** *Fix any $\vec{\mathsf{v}} \in (\mathbb{R}^d)^n$, with probability at least $1 - e^{-\Omega(\delta m/n)}$ over the randomness of $\mathbf{A}, \overrightarrow{\mathbf{W}}, \mathbf{B}$, it satisfies for every $\ell \in [L]$:*

$$
\|\widehat{\nabla}^{\vec{\mathsf{v}}}_{\mathbf{W}_\ell} F_i(\overrightarrow{\mathbf{W}})\|_F^2 \le O\Big(\frac{\|\mathsf{v}_i\|^2}{d}\times m\Big) \qquad \|\widehat{\nabla}^{\vec{\mathsf{v}}}_{\mathbf{W}_\ell} F(\overrightarrow{\mathbf{W}})\|_F^2 \le O\Big(\frac{\|\vec{\mathsf{v}}\|^2}{d}\times mn\Big)
$$

$$
\|\widehat{\nabla}^{\vec{\mathsf{v}}}_{\mathbf{W}_L} F(\overrightarrow{\mathbf{W}})\|_F^2 \ge \Omega\Big(\frac{\max_{i\in[n]}\|\mathsf{v}_i\|^2}{dn/\delta}\times m\Big)
$$

## 9.1 Proof of Lemma 9.3: Upper Bound

For each $i \in [n], \ell \in [L]$, we can calculate that

$$\begin{aligned}
\left\|\widehat{\nabla}^{\vec{\mathsf{v}}}_{\mathbf{W}_\ell} F_i(\overrightarrow{\mathbf{W}})\right\|_F &= \left\|\mathbf{D}_{i,\ell}(\mathsf{Back}_{i,\ell+1}^\top \cdot \mathsf{v}_i) \cdot h_{i,\ell-1}^\top\right\|_F \\
&= \left\|\mathbf{D}_{i,\ell}(\mathsf{Back}_{i,\ell+1}^\top \cdot \mathsf{v}_i)\right\|_2 \cdot \|h_{i,\ell-1}\|_2 \\
&\leq \|\mathsf{Back}_{i,\ell+1}\|_2 \cdot \|\mathsf{v}_i\|_2 \cdot \|h_{i,\ell-1}\|_2 \\
&\leq \|\mathbf{B}\mathbf{W}_L \mathbf{D}_{L-1} \cdots \mathbf{D}_{i,\ell+1}\mathbf{W}_{\ell+1}\|_2 \cdot \|\mathsf{v}_i\|_2 \cdot \|h_{i,\ell-1}\|_2 \\
&\overset{①}{\leq} O(\sqrt{m/d}) \cdot O(1) \cdot \|\mathsf{v}_i\|_2 \enspace.
\end{aligned}$$

where inequality ① uses Lemma 7.4b and Lemma 7.1 with high probability. Applying triangle inequality with respect to all $\ell \in [L]$, taking square on both sides, and summing up over all $i \in [n]$ finish the proof.

## 9.2 Proof of Lemma 9.3: Lower Bound

Let $i^* = \arg\max_{i\in[n]}\{\|\mathsf{v}_i\|\}$. Recall

$$\widehat{\nabla}^{\vec{\mathsf{v}}}_{[\mathbf{W}_L]_k} F(\overrightarrow{\mathbf{W}}) = \sum_{i=1}^n \langle \mathbf{B}_k, \mathsf{v}_i\rangle \cdot h_{i,L-1} \cdot \mathbb{1}_{(\mathbf{W}_L h_{i,L-1})_k \geq 0}$$

Let $\widehat{h} \overset{\text{def}}{=} \frac{h_{i^*,L-1}}{\|h_{i^*,L-1}\|}$. For analysis purpose, after $\widehat{h}$ is fixed (so after fixing the randomness of $\mathbf{A}, \mathbf{W}_1, \dots, \mathbf{W}_{L-1}$), we redefine $\mathbf{W}_L\widehat{h} = \sqrt{1-\theta^2}\widehat{g}_1 + \theta\widehat{g}_2$ where $\widehat{g}_1$ and $\widehat{g}_2$ are generated independently from $\mathcal{N}(0, \frac{2\mathbf{I}}{m})$. We can do so because the two sides are equal in distribution. In other words, we can set

$$\mathbf{W}'_L \overset{\text{def}}{=} \mathbf{W}_L(\mathbf{I} - \widehat{h}\widehat{h}^\top) - \sqrt{1-\theta^2}\widehat{g}_1\widehat{h}^\top \text{ and } \mathbf{W}''_L \overset{\text{def}}{=} \theta\widehat{g}_2\widehat{h}^\top,$$

then we have $\mathbf{W}_L = \mathbf{W}'_L + \mathbf{W}''_L$. In particular, the randomness of $\mathbf{W}'_L$ and $\mathbf{W}''_L$ are *independent.*

In the remainder of the proof, let us choose $\theta \overset{\text{def}}{=} \frac{\delta}{5n} \leq \frac{1}{5}$.

We first make two technical claims, and the proof of the first one can be found in Section 9.2.1.

**Claim 9.4.** *We have* $\mathbf{Pr}_{\mathbf{W}'_L,\mathbf{W}_{L-1},\dots,\mathbf{W}_1,\mathbf{A}}\left[|N_2| \geq \frac{\delta}{40n}m\right] \geq 1 - e^{\Omega(\delta m/n)}$

$$N_2 \overset{\text{def}}{=} \left\{k \in [m] \colon \left(\left|(\mathbf{W}'_L h_{i^*,L-1})_k\right| \leq \frac{\delta}{10n\sqrt{m}}\right) \bigwedge \left(\forall i \in [n]\setminus\{i^*\}, \quad \left|\left(\mathbf{W}'_L h_{i,L-1}\right)_k\right| \geq \frac{\delta}{4n\sqrt{m}}\right)\right\}$$

**Claim 9.5.** *Given set* $N_2 \subset [m]$ *and* $\vec{\mathsf{v}}$*, we have*

$$\mathbf{Pr}_{\mathbf{B}_k}\left[\left|\left\{k \in N_2 \colon |\langle \mathbf{B}_k, \mathsf{v}_{i^*}\rangle| \geq \frac{\|\mathsf{v}_{i^*}\|}{\sqrt{d}}\right\}\right| \geq \frac{|N_2|}{2}\right] \geq 1 - e^{-\Omega(|N_2|)}$$

*Proof of Claim 9.5.* Observe that each $\langle \mathbf{B}_k, \mathsf{v}_{i^*}\rangle$ follows from $\mathcal{N}(0, \|\mathsf{v}_{i^*}\|^2/d)$, so with probability at least 0.68 it satisfies $|\langle \mathbf{B}_k, \mathsf{v}_{i^*}\rangle| \geq \frac{\|\mathsf{v}_{i^*}\|}{\sqrt{d}}$. Using Chernoff bound we have the desired claim. □

Combining Claim 9.4 and Claim 9.5, we can obtain a set $N \subseteq [m]$ satisfying

$$\begin{aligned}
N \overset{\text{def}}{=} \Bigg\{k \in [m] \colon \left(\left|(\mathbf{W}'_L h_{i^*,L-1})_k\right| \leq \frac{\delta}{10n\sqrt{m}}\right) \bigwedge \left(\forall i \in [n]\setminus\{i^*\}, \quad \left|\left(\mathbf{W}'_L h_{i,L-1}\right)_k\right| \geq \frac{\delta}{4n\sqrt{m}}\right) \\
\bigwedge |\langle \mathbf{B}_k, \mathsf{v}_{i^*}\rangle| \geq \frac{\|\mathsf{v}_{i^*}\|}{\sqrt{d}}\Bigg\}
\end{aligned}$$

of cardinality $|N| \geq \frac{\delta}{100n}m$. Let us fix the randomness of $\mathbf{W}'_L$ so that $N$ is fixed. Let $k$ be any index in $N$. We can write

$$\widehat{\nabla}^{\vec{\mathsf{v}}}_{[\mathbf{W}_L]_k} F(\overrightarrow{\mathbf{W}}) = \sum_{i=1}^{n} \langle \mathbf{B}_k, \mathsf{v}_i \rangle \cdot h_{i,L-1} \cdot \mathbb{1}_{(\mathbf{W}'_L h_{i,L-1})_k + (\mathbf{W}''_L h_{i,L-1})_k \geq 0} .$$

The only remaining source of randomness comes from $\mathbf{W}''_L = \theta \widehat{g}_2 \widehat{h}^\top$.

Recalling that $\theta = \frac{1}{5n}$ and $\widehat{g}_2 \sim \mathcal{N}(0, \frac{2}{m}\mathbf{I})$, so since $\theta(\widehat{g}_2)_k \sim \mathcal{N}(0, \frac{2\theta^2}{m})$, using numerical values of Gaussian CDF, one can verify that

$$\mathbf{Pr}_{\widehat{g}_2}\left[ |\theta(\widehat{g}_2)_k| \in \Big( \frac{\delta}{9n\sqrt{m}}, \frac{\delta}{5n\sqrt{m}} \Big) \right] \geq 0.2 \enspace .$$

Let us denote this event of $\widehat{g}_2$ as $\mathfrak{E}_k$. Conditioning on $\mathfrak{E}_k$ happens, recalling $\|h_{i,L-1}\| \in [0.9, 1.1]$ from Lemma 7.1,

- For every $i \in [n] \setminus \{i^*\}$, we have

$$|(\mathbf{W}''_L h_{i,L-1})_k| = |(\theta \widehat{g}_2 \widehat{h}^\top h_{i,L-1})_k| \leq |(\theta \widehat{g}_2)_k| \cdot \|h_{i,L-1}\| < \frac{\delta}{5n\sqrt{m}} \cdot 1.1 < |(\mathbf{W}'_L h_{i,L-1})_k|$$

  and this means $\mathbb{1}_{(\mathbf{W}_L h_{i,L-1})_k \geq 0} = \mathbb{1}_{(\mathbf{W}'_L h_{i,L-1})_k \geq 0}$.
- For $i = i^*$, we have

$$|(\mathbf{W}''_L h_{i^*,L-1})_k| = |(\theta \widehat{g}_2 \widehat{h}^\top h_{i^*,L-1})_k| = |(\theta \widehat{g}_2)_k| \cdot \|h_{i^*,L-1}\| > \frac{\delta}{9n\sqrt{m}} \cdot 0.9 > |(\mathbf{W}'_L h_{i^*,L-1})_k|$$

  and this means $\mathbb{1}_{(\mathbf{W}_L h_{i^*,L-1})_k \geq 0} \neq \mathbb{1}_{(\mathbf{W}'_L h_{i^*,L-1})_k \geq 0}$ with probability exactly $\frac{1}{2}$— this is because, conditioning on event $\mathfrak{E}_k$, the sign of $(\theta \widehat{g}_2)_k$ is $\pm 1$ each with half probability.

Recall that for every $k \in N$,

$$\widehat{\nabla}^{\vec{\mathsf{v}}}_{[\mathbf{W}_L]_k} F(\overrightarrow{\mathbf{W}}) = \underbrace{\langle \mathbf{B}_k, \mathsf{v}_{i^*} \rangle \cdot h_{i^*,L-1} \cdot \mathbb{1}_{(\mathbf{W}_L h_{i^*,L-1})_k \geq 0}}_{\spadesuit} + \sum_{i \in [n] \setminus \{i^*\}} \underbrace{\langle \mathbf{B}_k, \mathsf{v}_i \rangle \cdot h_{i,L-1} \cdot \mathbb{1}_{(\mathbf{W}_L h_{i,L-1})_k \geq 0}}_{\clubsuit}$$

Now, fix the randomness of $\mathbf{A}, \mathbf{B}, \mathbf{W}_1, \dots, \mathbf{W}_{L-1}, \mathbf{W}'_L$ and let $\widehat{g}_2$ be the only randomness. Conditioning on $\mathfrak{E}_k$, we have that each term in $\clubsuit$ is fixed (i.e., independent of $\widehat{g}_2$) because $\mathbb{1}_{(\mathbf{W}_L h_{i,L-1})_k \geq 0} = \mathbb{1}_{(\mathbf{W}'_L h_{i,L-1})_k \geq 0}$. In contrast, conditioning on $\mathfrak{E}_k$, the indicator $\mathbb{1}_{(\mathbf{W}_L h_{i^*,L-1})_k \geq 0}$ of the $\spadesuit$ term may be 1 or 0 each with half probability. This means,

$$\mathbf{Pr}_{(\widehat{g}_2)_k}\left[ \|\widehat{\nabla}^{\vec{\mathsf{v}}}_{[\mathbf{W}_L]_k} F(\overrightarrow{\mathbf{W}})\|^2 \geq |\langle \mathbf{B}_k, \mathsf{v}_{i^*} \rangle|^2 \cdot \|h_{i^*,L-1}\|^2 \,\Big|\, k \in N \wedge \mathfrak{E}_k \right] \geq \frac{1}{2} \enspace .$$

Taking into account the fact that $|\langle \mathbf{B}_k, \mathsf{v}_{i^*} \rangle| \geq \frac{\|\mathsf{v}_{i^*}\|}{\sqrt{d}}$ (by definition of $N$), the fact that $\|h_{i,L-1}\| \geq 0.9$, and the fact that $\mathbf{Pr}_{(\widehat{g}_2)_k}[\mathfrak{E}] \geq 0.2$, we have

$$\mathbf{Pr}_{(\widehat{g}_2)_k}\left[ \|\widehat{\nabla}^{\vec{\mathsf{v}}}_{[\mathbf{W}_L]_k} F(\overrightarrow{\mathbf{W}})\|^2 \geq 0.8 \frac{\|\mathsf{v}_{i^*}\|^2}{d} \,\Big|\, k \in N \right] \geq \frac{1}{10} \enspace .$$

Using the independence of $(\widehat{g}_2)_k$ with respect to different $k \in N$, we can apply Chernoff bound and derive:

$$\mathbf{Pr}_{\widehat{g}_2}\left[ \sum_{k \in N} \|\widehat{\nabla}^{\vec{\mathsf{v}}}_{[\mathbf{W}_L]_k} F(\overrightarrow{\mathbf{W}})\|^2 \geq 0.8 \frac{\|\mathsf{v}_{i^*}\|^2}{d} \cdot \frac{|N|}{15} \,\Big|\, N \right] \geq 1 - e^{-\Omega(|N|)} \enspace .$$

Finally, using and $|N| \geq \frac{\delta}{100n}m$, we have

$$\mathbf{Pr}\left[ \|\widehat{\nabla}^{\vec{\mathsf{v}}}_{\mathbf{W}_L} F(\overrightarrow{\mathbf{W}})\|_F^2 \geq \frac{\|\mathsf{v}_{i^*}\|^2}{d} \frac{\delta}{2000n} m \right] \geq 1 - e^{-\Omega(\delta m / n)} \enspace .$$

We finish the upper bound proof of Lemma 9.3. ■

### 9.2.1 Proof of Claim 9.4

**Claim 9.4.** *We have* $\mathbf{Pr}_{\mathbf{W}'_L,\mathbf{W}_{L-1},\dots,\mathbf{W}_1,\mathbf{A}}\left[|N_2| \geq \frac{\delta}{40n}m\right] \geq 1 - e^{\Omega(\delta m/n)}$

$$N_2 \stackrel{\text{def}}{=} \left\{k \in [m] \colon \left(\left|(\mathbf{W}'_L h_{i^*,L-1})_k\right| \leq \frac{\delta}{10n\sqrt{m}}\right) \bigwedge \left(\forall i \in [n] \setminus \{i^*\}, \quad \left|\left(\mathbf{W}'_L h_{i,L-1}\right)_k\right| \geq \frac{\delta}{4n\sqrt{m}}\right)\right\}$$

*Proof of Claim 9.4.* Throughout the proof we assume $\mathbf{W}_{L-1},\dots,\mathbf{A}$ are good enough so that Lemma 7.1 holds (for $\varepsilon = 0.01$) and we fix their randomness. Define

$$N_1 \stackrel{\text{def}}{=} \left\{k \in [m] \colon \left|(\mathbf{W}'_L h_{i^*,L-1})_k\right| \leq \frac{\delta}{10n\sqrt{m}}\right\}$$

Since $\|h_{i^*,L-1}\|^2 \leq 1.1$ by Lemma 7.1, and since by definition of $\mathbf{W}'_L$ we have $(\mathbf{W}'_L h_{i^*,L-1})_k \sim \mathcal{N}(0, \frac{2(1-\theta^2)\|h_{i^*,L-1}\|^2}{m})$. By standard properties of Gaussian CDF (see Fact 9.6), we know $|(\mathbf{W}'_L h_{i^*,L-1})_k| \leq \frac{\delta}{10n\sqrt{m}}$ with probability at least $\frac{\delta}{25n}$ for each $k \in [m]$. By Chernoff bound,

$$\mathbf{Pr}_{\mathbf{W}'_L}\left[|N_1| \geq \frac{\delta}{30n}m\right] \geq 1 - e^{-\Omega(\delta m/n)}$$

Next, suppose we fix the randomness of $\mathbf{W}'_L\widehat{h}$. Define

$$N_2 \stackrel{\text{def}}{=} \left\{k \in N_1 \colon \forall i \in [n] \setminus \{i^*\}, \quad \left|\left(\mathbf{W}'_L h_{i,L-1}\right)_k\right| \geq \frac{\delta}{4n\sqrt{m}}\right\}$$

For each $k \in N_1$ and $i \in [n] \setminus \{i^*\}$, we can write

$$\mathbf{W}'_L h_{i,L-1} = \mathbf{W}'_L \widehat{h}(\widehat{h}^\top h_{i,L-1}) + \mathbf{W}'_L(\mathbf{I} - \widehat{h}\widehat{h}^\top)h_{i,L-1} \ .$$

Above, the first term on the right hand side is fixed (because we have fixed the randomness of $\mathbf{W}'_L\widehat{h}$); however, $\mathbf{W}'_L(\mathbf{I} - \widehat{h}\widehat{h}^\top)h_{i,L-1}$ is still fresh new random Gaussian. In symbols,

$$\mathbf{W}'_L h_{i,L-1} \sim \mathcal{N}\left(\mathbf{W}'_L\widehat{h}\widehat{h}^\top h_{i,L-1}, \frac{2\|(\mathbf{I} - \widehat{h}\widehat{h}^\top)h_{i,L-1}\|^2}{m}\mathbf{I}\right) \ .$$

According to Lemma 7.5, the variance here is at least $\frac{2}{m}\|(\mathbf{I} - \widehat{h}\widehat{h}^\top)h_{i,L-1}\|^2 \geq \frac{\delta^2}{2m}$. Using standard properties of Gaussian CDF (see Fact 9.6), we know $|(\mathbf{W}'_L h_{i,L-1})_k| \geq \frac{\delta}{4n\sqrt{m}}$ with probability at least $1 - \frac{1}{8n}$ for each $k \in [m]$. By union bound, for this $k \in [m]$, with probability at least $\frac{7}{8}$ we know $|(\mathbf{W}'_L h_{i,L-1})_k| \geq \frac{\delta}{4n\sqrt{m}}$ for all $i \in [n] \setminus \{i^*\}$. By Chernoff bound (over all $k \in N_1$), we conclude that

$$\mathbf{Pr}_{\mathbf{W}'_L}\left[|N_2| \geq \frac{3}{4}|N_1| \,\Big|\, N_1\right] \geq 1 - e^{-\Omega(|N_1|)} = 1 - e^{-\Omega(\delta m/n)} \ .$$

Combining the two bounds we finish the proof. □

**Fact 9.6.** *Suppose* $x \sim \mathcal{N}(0,\sigma^2)$ *is a Gaussian random variable. For any* $t \in (0,\sigma)$ *we have*

$$\mathbf{Pr}[x \geq t] \in \left[\tfrac{1}{2}(1 - \tfrac{4}{5}\tfrac{t}{\sigma}), \tfrac{1}{2}(1 - \tfrac{2}{3}\tfrac{t}{\sigma})\right] .$$

*Similarly, if* $x \sim \mathcal{N}(\mu,\sigma^2)$*, for any* $t \in (0,\sigma)$*, we have*

$$\mathbf{Pr}[|x| \geq t] \in \left[1 - \tfrac{4}{5}\tfrac{t}{\sigma}, 1 - \tfrac{2}{3}\tfrac{t}{\sigma}\right] .$$

## 10 Theorem 3: Gradient Bound at After Perturbation

In this section we prove our main theorem on the gradient upper and lower bounds.

**Theorem 3** (gradient bound, restated)**.** *Let* $\omega \stackrel{\text{def}}{=} O\big(\frac{\delta^{3/2}}{n^{9/2}L^6\log^3 m}\big)$*. With probability at least* $1 - e^{-\Omega(m\omega^{2/3}L)}$ *over the randomness of* $\overrightarrow{\mathbf{W}}^{(0)}, \mathbf{A}, \mathbf{B}$*, it satisfies for every* $\ell \in [L]$*, every* $i \in [n]$*, and every* $\overrightarrow{\mathbf{W}}$ *with* $\|\overrightarrow{\mathbf{W}} - \overrightarrow{\mathbf{W}}^{(0)}\|_2 \le \omega$*,*

$$\|\nabla_{\mathbf{W}_\ell} F_i(\overrightarrow{\mathbf{W}})\|_F^2 \le O\Big(\frac{F_i(\overrightarrow{\mathbf{W}})}{d} \times m\Big) \qquad \|\nabla_{\mathbf{W}_\ell} F(\overrightarrow{\mathbf{W}})\|_F^2 \le O\Big(\frac{F(\overrightarrow{\mathbf{W}})}{d} \times mn\Big)$$

$$\|\nabla_{\mathbf{W}_L} F(\overrightarrow{\mathbf{W}})\|_F^2 \ge \Omega\Big(\frac{\max_{i\in[n]} F_i(\overrightarrow{\mathbf{W}})}{dn/\delta} \times m\Big) \ .$$

*Remark* 10.1. Our Theorem 3 only gives gradient lower bound on $\|\nabla_{\mathbf{W}_L} F(\overrightarrow{\mathbf{W}})\|_F$. In principle, one can derive similar lower bounds on $\|\nabla_{\mathbf{W}_\ell} F(\overrightarrow{\mathbf{W}})\|_F$ for all $\ell = 1, 2, \dots, L-1$. However, the proof will be significantly more involved. We choose not to derive those bounds at the expense of losing a polynomial factor in $L$ in the final running time. For readers interested in the techniques for obtaining those bounds, we refer to them to the "randomness decomposition" part of our separate paper [6].

*Proof of Theorem 3.* Again we denote by $\mathbf{D}_{i,\ell}^{(0)}$ and $\mathbf{D}_{i,\ell}$ respectively the sign matrix at the initialization $\overrightarrow{\mathbf{W}}^{(0)}$ and at the current point $\overrightarrow{\mathbf{W}}$; and by $h_{i,\ell}^{(0)}$ and $h_{i,\ell}$ respectively the forward vector at $\overrightarrow{\mathbf{W}}^{(0)}$ and at $\overrightarrow{\mathbf{W}}$. Let us choose $s = O(m\omega^{2/3}L)$ which bounds the sparsity of $\|\mathbf{D}_{i,\ell} - \mathbf{D}_{i,\ell}^{(0)}\|_0$ by Lemma 8.2b. Recall

$$\begin{aligned}&\widehat{\nabla}^{\vec{\mathsf{v}}}_{\mathbf{W}_\ell} F(\overrightarrow{\mathbf{W}}^{(0)}) - \widehat{\nabla}^{\vec{\mathsf{v}}}_{\mathbf{W}_\ell} F(\overrightarrow{\mathbf{W}}) \\ &= \sum_{i=1}^n \Big( \big(\mathsf{v}_i^\top \mathbf{B}\mathbf{D}_{i,L}^{(0)}\mathbf{W}_L^{(0)} \cdots \mathbf{W}_{\ell+1}^{(0)}\mathbf{D}_{i,\ell}^{(0)}\big)^\top (h_{i,\ell-1}^{(0)})^\top - \big(\mathsf{v}_i^\top \mathbf{B}\mathbf{D}_{i,L}\mathbf{W}_L \cdots \mathbf{W}_{\ell+1}\mathbf{D}_{i,\ell}\big)^\top (h_{i,\ell-1})^\top \Big)\end{aligned} \tag{10.1}$$

By Lemma 8.7, we know that

$$\|\mathsf{v}_i^\top \mathbf{B}\mathbf{D}_{i,L}^{(0)}\mathbf{W}_L^{(0)} \cdots \mathbf{D}_{i,a}^{(0)}\mathbf{W}_a^{(0)}\mathbf{D}_{i,a-1}^{(0)} - \mathsf{v}_i^\top \mathbf{B}\mathbf{D}_{i,L}\mathbf{W}_L \cdots \mathbf{D}_{i,a}\mathbf{W}_a\mathbf{D}_{i,a-1}\| \le O(\omega^{1/3}L^2\sqrt{m\log m}/\sqrt{d}) \cdot \|\mathsf{v}_i\|$$

By Lemma 7.4b we know

$$\|\mathsf{v}_i^\top \mathbf{B}\mathbf{D}_{i,L}^{(0)}\mathbf{W}_L^{(0)} \cdots \mathbf{D}_{i,a}^{(0)}\mathbf{W}_a^{(0)}\mathbf{D}_{i,a-1}^{(0)}\| \le O(\sqrt{m/d}) \cdot \|\mathsf{v}_i\|$$

By Lemma 7.1 and Lemma 8.2c, we have

$$\|h_{i,\ell-1}\| \le 1.1 \quad \text{and} \quad \|h_{i,\ell-1} - h_{i,\ell-1}^{(0)}\| \le O(\omega L^{5/2}\sqrt{\log m})$$

Together, they imply

$$\begin{aligned}\Big\|\widehat{\nabla}^{\vec{\mathsf{v}}}_{\mathbf{W}_\ell} F(\overrightarrow{\mathbf{W}}^{(0)}) - \widehat{\nabla}^{\vec{\mathsf{v}}}_{\mathbf{W}_\ell} F(\overrightarrow{\mathbf{W}})\Big\|_F^2 &\le n\|\vec{\mathsf{v}}\|^2 \cdot O\left(\omega^{1/3}L^2\sqrt{m\log m}/\sqrt{d} + \sqrt{m/d} \times \omega L^{5/2}\sqrt{\log m}\right)^2 \\ &\le n\|\vec{\mathsf{v}}\|^2 \cdot O\left(\frac{m\log m}{d} \cdot \omega^{2/3}L^4\right) \ .\end{aligned} \tag{10.2}$$

With our parameter assumption on $\omega$, this together with Lemma 9.3 implies the same upper and lower bounds at point $\overrightarrow{\mathbf{W}} = \overrightarrow{\mathbf{W}}^{(0)} + \overrightarrow{\mathbf{W}}'$:

$$\|\widehat{\nabla}^{\vec{\mathsf{v}}}_{\mathbf{W}_\ell} F_i(\overrightarrow{\mathbf{W}}^{(0)} + \overrightarrow{\mathbf{W}}')\|_F^2 \le O\Big(\frac{\|\mathsf{v}_i\|^2}{d} \times m\Big) \quad \|\widehat{\nabla}^{\vec{\mathsf{v}}}_{\mathbf{W}_\ell} F(\overrightarrow{\mathbf{W}}^{(0)} + \overrightarrow{\mathbf{W}}')\|_F^2 \le O\Big(\frac{\|\vec{\mathsf{v}}\|^2}{d} \times mn\Big)$$

$$\|\widehat{\nabla}^{\vec{\mathsf{v}}}_{\mathbf{W}_L} F(\overrightarrow{\mathbf{W}}^{(0)} + \overrightarrow{\mathbf{W}}')\|_F^2 \ge \Omega\Big(\frac{\max_{i\in[n]} \|\mathsf{v}_i\|^2}{dn/\delta} \times m\Big) \ .$$

Finally, taking $\varepsilon$-net over all possible vectors $\vec{\mathsf{v}} = (\mathsf{v}_1, \dots, \mathsf{v}_n) \in (\mathbb{R}^d)^n$, we know that the above bounds hold not only for fixed $\vec{\mathsf{v}}$ but for all $\vec{\mathsf{v}}$. In particular, we can now plug in the choice of $\mathsf{v}_i = \mathsf{loss}_i = \mathbf{B}h_{i,L} - y_i^*$ and it implies our desired bounds on the true gradients. □

# 11 Theorem 4: Objective Semi-Smoothness

The purpose of this section is to prove

**Theorem 4** (objective semi-smoothness, restated)**.** *Let* $\omega \in \big[\Omega(\frac{d^{3/2}}{m^{3/2}L^{3/2}\log^{3/2} m}), O(\frac{1}{L^{4.5}\log^3 m})\big]$ *and* $\overrightarrow{\mathbf{W}}^{(0)}, \mathbf{A}, \mathbf{B}$ *be at random initialization. With probability at least* $1 - e^{-\Omega(m\omega^{2/3}L)}$ *over the randomness of* $\overrightarrow{\mathbf{W}}^{(0)}, \mathbf{A}, \mathbf{B}$*, we have for every* $\breve{\overrightarrow{\mathbf{W}}} \in (\mathbb{R}^{m\times m})^L$ *with* $\|\breve{\overrightarrow{\mathbf{W}}} - \overrightarrow{\mathbf{W}}^{(0)}\|_2 \le \omega$*, and for every* $\overrightarrow{\mathbf{W}}' \in (\mathbb{R}^{m\times m})^L$ *with* $\|\overrightarrow{\mathbf{W}}'\|_2 \le \omega$*, we have*

$$F(\breve{\overrightarrow{\mathbf{W}}} + \overrightarrow{\mathbf{W}}') \le F(\breve{\overrightarrow{\mathbf{W}}}) + \langle \nabla F(\breve{\overrightarrow{\mathbf{W}}}), \overrightarrow{\mathbf{W}}' \rangle + \sqrt{nF(\breve{\overrightarrow{\mathbf{W}}})} \cdot \frac{\omega^{1/3}L^2\sqrt{m\log m}}{\sqrt{d}} \cdot O(\|\overrightarrow{\mathbf{W}}'\|_2) + O\Big(\frac{nL^2m}{d}\Big)\|\overrightarrow{\mathbf{W}}'\|_2^2$$

We introduce the following notations before we go to proofs.

**Definition 11.1.** *For* $i \in [n]$ *and* $\ell \in [L]$*:*

$$\begin{aligned}
g_{i,0}^{(0)} &= \mathbf{A}x_i & \breve{g}_{i,0} &= \mathbf{A}x_i & g_{i,0} &= \mathbf{A}x_i \\
h_{i,0}^{(0)} &= \phi(\mathbf{A}x_i) & \breve{h}_{i,0} &= \phi(\mathbf{A}x_i) & h_{i,0} &= \phi(\mathbf{A}x_i) \\
g_{i,\ell}^{(0)} &= \mathbf{W}_\ell^{(0)} h_{i,\ell-1}^{(0)} & \breve{g}_{i,\ell} &= \breve{\mathbf{W}}_\ell \breve{h}_{i,\ell-1} & g_{i,\ell} &= (\breve{\mathbf{W}}_\ell + \mathbf{W}'_\ell)h_{i,\ell-1} \\
h_{i,\ell}^{(0)} &= \phi(\mathbf{W}_\ell^{(0)} h_{i,\ell-1}^{(0)}) & \breve{h}_{i,\ell} &= \phi(\breve{\mathbf{W}}_\ell \breve{h}_{i,\ell-1}) & h_{i,\ell} &= \phi((\breve{\mathbf{W}}_\ell + \mathbf{W}'_\ell)h_{i,\ell-1})
\end{aligned}$$

$$\breve{\mathsf{loss}}_i = B\breve{h}_{i,L} - y_i^*$$

*Define diagonal matrices* $\mathbf{D}_{i,\ell}^{(0)} \in \mathbb{R}^{m\times m}$ *and* $\breve{\mathbf{D}}_{i,\ell} \in \mathbb{R}^{m\times m}$ *respectively by letting*

$$(\mathbf{D}_{i,\ell}^{(0)})_{k,k} = \mathbb{1}_{(g_{i,\ell}^{(0)})_k \ge 0} \text{ and } (\breve{\mathbf{D}}_{i,\ell})_{k,k} = \mathbb{1}_{(\breve{g}_{i,\ell})_k \ge 0}, \forall k \in [m].$$

The following claim gives rise to a new recursive formula to calculate $h_{i,\ell} - \breve{h}_{i,\ell}$.

**Claim 11.2.** *There exist diagonal matrices* $\mathbf{D}''_{i,\ell} \in \mathbb{R}^{m\times m}$ *with entries in* $[-1, 1]$ *such that,*

$$\forall i \in [n], \forall \ell \in [L]: \quad h_{i,\ell} - \breve{h}_{i,\ell} = \sum_{a=1}^{\ell} (\breve{\mathbf{D}}_{i,\ell} + \mathbf{D}''_{i,\ell})\breve{\mathbf{W}}_\ell \cdots \breve{\mathbf{W}}_{a+1}(\breve{\mathbf{D}}_{i,a} + \mathbf{D}''_{i,a})\mathbf{W}'_a h_{i,a-1} \tag{11.1}$$

*Furthermore, we have* $\|h_{i,\ell} - \breve{h}_{i,\ell}\| \le O(L^{1.5})\|\mathbf{W}'\|_2$*,* $\|\mathbf{B}h_{i,\ell} - \mathbf{B}\breve{h}_{i,\ell}\| \le O(L\sqrt{m/d})\|\mathbf{W}'\|_2$ *and* $\|\mathbf{D}''_{i,\ell}\|_0 \le O(m\omega^{2/3}L)$*.*

*Proof of Theorem 4.* First of all, since

$$\frac{1}{2}\|\mathbf{B}h_{i,L} - y_i^*\|^2 = \frac{1}{2}\|\breve{\mathsf{loss}}_i + \mathbf{B}(h_{i,L} - \breve{h}_{i,L})\|^2 = \frac{1}{2}\|\breve{\mathsf{loss}}_i\|^2 + \breve{\mathsf{loss}}_i^\top \mathbf{B}(h_{i,L} - \breve{h}_{i,L}) + \frac{1}{2}\|\mathbf{B}(h_{i,L} - \breve{h}_{i,L})\|^2 \tag{11.2}$$

we can write

$$
\begin{aligned}
&F(\breve{\overrightarrow{\mathbf{W}}} + \overrightarrow{\mathbf{W}}') - F(\overrightarrow{\mathbf{W}}) - \langle \nabla F(\overrightarrow{\mathbf{W}}), \overrightarrow{\mathbf{W}}' \rangle \\
&\overset{①}{=} -\langle \nabla F(\breve{\overrightarrow{\mathbf{W}}}), \overrightarrow{\mathbf{W}}' \rangle + \frac{1}{2}\sum_{i=1}^{n} \|\mathbf{B}h_{i,L} - y^*_{i,L}\|^2 - \|\mathbf{B}\breve{h}_{i,L} - y^*_{i,L}\|^2 \\
&\overset{②}{=} -\langle \nabla F(\breve{\overrightarrow{\mathbf{W}}}), \overrightarrow{\mathbf{W}}' \rangle + \sum_{i=1}^{n} \breve{\mathsf{loss}}_i^\top \mathbf{B}(h_{i,L} - \breve{h}_{i,L}) + \frac{1}{2}\|\mathbf{B}(h_{i,L} - \breve{h}_{i,L})\|^2 \\
&\overset{③}{=} \sum_{i=1}^{n} \breve{\mathsf{loss}}_i^\top \mathbf{B}\left( (h_{i,L} - \breve{h}_{i,L}) - \sum_{\ell=1}^{L} \breve{\mathbf{D}}_{i,L}\breve{\mathbf{W}}_L \cdots \breve{\mathbf{W}}_{\ell+1}\breve{\mathbf{D}}_{i,\ell}\mathbf{W}'_\ell \breve{h}_{i,\ell-1} \right) + \frac{1}{2}\|\mathbf{B}(h_{i,L} - \breve{h}_{i,L})\|^2 \\
&\overset{④}{=} \sum_{i=1}^{n} \breve{\mathsf{loss}}_i^\top \mathbf{B}\left( \sum_{\ell=1}^{L} (\breve{\mathbf{D}}_{i,L} + \mathbf{D}''_{i,L})\breve{\mathbf{W}}_L \cdots \breve{\mathbf{W}}_{\ell+1}(\breve{\mathbf{D}}_{i,\ell} + \mathbf{D}''_{i,\ell})\mathbf{W}'_\ell h_{i,\ell-1} - \breve{\mathbf{D}}_{i,L}\breve{\mathbf{W}}_L \cdots \breve{\mathbf{W}}_{\ell+1}\breve{\mathbf{D}}_{i,\ell}\mathbf{W}'_\ell \breve{h}_{i,\ell-1} \right) \\
&\quad + \frac{1}{2}\sum_{i=1}^{n} \|\mathbf{B}(h_{i,L} - \breve{h}_{i,L})\|^2 \qquad (11.3)
\end{aligned}
$$

Above, ① is by the definition of $F(\cdot)$; ② is by (11.2); ③ is by the definition of $\nabla F(\cdot)$ (see Fact 2.6 for an explicit form of the gradient).

We next bound the RHS of (11.3). We first note that by Lemma 8.2b, we have $\|\breve{\mathbf{D}}_{i,\ell} + \mathbf{D}''_{i,\ell} - \mathbf{D}^{(0)}_{i,\ell}\|_0 \leq s$ and $\|\breve{\mathbf{D}}_{i,\ell} - \mathbf{D}^{(0)}_{i,\ell}\|_0 \leq s$ for $s = O(m\omega^{2/3}L)$.

We ignore subscripts in $i$ for notational convenience. We first use Claim 11.2 to get

$$\|\mathbf{B}(h_L - \breve{h}_L)\| \leq O(L\sqrt{m/d}) \cdot \|\overrightarrow{\mathbf{W}}'\|_2 \ . \qquad (11.4)$$

Next we calculate that

$$
\begin{aligned}
&\left| \breve{\mathsf{loss}}_i^\top \mathbf{B}(\breve{\mathbf{D}}_L + \mathbf{D}''_L)\breve{\mathbf{W}}_L \cdots (\breve{\mathbf{D}}_\ell + \mathbf{D}''_\ell)\mathbf{W}'_\ell h_{\ell-1} - \breve{\mathsf{loss}}_i^\top \mathbf{B}\breve{\mathbf{D}}_L\breve{\mathbf{W}}_L \cdots \breve{\mathbf{D}}_\ell \mathbf{W}'_\ell h_{\ell-1} \right| \\
&\leq \|\breve{\mathsf{loss}}_i\| \cdot \underbrace{\left\| \mathbf{B}(\breve{\mathbf{D}}_L + \mathbf{D}''_L)\breve{\mathbf{W}}_L \cdots \breve{\mathbf{W}}_{\ell-1}(\breve{\mathbf{D}}_\ell + \mathbf{D}''_\ell) - \mathbf{B}\breve{\mathbf{D}}_L\breve{\mathbf{W}}_L \cdots \breve{\mathbf{W}}_{\ell-1}\breve{\mathbf{D}}_\ell \right\|_2}_{\text{Lemma 8.7 with } s = O(m\omega^{2/3}L)} \cdot \|\mathbf{W}'_\ell h_{\ell-1}\| \\
&\leq \|\breve{\mathsf{loss}}_i\| \cdot O\left( \frac{\sqrt{L^3\omega^{2/3}Lm\log m}}{\sqrt{d}} \right) \cdot O(\|\mathbf{W}'_\ell\|_2) \ . \qquad (11.5)
\end{aligned}
$$

Finally, we also have

$$
\begin{aligned}
&\left| \breve{\mathsf{loss}}_i^\top \mathbf{B}\breve{\mathbf{D}}_L\breve{\mathbf{W}}_L \cdots \breve{\mathbf{D}}_\ell \mathbf{W}'_\ell (h_{\ell-1} - \breve{h}_{\ell-1}) \right| \\
&\overset{①}{\leq} \|\breve{\mathsf{loss}}_i\| \cdot O\left( \sqrt{m/d} + \frac{\omega^{1/3}L^2\sqrt{m\log m}}{\sqrt{d}} \right) \cdot \|\mathbf{W}'_\ell\|_2 \cdot \|h_\ell - \breve{h}_\ell\|_2 \\
&\overset{②}{\leq} O(L^{0.5}\sqrt{m/d}) \cdot \|\breve{\mathsf{loss}}_i\|_2 \cdot L^{1.5}\|\mathbf{W}'_\ell\|^2 \qquad (11.6)
\end{aligned}
$$

where ① uses Lemma 7.4b (and Lemma 8.7 for bounding the perturbation) and ② uses Claim 11.2 to bound $\|h_\ell - \breve{h}_\ell\|_2$ and our choice of $\omega$.

Putting (11.4), (11.5) and (11.6) back to (11.3), and using triangle inequality, we have the desired result. □

## 11.1 Proof of Claim 11.2

We first present a simple proposition about the ReLU function.

**Proposition 11.3.** *Given vectors $a, b \in \mathbb{R}^m$ and $\mathbf{D} \in \mathbb{R}^{m\times m}$ the diagonal matrix where $\mathbf{D}_{k,k} = \mathbb{1}_{a_k \geq 0}$. Then, then there exists a diagonal matrix $\mathbf{D}'' \in \mathbb{R}^{m\times m}$ with*

- $|\mathbf{D}_{k,k} + \mathbf{D}''_{k,k}| \leq 1$ *and* $|\mathbf{D}''_{k,k}| \leq 1$ *for every* $k \in [m]$,
- $\mathbf{D}''_{k,k} \neq 0$ *only when* $\mathbb{1}_{a_k\geq 0} \neq \mathbb{1}_{b_k \geq 0}$, *and*
- $\phi(a) - \phi(b) = (\mathbf{D} + \mathbf{D}'')(a - b)$

*Proof.* We verify coordinate by coordinate for each $k \in [m]$.

- If $a_k \geq 0$ and $b_k \geq 0$, then $(\phi(a) - \phi(b))_k = a_k - b_k = \big(\mathbf{D}(a-b)\big)_k$.
- If $a_k < 0$ and $b_k < 0$, then $(\phi(a) - \phi(b))_k = 0 - 0 = \big(\mathbf{D}(a-b)\big)_k$.
- If $a_k \geq 0$ and $b_k < 0$, then $(\phi(a) - \phi(b))_k = a_k = (a_k - b_k) + \frac{b_k}{a_k - b_k}(a_k - b_k) = \big(\mathbf{D}(a-b) + \mathbf{D}''(a-b)\big)_k$, if we define $(\mathbf{D}'')_{k,k} = \frac{b_k}{a_k-b_k} \in [-1, 0]$.
- If $a_k < 0$ and $b_k \geq 0$, then $(\phi(a) - \phi(b))_k = -b_k = 0 \cdot (a_k - b_k) - \frac{b_k}{b_k - a_k}(a_k - b_k) = \big(\mathbf{D}(a-b) + \mathbf{D}''(a-b)\big)_k$, if we define $(\mathbf{D}'')_{k,k} = \frac{b_k}{b_k - a_k} \in [0, 1]$. □

*Proof of Claim 11.2.* We ignore the subscript in $i$ for cleanness, and calculate that

$$
\begin{aligned}
h_\ell - \breve{h}_\ell &\overset{①}{=} \phi((\breve{\mathbf{W}}_\ell + \mathbf{W}'_\ell)h_{\ell-1}) - \phi(\breve{\mathbf{W}}_\ell \breve{h}_{\ell-1}) \\
&\overset{②}{=} (\breve{\mathbf{D}}_\ell + \mathbf{D}''_\ell)\Big((\breve{\mathbf{W}}_\ell + \mathbf{W}'_\ell)h_{\ell-1} - \breve{\mathbf{W}}_\ell \breve{h}_{\ell-1}\Big) \\
&= (\breve{\mathbf{D}}_\ell + \mathbf{D}''_\ell)\breve{\mathbf{W}}_\ell(h_{\ell-1} - \breve{h}_{\ell-1}) + (\breve{\mathbf{D}}_\ell + \mathbf{D}''_\ell)\mathbf{W}'_\ell h_{\ell-1} \\
&\overset{③}{=} \sum_{a=1}^{\ell} (\breve{\mathbf{D}}_\ell + \mathbf{D}''_\ell)\breve{\mathbf{W}}_\ell \cdots \breve{\mathbf{W}}_{a+1}(\breve{\mathbf{D}}_a + \mathbf{D}''_a)\mathbf{W}'_a h_{a-1}
\end{aligned}
$$

Above, ① is by the recursive definition of $h_\ell$ and $\breve{h}_\ell$; ② is by Proposition 11.3 and $\mathbf{D}''_\ell$ is defined according to Proposition 11.3; and inequality ③ is by recursively computing $h_{\ell-1} - \breve{h}_{\ell-1}$. As for the remaining properties:

- We have $\|\mathbf{D}''_\ell\|_0 \leq O(m\omega^{2/3}L)$.

  This is because, $(\mathbf{D}''_\ell)_{k,k}$ is non-zero only at the coordinates $k \in [m]$ where the signs of $\breve{g}_\ell$ and $g_\ell$ are opposite (by Proposition 11.3). Such a coordinate $k$ must satisfy either $(\mathbf{D}^{(0)}_\ell)_{k,k} \neq (\breve{\mathbf{D}}_\ell)_{k,k}$ or $(\mathbf{D}^{(0)}_\ell)_{k,k} \neq (\mathbf{D}_\ell)_{k,k}$, and therefore by Lemma 8.2b there are at most $O(m\omega^{2/3}L)$ such coordinates $k$.

- We have $\|h_\ell - \breve{h}_\ell\| \leq O(L^{1.5})\|\overrightarrow{\mathbf{W}}'\|_2$.

  This is because we have $\big\|(\breve{\mathbf{D}}_\ell + \mathbf{D}''_\ell)\breve{\mathbf{W}}_\ell \cdots \breve{\mathbf{W}}_{a+1}(\breve{\mathbf{D}}_a + \mathbf{D}''_a)\big\|_2 \leq O(\sqrt{L})$ from Lemma 8.6b, we have $\|h_{a-1}\| \leq O(1)$ (by $\|h^{(0)}_{a-1}\| \leq O(1)$ from Lemma 7.1 and $\|h^{(0)}_{a-1} - h_{a-1}\| \leq o(1)$ from Lemma 8.2c); and and $\|\mathbf{W}'_a h_{a-1}\| \leq \|\mathbf{W}'_a\|_2 \|h_{a-1}\| \leq O(\|\overrightarrow{\mathbf{W}}'\|_2)$.

- We have $\|\mathbf{B}h_\ell - \mathbf{B}\breve{h}_\ell\| \leq O(L\sqrt{m/d})\|\overrightarrow{\mathbf{W}}'\|_2$.

  This is because we have $\big\|\mathbf{B}(\breve{\mathbf{D}}_\ell + \mathbf{D}''_\ell)\breve{\mathbf{W}}_\ell \cdots \breve{\mathbf{W}}_{a+1}(\breve{\mathbf{D}}_a + \mathbf{D}''_a)\big\|_2 \leq O(\sqrt{m/d})$ from Lemma 7.4b (along with perturbation bound Lemma 8.7), we have $\|h_{a-1}\| \leq O(1)$ (by $\|h^{(0)}_{a-1}\| \leq O(1)$ from Lemma 7.1 and $\|h^{(0)}_{a-1} - h_{a-1}\| \leq o(1)$ from Lemma 8.2c); and and $\|\mathbf{W}'_a h_{a-1}\| \leq \|\mathbf{W}'_a\|_2 \|h_{a-1}\| \leq O(\|\overrightarrow{\mathbf{W}}'\|_2)$.

□

## 12 Theorem 1: Convergence Rate of GD

**Theorem 1** (gradient descent, restated)**.** *For any* $\varepsilon \in (0,1]$*,* $\delta \in \big(0, O(\frac{1}{L})\big]$*. Let* $m \geq \widetilde{\Omega}\big(\mathsf{poly}(n, L, \delta^{-1}) d\big)$*,* $\eta = \Theta\big(\frac{d\delta}{\mathsf{poly}(n,L)m}\big)$*, and* $\overrightarrow{\mathbf{W}}^{(0)}, \mathbf{A}, \mathbf{B}$ *are at random initialization. Then, with probability at least* $1 - e^{-\Omega(\log^2 m)}$*, suppose we start at* $\overrightarrow{\mathbf{W}}^{(0)}$ *and for each* $t = 0, 1, \dots, T-1$*,*

$$\overrightarrow{\mathbf{W}}^{(t+1)} = \overrightarrow{\mathbf{W}}^{(t)} - \eta \nabla F(\overrightarrow{\mathbf{W}}^{(t)}) \enspace .$$

*Then, it satisfies*

$$F(\overrightarrow{\mathbf{W}}^{(T)}) \leq \varepsilon \quad \textit{for} \quad T = \Theta\left(\frac{\mathsf{poly}(n,L)}{\delta^2} \log \frac{1}{\varepsilon}\right) \enspace .$$

*In other words, the training loss drops to* $\varepsilon$ *in a linear convergence speed.*

*Proof of Theorem 1.* Using Lemma 7.1 we have $\|h_{i,L}\|_2 \leq 1.1$ and then using the randomness of $\mathbf{B}$, it is easy to show that $\|\mathbf{B} h_{i,L}^{(0)} - y_i^*\|^2 \leq O(\log^2 m)$ with at least $1 - e^{-\Omega(\log^2 m)}$ (where $h_{i,L}^{(0)}$ is defined with respect to the random initialization $\overrightarrow{\mathbf{W}}^{(0)}$), and therefore

$$F(\overrightarrow{\mathbf{W}}^{(0)}) \leq O(n \log^2 m) \enspace .$$

Let us assume for every $t = 0, 1, \dots, T-1$, the following holds

$$\|\overrightarrow{\mathbf{W}}^{(t)} - \overrightarrow{\mathbf{W}}^{(0)}\|_F \leq \omega \stackrel{\mathrm{def}}{=} O\left(\frac{n^3 \sqrt{d}}{\delta \sqrt{m}} \log m\right) \enspace . \tag{12.1}$$

We shall prove the convergence of GD assuming (12.1) holds, so that previous statements such as Theorem 4 and Theorem 3 can be applied. At the end of the proof, we shall verify that (12.1) is satisfied.

To make the proof simple, we choose

$$m \geq \Omega\big(\frac{n^{24} L^{12} d \log^5 m}{\delta^8}\big), \quad \eta = \Theta\big(\frac{d\delta}{n^4 L^2 m}\big), \quad T = \Theta\left(\frac{n^6 L^2}{\delta^2} \log \frac{1}{\varepsilon}\right)$$

We emphasize that

- Most of the polynomial dependency in $n, L, \delta^{-1}$ come from the non-smoothness of the ReLU activation; if one instead studies smooth activations, their power can be significantly reduced. For instance, for smooth activation functions, one does not need the semi-smoothness Theorem 4.
- We have not tried to tighten the polynomial dependency on $n, L, \delta^{-1}$. We are aware of many ways to improve the constant in the exponents at the expense of complicating the proofs. Since the main focus of this paper is to derive the first *polynomial* running time, we do not include such improvements.

Letting $\nabla_t = \nabla F(\overrightarrow{\mathbf{W}}^{(t)})$, we calculate that

$$\begin{aligned}
& F(\overrightarrow{\mathbf{W}}^{(t+1)}) \\
\overset{①}{\leq}\ & F(\overrightarrow{\mathbf{W}}^{(t)}) - \eta \|\nabla F(\overrightarrow{\mathbf{W}}^{(t)})\|_F^2 + \eta \sqrt{n F(\overrightarrow{\mathbf{W}}^{(t)})} \cdot O\left(\frac{\omega^{1/3} L^2 \sqrt{m \log m}}{\sqrt{d}}\right) \cdot \|\nabla_t\|_2 + O\big(\eta^2 \frac{n L^2 m}{d}\big) \|\nabla_t\|_2^2 \\
\overset{②}{\leq}\ & F(\overrightarrow{\mathbf{W}}^{(t)}) - \eta \|\nabla F(\overrightarrow{\mathbf{W}}^{(t)})\|_F^2 + O\left(\frac{\eta n L^2 m \omega^{1/3} \sqrt{\log m}}{d} + \frac{\eta^2 n^2 L^2 m^2}{d^2}\right) \cdot F(\overrightarrow{\mathbf{W}}^{(t)}) \\
\overset{③}{\leq}\ & \left(1 - \Omega\big(\frac{\eta \delta m}{d n^2}\big)\right) F(\overrightarrow{\mathbf{W}}^{(t)}) \enspace .
\end{aligned} \tag{12.2}$$

Above, ① uses Theorem 4; ② uses Theorem 3 (which gives $\|\nabla_t\|_2^2 \leq \max_{\ell\in[L]} \|\nabla_{\mathbf{W}_\ell} F(\overrightarrow{\mathbf{W}}^{(t)})\|_F^2 \leq O\big(\frac{F(\overrightarrow{\mathbf{W}}^{(t)})}{d} \times mn\big)$); ③ use gradient lower bound from Theorem 3 and our choice of $\eta$. In other words, after $T = \Theta(\frac{dn^2}{\eta\delta m}) \log \frac{n \log m}{\varepsilon}$ iterations we have $F(\overrightarrow{\mathbf{W}}^{(T)}) \leq \varepsilon$.

We need to verify for each $t$, $\|\overrightarrow{\mathbf{W}}^{(t)} - \overrightarrow{\mathbf{W}}^{(0)}\|_F$ is small so that (12.1) holds. By Theorem 3,

$$
\begin{aligned}
\|\mathbf{W}_\ell^{(t)} - \mathbf{W}_\ell^{(0)}\|_F &\leq \sum_{i=0}^{t-1} \|\eta \nabla_{\mathbf{W}_\ell} F(\overrightarrow{\mathbf{W}}^{(i)})\|_F \leq O(\eta\sqrt{nm/d}) \cdot \sum_{i=0}^{t-1} \sqrt{F(\overrightarrow{\mathbf{W}}^{(i)})} \\
&\leq O(\eta\sqrt{nm/d}) \cdot \Theta(\frac{dn^2}{\eta\delta m}) \cdot O(\sqrt{n \log^2 m}) \leq O\left(\frac{n^3\sqrt{d}}{\delta\sqrt{m}} \log m\right) .
\end{aligned}
$$

where the last step follows by our choice of $T$. □

# 13 Theorem 2: Convergence Rate of SGD

**Theorem 2** (stochastic gradient descent, stated). *For any $\varepsilon \in (0,1]$, $\delta \in \big(0, O(\frac{1}{L})\big]$, $b \in [n]$. Let $m \geq \widetilde{\Omega}\big(\frac{\mathsf{poly}(n,L,\delta^{-1})\cdot d}{b}\big)$, $\eta \stackrel{\text{def}}{=} \Theta(\frac{b\delta d}{\mathsf{poly}(n,L) m \log^2 m})$, and $\overrightarrow{\mathbf{W}}^{(0)}, \mathbf{A}, \mathbf{B}$ are at random initialization. Suppose we start at $W^{(0)}$ and for each $t = 0, 1, \ldots, T-1$,*

$$
W^{(t+1)} = W^{(t)} - \eta \cdot \frac{n}{|S_t|} \sum_{i\in S_t} \nabla F(W^{(t)}) \qquad \text{(for a random subset } S_t \subseteq [n] \text{ of fixed cardinality } b.)
$$

*Then, it satisfies with probability at least $1 - e^{-\Omega(\log^2 m)}$ over the randomness of $S_1, \ldots, S_T$:*

$$
F(W^{(T)}) \leq \varepsilon \quad \textit{for all} \quad T = \Theta\Big(\frac{\mathsf{poly}(n,L) \log^2 m}{b\delta^2} \log \frac{n \log m}{\varepsilon}\Big) .
$$

The proof of Theorem 2 is the same as Theorem 1 plus the careful use of martingale concentration.

*Proof of Theorem 2.* Using similar argument as the proof of Theorem 1, we have with at least $1 - e^{-\Omega(\log^2 m)}$ probability

$$
F(\overrightarrow{\mathbf{W}}^{(0)}) \leq O(n \log^2 m) .
$$

Let us assume for every $t = 0, 1, \ldots, T-1$, the following holds

$$
\|\overrightarrow{\mathbf{W}}^{(t)} - \overrightarrow{\mathbf{W}}^{(0)}\|_F \leq \omega \stackrel{\text{def}}{=} O\left(\frac{n^{3.5}\sqrt{d}}{\delta\sqrt{bm}} \log m\right) . \tag{13.1}
$$

We shall prove the convergence of SGD assuming (13.1) holds, so that previous statements such as Theorem 4 and Theorem 3 can be applied. At the end of the proof, we shall verify that (13.1) is satisfied throughout the SGD with high probability.

To make the proof simple, we choose

$$
m \geq \Omega\big(\frac{n^{24} L^{12} b d \log^5 m}{\delta^8}\big), \quad \eta = \Theta(\frac{b\delta d}{n^5 L^2 m \log^2 m}), \quad T = \Theta\Big(\frac{dn^2}{\eta\delta m} \log \frac{n \log m}{\varepsilon}\Big) = \Theta\Big(\frac{n^7 L^2 \log^2 m}{b\delta^2} \log \frac{n \log m}{\varepsilon}\Big)
$$

We emphasize that

- Most of the polynomial dependency in $n, L, \delta^{-1}$ come from the non-smoothness of the ReLU activation; if one instead studies smooth activations, their power can be significantly reduced. For instance, for smooth activation functions, one does not need the semi-smoothness Theorem 4.
- We have not tried to tighten the polynomial dependency on $n, L, \delta^{-1}$. We are aware of many ways to improve the constant in the exponents at the expense of complicating the proofs. Since the main focus of this paper is to derive the first *polynomial* running time, we do not include such improvements.

For each $t = 0, 1, \dots, T-1$, using the same notation as Theorem 1, except that we choose $\nabla_t = \frac{n}{|S_t|}\sum_{i \in S_t} \nabla F_i(\overrightarrow{\mathbf{W}}^{(t)})$. We have $\mathbb{E}_{S_t}[\nabla_t] = \nabla F(\overrightarrow{\mathbf{W}}^{(t)})$ and therefore

$$\begin{aligned}
&\mathop{\mathbb{E}}_{S_t}[F(\overrightarrow{\mathbf{W}}^{(t+1)})] \\
&\overset{①}{\leq} F(\overrightarrow{\mathbf{W}}^{(t)}) - \eta\|\nabla F(\overrightarrow{\mathbf{W}}^{(t)})\|_F^2 + \eta\sqrt{nF(\overrightarrow{\mathbf{W}}^{(t)})} \cdot O\left(\frac{\omega^{1/3}L^2\sqrt{m\log m}}{\sqrt{d}}\right) \cdot \mathop{\mathbb{E}}_{S_t}[\|\nabla_t\|_2] \\
&\quad + O\big(\eta^2\frac{nL^2m}{d}\big)\mathop{\mathbb{E}}_{S_t}[\|\nabla_t\|_2^2] \\
&\overset{②}{\leq} F(\overrightarrow{\mathbf{W}}^{(t)}) - \eta\|\nabla_t\|_F^2 + O\left(\frac{\eta nL^2m\omega^{1/3}\sqrt{\log m}}{d} + \frac{\eta^2n^2L^2m^2}{d^2}\right) \cdot F(\overrightarrow{\mathbf{W}}^{(t)}) \\
&\overset{③}{\leq} \left(1 - \Omega\big(\frac{\eta\delta m}{dn^2}\big)\right) F(\overrightarrow{\mathbf{W}}^{(t)}) \ . 
\end{aligned} \tag{13.2}$$

Above, ① uses Theorem 4 and $\mathbb{E}_{S_t}[\nabla_t] = \nabla F(\overrightarrow{\mathbf{W}}^{(t)})$; ② uses Theorem 3 which give

$$\begin{aligned}
\mathop{\mathbb{E}}_{S_t}\left[\|\nabla_t\|_2^2\right] &\leq \frac{n^2}{b}\mathop{\mathbb{E}}_{S_t}\left[\sum_{i\in S_t}\max_{\ell\in[L]}\left\|\nabla_{\mathbf{W}_\ell}F_i(\overrightarrow{\mathbf{W}}^{(t)})\right\|_F^2\right] \leq O\Big(\frac{nmF(\overrightarrow{\mathbf{W}}^{(t)})}{d}\Big) \\
\mathop{\mathbb{E}}_{S_t}\left[\|\nabla_t\|_2\right] &\leq \left(\mathop{\mathbb{E}}_{S_t}\left[\|\nabla_t\|_2^2\right]\right)^{1/2} \leq O\Big(\big(\frac{nmF(\overrightarrow{\mathbf{W}}^{(t)})}{d}\big)^{1/2}\Big) \ ;
\end{aligned}$$

③ use gradient lower bound from Theorem 3 and our choice of $\eta$.

At the same time, we also have the following absolute value bound:

$$\begin{aligned}
F(\overrightarrow{\mathbf{W}}^{(t+1)}) &\overset{①}{\leq} F(\overrightarrow{\mathbf{W}}^{(t)}) + \eta\|\nabla F(\overrightarrow{\mathbf{W}}^{(t)})\|_F \cdot \|\nabla_t\|_F \\
&\qquad + \eta\sqrt{nF(\overrightarrow{\mathbf{W}}^{(t)})} \cdot O\left(\frac{\omega^{1/3}L^2\sqrt{m\log m}}{\sqrt{d}}\right) \cdot \|\nabla_t\|_2 + O\Big(\eta^2\frac{nL^2m}{d}\Big) \cdot \|\nabla_t\|_2^2 \\
&\overset{②}{\leq} F(\overrightarrow{\mathbf{W}}^{(t)}) + \eta \cdot O\left(\sqrt{\frac{LF(\overrightarrow{\mathbf{W}}^{(t)})mn}{d}}\right) \cdot O\left(\sqrt{\frac{n^2mLF(\overrightarrow{\mathbf{W}}^{(t)})}{bd}}\right) \\
&\qquad + \eta\sqrt{nF(\overrightarrow{\mathbf{W}}^{(t)})} \cdot O\left(\frac{\omega^{1/3}L^2\sqrt{m\log m}}{\sqrt{d}}\right) \cdot \frac{\sqrt{n^2mF(\overrightarrow{\mathbf{W}}^{(t)})}}{\sqrt{bd}} \\
&\qquad + O\Big(\eta^2\frac{nL^2m}{d}\Big) \cdot \frac{n^2}{b}O\Big(\frac{mF(\overrightarrow{\mathbf{W}}^{(t)})}{d}\Big) \\
&\overset{③}{\leq} \Big(1 + O\Big(\frac{\eta Lmn^{1.5}}{\sqrt{b}d} + \frac{\eta n^{1.5}\omega^{1/3}L^2m\sqrt{\log m}}{\sqrt{b}d} + \frac{\eta^2n^3L^2m^2}{d^2b}\Big)\Big)F(\overrightarrow{\mathbf{W}}^{(t)}) \ .
\end{aligned} \tag{13.3}$$

Above, ① uses Theorem 4 and Cauchy-Schwarz $\langle A, B\rangle \leq \|A\|_F\|B\|_F$, and ② uses Theorem 3 which

give

$$\|\nabla_t\|_2^2 \leq \frac{n^2}{b}\left[\sum_{i\in S_t}\max_{\ell\in[L]}\left\|\nabla_{\mathbf{W}_\ell}F_i(\overrightarrow{\mathbf{W}}^{(t)})\right\|_F^2\right] \leq \frac{n^2}{b}O\Big(\frac{mF(\overrightarrow{\mathbf{W}}^{(t)})}{d}\Big)$$

$$\|\nabla_t\|_F^2 \leq \frac{n^2}{b}\left[\sum_{i\in S_t}\sum_{\ell=1}^{L}\left\|\nabla_{\mathbf{W}_\ell}F_i(\overrightarrow{\mathbf{W}}^{(t)})\right\|_F^2\right] \leq \frac{Ln^2}{b}O\Big(\frac{mF(\overrightarrow{\mathbf{W}}^{(t)})}{d}\Big)$$

and the derivation from (13.2).

Next, taking logarithm on both sides of (13.2) and (13.3), and using Jensen's inequality $\mathbb{E}[\log X] \leq \log \mathbb{E}[X]$, we have

$$\mathbb{E}[\log F(\overrightarrow{\mathbf{W}}^{(t+1)})] \leq \log F(\overrightarrow{\mathbf{W}}^{(t)}) - \Omega\Big(\frac{\eta\delta m}{dn^2}\Big) \quad \text{and} \quad \log F(\overrightarrow{\mathbf{W}}^{(t+1)}) \leq \log F(\overrightarrow{\mathbf{W}}^{(t)}) + O\Big(\frac{\eta L m n^{1.5}}{\sqrt{b}d}\Big)$$

By (one-sided) martingale concentration, we have with probability at least $1-e^{-\Omega(\log^2 m)}$, for every $t=1,2,\dots,T$:

$$\log F(\overrightarrow{\mathbf{W}}^{(t)}) - \mathbb{E}[\log F(\overrightarrow{\mathbf{W}}^{(t)})] \leq \sqrt{t}\cdot O\Big(\frac{\eta L m n^{1.5}}{\sqrt{b}d}\Big)\cdot \log m \enspace .$$

This implies for every $t=1,2,\dots,T$, we have

$$\begin{aligned}
\log F(\overrightarrow{\mathbf{W}}^{(t)}) &\leq \sqrt{t}\cdot O\Big(\frac{\eta L m n^{1.5}}{\sqrt{b}d}\Big)\cdot\log m + \log F(\overrightarrow{\mathbf{W}}^{(0)}) - \Omega\Big(\frac{\eta\delta m}{dn^2}\Big)t \\
&\overset{①}{=} \log F(\overrightarrow{\mathbf{W}}^{(0)}) - \left(\sqrt{\frac{\eta\delta m}{dn^2}}\cdot\Omega(\sqrt{t}) - \sqrt{\frac{dn^2}{\eta\delta m}}\cdot O\Big(\frac{\eta L m n^{1.5}}{\sqrt{b}d}\log m\Big)\right)^2 \\
&\quad + O\Big(\frac{\eta L^2 m n^5}{b\delta d}\log^2 m\Big) \\
&\overset{②}{\leq} \log F(\overrightarrow{\mathbf{W}}^{(0)}) + 1 - \left(\sqrt{\frac{\eta\delta m}{dn^2}}\cdot\Omega(\sqrt{t}) - \sqrt{\frac{dn^2}{\eta\delta m}}\cdot O\Big(\frac{\eta L m n^{1.5}}{\sqrt{b}d}\log m\Big)\right)^2 \\
&\overset{③}{\leq} \log F(\overrightarrow{\mathbf{W}}^{(0)}) + 1 - \mathbb{1}\left[t \geq \Theta\Big(\frac{L^2n^7}{b\delta^2}\log^2 m\Big)\right]\cdot\Omega\left(\frac{\eta\delta m}{dn^2}t\right) \\
&\overset{④}{\leq} \log F(\overrightarrow{\mathbf{W}}^{(0)}) + 1 - \mathbb{1}\left[t \geq \Theta\Big(\frac{L^2n^7}{b\delta^2}\log^2 m\Big)\right]\cdot\Omega\left(\frac{b\delta^2}{L^2n^7\log^2 m}t\right) \enspace .
\end{aligned}$$

Above, in ① we have used $2a\sqrt{t} - b^2t = -(b\sqrt{t}-a/b)^2 + a^2/b^2$; in ② we have used our choice of $\eta$; in ③ we have used $-(a\sqrt{t}-b)^2 \leq -\mathbb{1}[t \geq 2b^2/a^2]\cdot\frac{a^2t}{4}$; and in ④ we have used our choice of $\eta$ again. We can read two things from the above formula:

- If $T \geq \Omega\big(\frac{L^2n^7}{b\delta^2}\log^2 m \log\frac{n\log m}{\varepsilon}\big)$ then we have

$$\log F(\overrightarrow{\mathbf{W}}^{(T)}) \leq \log O(n\log^2 m) - \Omega\Big(\log\frac{n\log^2 m}{\varepsilon}\Big) \leq \log\varepsilon \enspace .$$

  so $F(\overrightarrow{\mathbf{W}}^{(T)}) \leq \varepsilon$.

- Letting $T_0 = \Omega\big(\frac{L^2n^7}{b\delta^2}\log^2 m\big)$, we have

$$\sum_{i=0}^{t-1}\sqrt{F(\overrightarrow{\mathbf{W}}^{(i)})} \leq \sqrt{n\log^2 m}\cdot 2T_0 + \frac{\sqrt{n\log^2 m}}{2}\cdot 2T_0 + \frac{\sqrt{n\log^2 m}}{4}\cdot 2T_0 + \cdots \leq O\big(\sqrt{n\log^2 m}T_0\big)$$

and therefore one can verify that $\|\overrightarrow{\mathbf{W}}^{(t)} - \overrightarrow{\mathbf{W}}^{(0)}\|_F$ is small and (13.1) holds: by Theorem 3,

$$\begin{aligned}\|\mathbf{W}_\ell^{(t)} - \mathbf{W}_\ell^{(0)}\|_F &\leq \sum_{i=0}^{t-1} \Big\|\eta \frac{n}{|S_t|} \sum_{i \in S_t} \nabla_{\mathbf{W}_\ell} F_i(\overrightarrow{\mathbf{W}}^{(t)})\Big\|_F \leq O\left(\eta\sqrt{\frac{n^2 m}{bd}}\right) \cdot \sum_{i=0}^{t-1} \sqrt{F(\overrightarrow{\mathbf{W}}^{(i)})} \\ &\leq O\left(\eta\sqrt{\frac{n^2 m}{bd}}\right) \cdot O(T_0 \sqrt{n} \log m) \leq O\left(\frac{n^{3.5}\sqrt{d}}{\delta\sqrt{bm}} \log m\right) \ . \qquad \square\end{aligned}$$

# 14 Theorem 5: Equivalence to Neural Tangent Kernel

Recall on input $x \in \mathbb{R}^{\mathfrak{d}}$, the network output $y(\overrightarrow{\mathbf{W}}; x) \stackrel{\text{def}}{=} y = \mathbf{B} h_L \in \mathbb{R}^d$ is a function of the weights $\overrightarrow{\mathbf{W}}$. The *neural tangent kernel (NTK)* [32] is usually referred to as the feature space defined by the network gradient at random initialization. In other words, for the $j$-th output dimension,

- the NTK kernel function $K_j^{\mathsf{ntk}}(x, \widetilde{x}) \stackrel{\text{def}}{=} \langle \nabla y_j(\overrightarrow{\mathbf{W}}^{(0)}; x), \nabla y_j(\overrightarrow{\mathbf{W}}^{(0)}; \widetilde{x})\rangle$
- the NTK objective $y_j^{\mathsf{ntk}}(\overrightarrow{\mathbf{W}}'; x) \stackrel{\text{def}}{=} \langle \nabla y_j(\overrightarrow{\mathbf{W}}^{(0)}; x), \overrightarrow{\mathbf{W}}'\rangle$.

We have the following theorem whose proof is subsumed by the proofs of Theorem 3 and 4. We prove it here for completeness' sake.

**Theorem 5.** *Let $\overrightarrow{\mathbf{W}}^{(0)}, \mathbf{A}, \mathbf{B}$ be at random initialization. For every fixed unit vector $x \in \mathbb{R}^{\mathfrak{d}}$, every (small) parameter $\omega \in \big[\Omega(\frac{d^{3/2}}{m^{3/2} L^{3/2} \log^{3/2} m}), O(\frac{1}{L^{4.5} \log^3 m})\big]$, with probability at least $1 - e^{-\Omega(m\omega^{2/3} L)}$ over $\overrightarrow{\mathbf{W}}^{(0)}, \mathbf{A}, \mathbf{B}$, we have for all $\overrightarrow{\mathbf{W}}' \in (\mathbb{R}^{m\times m})^L$ with $\|\overrightarrow{\mathbf{W}}'\|_2 \leq \omega$, for all $j \in [d]$,*

*(a) $\|\nabla y_j(\overrightarrow{\mathbf{W}}^{(0)} + \overrightarrow{\mathbf{W}}'; x) - \nabla y_j^{\mathsf{ntk}}(\overrightarrow{\mathbf{W}}'; x)\|_F \leq O\big(\sqrt{\log m} \cdot \omega^{1/3} L^3\big) \cdot \|\nabla y_j^{\mathsf{ntk}}(\overrightarrow{\mathbf{W}}'; x)\|_F$; and*

*(b) $y_j(\overrightarrow{\mathbf{W}}^{(0)} + \overrightarrow{\mathbf{W}}'; x) = y_j(\overrightarrow{\mathbf{W}}^{(0)}; x) + y_j^{\mathsf{ntk}}(\overrightarrow{\mathbf{W}}'; x) + O\big(\frac{L^3 \omega^{4/3} \sqrt{m \log m}}{\sqrt{d}}\big)$.*

*(c) If $x, \widetilde{x} \in \mathbb{R}^{\mathfrak{d}}$ are two fixed unit vectors, and $\omega \leq O(\frac{1}{L^9 \log^{3/2} m})$, then*

$$\begin{aligned}&\big|\langle \nabla y_j(\overrightarrow{\mathbf{W}}^{(0)} + \overrightarrow{\mathbf{W}}'; x), \nabla y_j(\overrightarrow{\mathbf{W}}^{(0)} + \overrightarrow{\mathbf{W}}'; \widetilde{x})\rangle - K_j^{\mathsf{ntk}}(x, \widetilde{x})\big| \\ &\qquad \leq O\big(\sqrt{\log m} \cdot \omega^{1/3} L^3\big) \cdot \sqrt{K_j^{\mathsf{ntk}}(x, x) K_j^{\mathsf{ntk}}(\widetilde{x}, \widetilde{x})} \ .\end{aligned}$$

*Proof of Theorem 5.* As before we denote by $\mathbf{D}_1^{(0)}, \dots \mathbf{D}_L^{(0)}$ and $h_1^{(0)}, \dots, h_L^{(0)}$ the diagonal sign matrices and forward vectors determined at random initialization $\overrightarrow{\mathbf{W}}^{(0)}$ and by $\mathbf{D}_1, \dots, \mathbf{D}_L$ and $h_1, \dots, h_L$ those determined at $\overrightarrow{\mathbf{W}} \stackrel{\text{def}}{=} \overrightarrow{\mathbf{W}}^{(0)} + \overrightarrow{\mathbf{W}}'$. Recall $\|\mathbf{D}'_\ell\|_0 = \|\mathbf{D}_\ell - \mathbf{D}_\ell^{(0)}\|_0 \leq s \stackrel{\text{def}}{=} O(m\omega^{2/3} L)$ from Lemma 8.2b.

(a) Let $\mathbf{e}_j \in \mathbb{R}^d$ be the $j$-th basis vector and $\ell$ be in $[L]$. We have

$$\begin{aligned}&\nabla_{\mathbf{W}_\ell} y_j^{\mathsf{ntk}}(\overrightarrow{\mathbf{W}}'; x) - \nabla_{\mathbf{W}_\ell} y_j(\overrightarrow{\mathbf{W}}^{(0)} + \overrightarrow{\mathbf{W}}'; x) \\ &= \big(\mathbf{e}_j^\top \mathbf{B} \mathbf{D}_L^{(0)} \mathbf{W}_L^{(0)} \cdots \mathbf{W}_{\ell+1}^{(0)} \mathbf{D}_\ell^{(0)}\big)^\top (h_{\ell-1}^{(0)})^\top - \big(\mathbf{e}_j^\top \mathbf{B} \mathbf{D}_L \mathbf{W}_L \cdots \mathbf{W}_{\ell+1} \mathbf{D}_\ell\big)^\top (h_{\ell-1})^\top\end{aligned}$$

This difference matrix is precisely (10.1) (by setting $n = 1$ and $\mathsf{v} = e_j$). Using the bound (10.2) we have its Frobenius norm is at most $O\left(\sqrt{m \log m / d} \cdot \omega^{1/3} L^2\right)$. On the other hand, one can calculate for every $k \in [m]$,

$$\nabla_{[\mathbf{W}_L]_k} y_j^{\mathsf{ntk}}(\overrightarrow{\mathbf{W}}'; x) = (\mathbf{B}^\top \mathbf{e}_j)_k \cdot h_{L-1}^{(0)} \cdot \mathbb{1}_{\langle [\mathbf{W}_L]_k, h_{L-1}^{(0)}\rangle \geq 0} \ .$$

We already know $\|h^{(0)}_{L-1}\| \geq \Omega(1)$ from Lemma 7.1. Now, regardless of the randomness of $h^{(0)}_{L-1}$, we have $\mathbb{1}_{\langle [\mathbf{W}_L]_k, h^{(0)}_{L-1}\rangle \geq 0} = 1$ with exactly half probability; also, regardless of the randomness of $\overrightarrow{\mathbf{W}}^{(0)}$ and $\mathbf{A}$, we have $(\mathbf{B}^\top \mathbf{e}_j)_k \sim \mathcal{N}(0, \frac{1}{d})$. Therefore, we conclude that with probability at least $1 - e^{-\Omega(m)}$ it satisfies $\|\nabla_{\mathbf{W}_L} y^{\mathsf{ntk}}_j(\overrightarrow{\mathbf{W}}'; x)\|_F^2 \geq \Omega(m/d)$ . Putting the two abounds together we finish the proof.

(b) This statement can be derived from (11.3), (11.5) and (11.6). For completeness' sake, below we provide a direct proof without invoking them. We first calculate that

$$\left| y_j(\overrightarrow{\mathbf{W}}^{(0)} + \overrightarrow{\mathbf{W}}'; x) - \mathbf{e}_j^\top \mathbf{B}\mathbf{D}^{(0)}_L \mathbf{W}_L \cdots \mathbf{D}^{(0)}_1 \mathbf{W}_1 x \right| = \left| \sum_{\ell=1}^L \mathbf{e}_j^\top \mathbf{B}\mathbf{D}^{(0)}_L \mathbf{W}_L \cdots \mathbf{D}^{(0)}_{\ell+1} \mathbf{W}_{\ell+1} \mathbf{D}'_\ell g_{\ell-1} \right|$$

$$\begin{aligned}
&\leq \sum_{\ell=1}^L \Big( \underbrace{\left\| \mathbf{B}\mathbf{D}^{(0)}_L \mathbf{W}^{(0)}_L \cdots \mathbf{D}^{(0)}_{\ell+1} \mathbf{W}^{(0)}_{\ell+1} \mathbf{D}'_\ell \right\|_2}_{\text{Lemma 7.4a}} \\
&\qquad + \underbrace{\left\| \mathbf{B}\mathbf{D}^{(0)}_L \mathbf{W}^{(0)}_L \cdots \mathbf{D}^{(0)}_{\ell+1} \mathbf{W}^{(0)}_{\ell+1} - \mathbf{B}\mathbf{D}^{(0)}_L \mathbf{W}_L \cdots \mathbf{D}^{(0)}_{\ell+1} \mathbf{W}_{\ell+1} \right\|_2}_{\text{Lemma 8.7}} \Big) \cdot \underbrace{\|\mathbf{D}'_\ell g_{\ell-1}\|}_{\text{Lemma 8.2b}} \\
&\leq L \cdot \left( O\big(\frac{\sqrt{s \log m}}{\sqrt{d}}\big) + O\big(\omega L^{1.5} \frac{\sqrt{m}}{\sqrt{d}}\big) \right) \cdot O(\omega L^{3/2}) \leq O\big(\frac{L^3 \omega^{4/3} \sqrt{m \log m}}{\sqrt{d}}\big) \qquad (14.1)
\end{aligned}$$

We next calculate that

$$\begin{aligned}
&\left| \mathbf{e}_j^\top \mathbf{B}\mathbf{D}^{(0)}_L \mathbf{W}_L \cdots \mathbf{D}^{(0)}_1 \mathbf{W}_1 x - y_j(\overrightarrow{\mathbf{W}}^{(0)}) - y^{\mathsf{ntk}}(\overrightarrow{\mathbf{W}}'; x) \right| \\
&= \left| \sum_{\ell=1}^L \mathbf{e}_j^\top \mathbf{B}\mathbf{D}^{(0)}_L \mathbf{W}_L \cdots \mathbf{W}_{\ell+1} \mathbf{D}^{(0)}_\ell \mathbf{W}'_\ell h^{(0)}_{\ell-1} - \mathbf{e}_j^\top \mathbf{B}\mathbf{D}^{(0)}_L \mathbf{W}^{(0)}_L \cdots \mathbf{W}^{(0)}_{\ell+1} \mathbf{D}^{(0)}_\ell \mathbf{W}'_\ell h^{(0)}_{\ell-1} \right| \\
&\leq \sum_{\ell=1}^L \underbrace{\left\| \mathbf{B} \left( \mathbf{D}^{(0)}_L \mathbf{W}_L \cdots \mathbf{W}_{\ell+1} \mathbf{D}^{(0)}_\ell - \mathbf{D}^{(0)}_L \mathbf{W}^{(0)}_L \cdots \mathbf{W}^{(0)}_{\ell+1} \mathbf{D}^{(0)}_\ell \right) \right\|_2}_{\text{Lemma 8.7}} \cdot \|\mathbf{W}'_\ell\|_2 \cdot \underbrace{\|h^{(0)}_{\ell-1}\|}_{\text{Lemma 7.1}} \\
&\leq L \cdot O\big(\omega L^{1.5} \frac{\sqrt{m}}{\sqrt{d}}\big) \cdot \omega \cdot O(1) \leq O\big(\omega^2 L^{2.5} \frac{\sqrt{m}}{\sqrt{d}}\big) \qquad (14.2)
\end{aligned}$$

Putting (14.1) and (14.2) together finishes the proof.

(c) This is a direct corollary of (a). □

# Appendix

## A Extension to Other Loss Functions

For simplicity, in them main body of this paper we have used the $\ell_2$ regression loss. Our results generalize easily to other Lipschitz smooth (but possibly nonconvex) loss functions.

Suppose we are given loss function $f(z;y)$ that takes as input a neural-network output $z \in \mathbb{R}^d$ and a label $y$. Then, our training objective for the $i$-th training sample becomes $F_i(\mathbf{W}) = f(\mathbf{B}h_{i,L}; y_i^*)$. We redefine the loss vector $\mathsf{loss}_i \stackrel{\text{def}}{=} \nabla f(\mathbf{B}h_{i,L}; y_i^*) \in \mathbb{R}^d$ (where the gradient is with respect to $z$). Note that if $f(z;y) = \frac{1}{2}\|z-y\|^2$ is the $\ell_2$ loss, then this notion coincides with Section 2. We assume that $f(z;y)$ is 1-Lipscthiz (upper) smooth with respect to $z$.[19]

All the results in Section 7, 8 and 9 remain unchanged. Section 10 also remains unchanged, except we need to restate Theorem 3 with respect to this new notation:

$$\|\nabla_{\mathbf{W}_\ell} F_i(\overrightarrow{\mathbf{W}})\|_F^2 \le O\Big(\frac{\|\mathsf{loss}_i\|^2}{d} \times m\Big) \qquad \|\nabla_{\mathbf{W}_\ell} F(\overrightarrow{\mathbf{W}})\|_F^2 \le O\Big(\frac{\|\mathsf{loss}\|^2}{d} \times mn\Big)$$

$$\|\nabla_{\mathbf{W}_L} F(\overrightarrow{\mathbf{W}})\|_F^2 \ge \Omega\Big(\frac{\max_{i\in[n]} \|\mathsf{loss}_i\|^2}{dn/\delta} \times m\Big) \ .$$

Section 11 also remains unchanged, except that we need to replace the precise definition of $\ell_2$ loss in (11.2) with the semi-smoothness condition:

$$\begin{aligned} F_i(\overrightarrow{\mathbf{W}}) = f(\mathbf{B}h_{i,L}; y_i^*) &\le f(\mathbf{B}\breve{h}_{i,L}; y_i^*) + \langle \nabla f(\mathbf{B}\breve{h}_{i,L}, y_i^*), \mathbf{B}(h_{i,L} - \breve{h}_{i,L})\rangle + \frac{1}{2}\|\mathbf{B}(h_{i,L} - \breve{h}_{i,L})\|^2 \\ &= F_i(\breve{\overrightarrow{\mathbf{W}}}) + \langle \breve{\mathsf{loss}}_i, \mathbf{B}(h_{i,L} - \breve{h}_{i,L})\rangle + \frac{1}{2}\|\mathbf{B}(h_{i,L} - \breve{h}_{i,L})\|^2 \end{aligned} \tag{A.1}$$

and the rest of the proof remains unchanged.

As for the final convergence theorem of gradient descent, we can replace (12.2) with

$$F(\overrightarrow{\mathbf{W}}^{(t+1)}) \le F(\overrightarrow{\mathbf{W}}^{(t)}) - \Omega\Big(\frac{\eta\delta m}{dn^2}\Big) \cdot \|\mathsf{loss}^{(t)}\|^2 \ . \tag{A.2}$$

This means many things:

- If the loss is nonconvex but satisfies the Polyak-Łojasiewicz condition $\|\nabla f(z;y)\|^2 \ge \sigma(f(z;y) - f(z^*;y))$, then in $T = \Omega\Big(\frac{dn^2}{\eta\delta m\sigma}\Big) = O\big(\frac{n^6 L^2}{\delta^2\sigma}\log\frac{1}{\varepsilon}\big)$ iterations, GD can find a point $\overrightarrow{\mathbf{W}}^{(T)}$ with $\|\mathsf{loss}^{(T)}\| \le \varepsilon$. It suffices to choose $m \ge \widetilde{\Omega}\big(\mathsf{poly}(n, L, \delta^{-1}) \cdot d\sigma^{-2}\big)$ for same reason as before.
- If the loss is nonconvex but bounded (say, $|f(z;y)| \le O(1)$), then in $T = O\Big(\frac{dn^2}{\eta\delta m\varepsilon^2}\Big) = O\big(\frac{n^6 L^2}{\delta^2\varepsilon^2}\big)$ iterations, we can find a point $\overrightarrow{\mathbf{W}}^{(T)}$ with $\|\mathsf{loss}^{(T)}\| \le \varepsilon$. If suffices to choose $m \ge \widetilde{\Omega}\big(\mathsf{poly}(n, L, \delta^{-1}) \cdot d\varepsilon^{-1}\big)$.)
- If the loss is convex and its minimizer has bounded norm, meaning there exists $z^*$ so that $f(z^*;y) = \min_z f(z;y)$ and $\|z - z^*\| \le D$. Then, by convexity

$$f(z;y) - f(z^*;y) \le \langle \nabla f(z;y), z - z^*\rangle \le D\|\nabla f(z;y)\|$$

[19] That is, $f(z+z';y) \le f(z) + \langle \nabla f(z;y), z'\rangle + \frac{1}{2}\|z'\|^2$.

Putting this into (A.2), we have (here $\overrightarrow{\mathbf{W}}^* = \arg\min_{\overrightarrow{\mathbf{W}}} F_i(\overrightarrow{\mathbf{W}})$ for all $i \in [n]$)

$$F(\overrightarrow{\mathbf{W}}^{(t+1)}) - F(\overrightarrow{\mathbf{W}}^*) \le F(\overrightarrow{\mathbf{W}}^{(t)}) - F(\overrightarrow{\mathbf{W}}^*) - \Omega\Big(\frac{\eta\delta m}{dn^2 D^2}\Big) \cdot \sum_{i\in[n]} \big(F_i(\overrightarrow{\mathbf{W}}^{(t)}) - F_i(\overrightarrow{\mathbf{W}}^*)\big)^2$$
$$\le F(\overrightarrow{\mathbf{W}}^{(t)}) - F(\overrightarrow{\mathbf{W}}^*) - \Omega\Big(\frac{\eta\delta m}{dn^3 D^2}\Big) \cdot \big(F(\overrightarrow{\mathbf{W}}^{(t)}) - F(\overrightarrow{\mathbf{W}}^*)\big)^2 \enspace .$$

This implies (see for instance the classical calculation steps in [42]) that after $T = O\Big(\frac{dn^3 D^2}{\eta\delta m\varepsilon}\Big) = O\big(\frac{n^7 L^2 D^2}{\delta^2 \varepsilon}\big)$ iterations, we can have $F(\overrightarrow{\mathbf{W}}^{(T)}) - F(\overrightarrow{\mathbf{W}}^*) \le \varepsilon$. The amount of over-parameterization needed is $m \ge \widetilde{\Omega}\big(\mathsf{poly}(n, L, \delta^{-1}) \cdot d \log \varepsilon^{-1}\big)$.

- If the loss is cross entropy $f(z; y) = \frac{e^{z_y}}{\sum_{i=1}^d e^{z_i}}$ for classification, then $\|\nabla f(z;y)\| < 1/4$ implies perfect classification.[20] Thus, we have 100% training accuracy in $T = O\big(\frac{n^6 L^2}{\delta^2}\big)$ iterations. If suffices to choose $m \ge \widetilde{\Omega}\big(\mathsf{poly}(n, L, \delta^{-1}) \cdot d\big)$

# B Extension to Convolutional Neural Networks

There are numerous versions of convolutional neural networks (CNNs) that are used in practice. To demonstrate the capability of applying our techniques to such convolutional settings, in this section, we study a simple enough CNN for the $\ell_2$ regression task.

**A Simple CNN Model.** We assume that for the input layer (corresponding to $\mathbf{A}$) and for each hidden layer $\ell = 1, 2, \dots, L-1$ (corresponding to $\mathbf{W}_1, \dots, \mathbf{W}_{L-1}$), there are $\mathfrak{d}$ positions each consisting of $m$ channels. (Each position can be thought as a pixel of an image in computer vision tasks.) We assume the last hidden layer $\ell = L$ (corresponding to $\mathbf{W}_L$) and the output layer (corresponding to $\mathbf{B}$) are fully connected. We assume for each $j \in [\mathfrak{d}]$, there exists a set $\mathcal{Q}_j \subseteq [\mathfrak{d}]$ of fixed cardinality $q \in [\mathfrak{d}]$ so that the value at position $j$ in any convolutional layer is completely determined by positions $k \in \mathcal{Q}_j$ of the previous layer.

**Assumption B.1.** *We assume that $(\mathcal{Q}_1, \dots, \mathcal{Q}_{\mathfrak{d}})$ give rise to a $q$-regular bipartite graph: each $\mathcal{Q}_j$ has exactly $q$ entries and each $k \in [\mathfrak{d}]$ appears in exactly $q$ different sets $\mathcal{Q}_j$.*
*(In vision tasks, if $3\times 3$ kernels are used then $|\mathcal{Q}_j| = 9$. We ignore the padding issue for simplicity.)*

The output of each convolutional layer $\ell = 0, 1, 2, \dots, L-1$ is represented by a $\mathfrak{d}m$-dimensional vector $h_\ell = (h_{\ell,1}, \dots, h_{\ell,\mathfrak{d}})$ where each $h_{\ell,j} \in \mathbb{R}^m, \forall j \in [\mathfrak{d}]$. In the input layer and each $j \in [\mathfrak{d}]$, we assume

$$h_{0,j} = \phi\big(\mathbf{A}_j x_{\mathcal{Q}_j}\big) \in \mathbb{R}^m$$

where $x_{\mathcal{Q}_j} \in \mathbb{R}^q$ denotes the concatenation of $x_k$ for all $k \in \mathcal{Q}_j$ given input $x \in \mathbb{R}^{\mathfrak{d}}$, and $\mathbf{A}_j \in \mathbb{R}^{m\times q}$ is randomly initialized at $\mathcal{N}(0, \frac{2}{\sqrt{qm}})$ per entry. For notational simplicity, we define matrix $\mathbf{A} \in \mathbb{R}^{\mathfrak{d}m\times\mathfrak{d}}$ so that it satisfies $h_1 = \phi\big(\mathbf{A}x\big)$. Each row of $\mathbf{A}$ has $q$ non-zero entries.

For each layer $\ell = 1, \dots, L-1$ and each $j \in [\mathfrak{d}]$, we assume

$$h_{\ell,j} = \phi\big(\mathbf{W}_{\ell,j} h_{\ell-1,\mathcal{Q}_j} + \tau \cdot \mathbf{b}_{\ell,j}\big) \in \mathbb{R}^m$$

where $h_{\ell-1,\mathcal{Q}_j} \in \mathbb{R}^{qm}$ denotes the concatenation of $h_{\ell-1,k}$ for all $k \in \mathcal{Q}_j$, the weights $\mathbf{W}_{\ell,j} \in \mathbb{R}^{m\times(qm)}$ and the bias the $\mathbf{b}_{\ell,j} \in \mathbb{R}^m$ are randomly initialized at $\mathcal{N}(0, \frac{2}{qm})$ per entry, and $\tau$ is a small parameter

[20]Recall $\frac{\partial f(z;y)}{z_y} = p_y(1-p_y)$ where $p_j = \frac{e^{z_j}}{\sum_{i=1}^d e^{z_i}}$. If $p_y > 1/2$, then $z$ correctly predicts the target label $y$ because $p_y > p_j$ for $j \neq z$.

(say, $\tau = \frac{\delta^2}{100\mathfrak{d}L}$) for bias. For notational simplicity, we define matrix $\mathbf{W}_\ell \in \mathbb{R}^{\mathfrak{d}m\times\mathfrak{d}m}$ and vector $\mathbf{b}_\ell \in \mathbb{R}^{\mathfrak{d}m}$ so that it satisfies $h_\ell = \phi\big(\mathbf{W}_\ell h_{\ell-1} + \tau\mathbf{b}_\ell\big)$, and define vector $g_\ell \stackrel{\text{def}}{=} \mathbf{W}_\ell h_{\ell-1} + \tau\mathbf{b}_\ell \in \mathbb{R}^{\mathfrak{d}m}$. Note that each row of $\mathbf{W}_\ell$ has $qm$ non-zero entries.

We assume the last layer $\mathbf{W}_L$ and the output layer $\mathbf{B}$ are simply fully connected (say without bias). That is, each entry of $\mathbf{W}_L \in \mathbb{R}^{\mathfrak{d}m\times\mathfrak{d}m}$ is from $\mathcal{N}(0, \frac{2}{qm})$, and of $\mathbf{B} \in \mathbb{R}^{d\times\mathfrak{d}m}$ is from $\mathcal{N}(0, \frac{1}{d})$.

We denote by $h_{i,\ell}$ the value of $h_\ell$ when the input vector is $x_i$, and define $g_{i,\ell}$, $\mathbf{D}_{i,\ell}$ in the same way as before.

## B.1 Changes in the Proofs

If one is willing to loose polynomial factors in $L$ and $\mathfrak{d}$ in the final complexity, then changes to each of the lemmas of this paper is very little.[21]

**Changes to Section 7.** The first main result is Lemma 7.1: $\|h_{i,\ell}\|$ is in $[1-\varepsilon, 1+\varepsilon]$ with high probability. In the CNN case, for every $j \in [\mathfrak{d}]$, recalling that $h_{i,\ell,j} = \phi(\mathbf{W}_{\ell,j} h_{i,\ell-1,\mathcal{Q}_j} + \tau\mathbf{b}_\ell)$. Applying Fact 7.2, we have that $\frac{qm\|h_{i,\ell,j}\|^2}{2(\|h_{i,\ell-1,\mathcal{Q}_j}\|^2+\tau^2)}$ is distributed as $\chi^2_\omega$ distribution with $\omega \sim \mathcal{B}(m, \frac{1}{2})$. Due to the concentration of $\chi^2$ distribution and the concentration of binomial distribution, $\|h_{i,\ell,j}\|^2$ is extremely close to $\frac{\|h_{i,\ell-1,\mathcal{Q}_j}\|^2+\tau^2}{q}$ (a careful argument of this can be found in the proof of Lemma 7.1). Summing this up over all $j \in [\mathfrak{d}]$, and using Assumption B.1, we have $\|h_{i,\ell,j}\|^2$ is concentrated at $\|h_{i,\ell-1}\|^2 + \frac{\tau^2\mathfrak{d}}{q}$. Applying induction, we have $\|h_{i,\ell}\|$ is in $[1-\varepsilon, 1+\varepsilon]$ with probability at least $1 - e^{-\Omega(m\varepsilon^2/L^2)}$, as long as $\tau^2 \leq \frac{\varepsilon q}{100\mathfrak{d}L}$.[22]

The changes to Lemma 7.3 and Lemma 7.4 are the same as above, but we loose some polynomial factors in $L$ (because we are not careful in the argument above). For instance, the intermediate bound in Lemma 7.3a becomes $\|\mathbf{W}_b\mathbf{D}_{i,b-1}\mathbf{W}_{b-1}\cdots\mathbf{D}_{i,a}\mathbf{W}_a\|_2 \leq O(L)$.

As for the $\delta$-separateness Lemma 7.5, we need to redefine the notion of $\delta$-separateness between $h_{i,\ell}$ and $h_{j,\ell}$:

$$\sum_{k\in[\mathfrak{d}]} \Big\|(\mathbf{I} - \tfrac{h_{i,\ell,k}h_{i,\ell,k}^\top}{\|h_{i,\ell,k}\|^2})h_{j,\ell,k}\Big\|^2 \geq \Omega(\delta^2) \tag{B.1}$$

Then, denoting by $\widehat{h}_k = h_{i,\ell-1,k}/\|h_{i,\ell-1,k}\|$, we have

$$h_{j,\ell,k} = \phi(\mathbf{W}_{\ell,k}h_{j,\ell-1,\mathcal{Q}_j} + \tau\mathbf{b}_{\ell,k}) = \phi\Big(\vec{g}_1 + \Big(\sum_{z\in\mathcal{Q}_k}\|(\mathbf{I} - \widehat{h}_z\widehat{h}_z^\top)h_{j,\ell-1,z}\|^2\Big)^{1/2}\vec{g}_2\Big)$$

where $\vec{g}_2 \sim \mathcal{N}(0, \frac{2}{qm}\mathbf{I})$ is independent of the randomness of $h_{i,\ell,k}$ once $\mathbf{A}, \mathbf{W}_1, \dots, \mathbf{W}_{\ell-1}$ are fixed. One can use this to replace (7.5) and the rest of the proof follows.

**Changes to Section 8.** The first main result is Lemma 8.2, and we discuss necessary changes here to make it work for CNN. The first change in the proof is to replace $2c_1L^{1.5}$ with $2c_1L^2$ due to the above additional factor from Lemma 7.3a. Next, call that the proof of Lemma 8.2 relied on Claim 8.3 and Claim 8.5:

- For Claim 8.3, we can replace the definition of $x$ with $x = \mathbf{D}'(\mathbf{W}^{(0)}h^{(0)} + \tau\mathbf{b} + g')$ for $\mathbf{b} \in \mathcal{N}(0, \frac{2}{qm}\mathbf{I})$. This time, instead of using the randomness of $\mathbf{W}^{(0)}$ like in the old proof (because

[21] We acknowledge the existence of more careful modifications to avoid loosing too many such factors, but do not present such result for the simplicity of this paper.

[22] We note that in all of our applications of Lemma 7.1, the minimal choice of $\varepsilon$ is around $\delta^3$ from the proof of $\delta$-separateness. Therefore, choosing $\tau = \frac{\delta^2}{100\mathfrak{d}L}$ is safe. We are aware of slightly more involved proofs that are capable of handling much larger values of $\tau$.

$\mathbf{W}^{(0)}$ is no longer a full matrix), we use the randomness of $\tau\mathbf{b}$. The new statement becomes

$$\|x\|_0 \le O\Big(\frac{\mathfrak{d}m}{\tau^{2/3}}\|g_1'\|^{2/3} + \frac{1}{\tau}\|g_2'\|_\infty(\mathfrak{d}m)^{3/2}\Big) \quad \text{and} \quad \|x\| \le O\Big(\|g_1'\| + \frac{1}{\sqrt{\tau}}\|g_2'\|_\infty^{3/2}(\mathfrak{d}m)^{3/4}\Big) \ .$$

and its proof is by re-scaling $x$ by $\frac{1}{\tau}$ and then applying the old proof (with dimension $m$ replaced with $\mathfrak{d}m$).

- For Claim 8.5, it becomes $\|y_1\| \le O\big(\sqrt{qs/m}\log m\big)$ and $\|y_2\|_\infty \le \frac{2\sqrt{\log m}}{\sqrt{qm}}$ .

After making all of these changes, we loose at most some polynomial factors in $L$ and $\mathfrak{d}$ for the new statement of Lemma 8.2:

(a) $\|\mathbf{D}'_{i,\ell}\|_0 \le m\omega^{2/3}\mathsf{poly}(L,\mathfrak{d})$ and $\|\mathbf{D}'_{i,\ell}g_{i,\ell}\| \le \omega\mathsf{poly}(L,\mathfrak{d})$.

(b) $\|g'_{i,\ell}\|, \|h'_{i,\ell}\| \le \omega\mathsf{poly}(L,\mathfrak{d})\sqrt{\log m}$.

Finally, the statements of Lemma 8.6 and Lemma 8.7 only loose polynomial factors in $L$ and $\mathfrak{d}$.

**Changes to Section 9.** The norm upper bound part is trivial to modify so we only focus on the gradient norm lower bound. Since we have assumed $\mathbf{W}_L$ to be fully connected, the gradient on $\mathbf{W}_L$ is the same as before:

$$\widehat{\nabla}^{\vec{\mathsf{v}}}_{[\mathbf{W}_L]_k} F(\overrightarrow{\mathbf{W}}) = \sum_{i=1}^n \langle \mathbf{B}_k, \mathsf{v}_i\rangle \cdot h_{i,L-1} \cdot \mathbb{1}_{(\mathbf{W}_L h_{i,L-1})_k \ge 0}$$

Since we still have $\delta$-separateness (B.1), one can verify for $\ell = L-1$,

$$\|h_{i,\ell} - h_{j,\ell}\|^2 = \sum_{k\in[\mathfrak{d}]}\|h_{i,\ell,k} - h_{j,\ell,k}\|^2 \ge \sum_{k\in[\mathfrak{d}]}\Big\|(\mathbf{I} - \frac{h_{i,\ell,k}h_{i,\ell,k}^\top}{\|h_{i,\ell,k}\|^2})h_{j,\ell,k}\Big\|^2 \ge \Omega(\delta^2) \ .$$

Since $\|h_{i,\ell}\| \approx 1$ and $\|h_{j,\ell}\| \approx 1$, this gives back the old definition of $\delta$-separateness: $(\mathbf{I} - h_{i,\ell}h_{i,\ell}^\top/\|h_{i,\ell}\|^2)h_{j,\ell}$ has norm at least $\Omega(\delta)$. Therefore, the entire rest of Section 9 follows as before.

**Final Theorem.** Since Section 10 and 11 rely on previous sections, they do not need to be changed (besides some polynomial factor blowup in $L$ and $\mathfrak{d}$). Our final theorem becomes

**Theorem 7** (CNN)**.** *Let $m \ge \widetilde{\Omega}\big(\mathsf{poly}(n, L, \mathfrak{d}, \delta^{-1})\cdot d\big)$. For the convolutional neural network defined in this section, with probability at least $1 - e^{-\Omega(\log^2 m)}$ over the random initialization, GD and SGD respectively need at most $T = \frac{\mathsf{poly}(n,L,\mathfrak{d})}{\delta^2}\log\frac{1}{\varepsilon}$ and $T = \frac{\mathsf{poly}(n,L,\mathfrak{d})\cdot\log^2 m}{\delta^2}\log\frac{1}{\varepsilon}$ iterations to find a point $F(\overrightarrow{\mathbf{W}}) \le \varepsilon$.*

# C Extension to Residual Neural Networks

Again as we have discussed in Section C, there are numerous versions of residual neural networks that are used in practice. To demonstrate the capability of applying our techniques to residual settings, in this section, we study a simple enough residual network for the $\ell_2$ regression task (without convolutional layers).

**A Simple Residual Model.** We consider an input layer $h_0 = \phi(\mathbf{A}x)$, $L-1$ residual layers $h_\ell = \phi(h_{\ell-1} + \tau\mathbf{W}_\ell h_{\ell-1})$ for $\ell = 1, 2, \dots, L-1$, a fully-connected layer $h_L = \phi(\mathbf{W}_L h_{L-1})$ and an output layer $y = \mathbf{B}h_L$. We assume that $h_0, \dots, h_L \in \mathbb{R}^m$ and the entries of $\mathbf{W}_\ell \in \mathbb{R}^{m\times m}$ are from $\mathcal{N}(0, \frac{2}{m})$ as before. We choose $\tau = \frac{1}{\Omega(L\log m)}$ which is similar as previous work [61].

We denote by $g_0 = \mathbf{A}x$, $g_\ell = h_{\ell-1} + \tau \mathbf{W}_\ell h_{\ell-1}$ for $\ell = 1, 2, \dots, L-1$ and $g_L = \mathbf{W}_L h_{L-1}$. For analysis, we use $h_{i,\ell}$ and $g_{i,\ell}$ to denote the value of $h_\ell$ when the input vector is $x_i$, and $\mathbf{D}_{i,\ell}$ the diagonal sign matrix so that $[\mathbf{D}_{i,\ell}]_{k,k} = \mathbb{1}_{(g_{i,\ell})_k \ge 0}$.

## C.1 Changes in the Proofs

Conceptually, we need to replace all the occurrences of $\mathbf{W}_\ell$ with $(\mathbf{I} + \tau \mathbf{W}_\ell)$ for $\ell = 1, 2, \dots, L-1$. Many of the proofs in the residual setting becomes much simpler when residual links are present. The main property we shall use is that the spectral norm

$$\|(\mathbf{I} + \tau \mathbf{W}_a) \mathbf{D}_{i,a+1} \cdots \mathbf{D}_{i,b} (\mathbf{I} + \tau \mathbf{W}_b)\|_2 \le 1.01 \tag{C.1}$$

for any $L - 1 \ge a \ge b \ge 1$ with our choice of $\tau$.

**Changes to Section 7.** For Lemma 7.1, ignoring subscripts in $i$ for simplicity, we can combine the old proof with (C.1) to derive that $\|h_\ell\| \le 1.02$ for every $i$ and $\ell$. We also have $\|h_\ell\| \ge \frac{1}{\sqrt{20}}$ by the following argument.

- Fact 7.2 says each coordinate of $h_0$ follows i.i.d. from a distribution which is 0 with half probability, and $|\mathcal{N}(0, \frac{2}{m})|$ with half probability. Therefore, with high probability, at least $m/4$ of the coordinates $k \in [m]$ will satisfy $|(h_0)_k| \ge \frac{0.6}{\sqrt{m}}$. Denote this set as $M_0 \subseteq [m]$.
- In the following layer $\ell = 1$, $(h_\ell)_k \ge (h_{\ell-1})_k - \tau |(\mathbf{W}_\ell h_{\ell-1})_k|$. Since $\mathbf{W}_\ell h_{\ell-1} \sim \mathcal{N}(0, \frac{2\|h_{\ell-1}\|^2}{m} \mathbf{I})$ and $\|h_{\ell-1}\| \le 1.02$, we know with high probability, at least $1 - \frac{1}{10L}$ fraction of the coordinates in $M_0$ will satisfy $|(\mathbf{W}_\ell h_{\ell-1})_k| \le O(\frac{\log L}{\sqrt{m}})$. Therefore, for each of these $(1 - \frac{1}{10L})|M_0|$ coordinates, we have $(h_\ell)_k \ge (h_{\ell-1})_k - \frac{1}{10L}$ by our choice of $\tau$. Denote this set as $M_1 \subseteq M_0$, then we have $(h_\ell)_k \ge \frac{0.6}{\sqrt{m}} - \frac{1}{10L\sqrt{m}}$ for each $k \in M_1$.
- Continuing this argument for $\ell = 2, 3, \dots, L-1$, we know that every time we move from $M_{\ell-1}$ to $M_\ell$, its size shrinks by a factor $1 - \frac{1}{10L}$, and the magnitude of $(h_\ell)_k$ for $k \in M_\ell$ decreases by $\frac{1}{10L\sqrt{m}}$. Putting this together, we know $\|h_\ell\|^2 \ge (\frac{0.6}{\sqrt{m}} - \frac{1}{10\sqrt{m}})^2 \cdot (1 - \frac{1}{10L})^L \cdot \frac{m}{4} \ge \frac{1}{20}$ for all $\ell = 1, 2, \dots, L-1$. The proof of the last layer $h_L$ is the same as the old proof.

Lemma 7.3 is not needed anymore because of (C.1). Lemma 7.4 becomes trivial to prove using (C.1): for instance for Lemma 7.4a, we have $\|\mathbf{D}_{i,L} \mathbf{W}_L \mathbf{D}_{i,L-1} (\mathbf{I} + \tau \mathbf{W}_{L-1}) \cdots \mathbf{D}_{i,a} (\mathbf{I} + \tau \mathbf{W}_a) u\| \le O(\|u\|)$ and thus $\|\mathbf{B} \mathbf{D}_{i,L} \mathbf{W}_L \mathbf{D}_{i,L-1} (\mathbf{I} + \tau \mathbf{W}_{L-1}) \cdots \mathbf{D}_{i,a} (\mathbf{I} + \tau \mathbf{W}_a) u\| \le O\big(\frac{\sqrt{s \log m}}{\sqrt{d}}\big) \|u\|$ for all $s$-sparse vectors $u$.

Lemma 7.5 needs the following changes in the same spirit as our changes to Lemma 7.1. With probability at least $1 - e^{-\Omega(\log^2 m)}$ it satisfies $\|\mathbf{W}_\ell h_{i,\ell}\|_\infty \le O(\frac{\log m}{\sqrt{m}})$ for all $i \in [n]$ and $\ell \in L$. In the following proof we condition on this event happens.[23] Consider $i, j \in [n]$ with $i \ne j$.

- In the input layer, since $\|x_i - x_j\| \ge \delta$, the same Claim 7.6 shows that, with high probability, there are at least $\frac{3}{4} m$ coordinates $k \in [m]$ with $|(h_{i,0} - h_{j,0})_k| \ge \frac{\delta}{10\sqrt{m}}$. At the same time, at least $\frac{3}{4} m$ coordinates $k \in [m]$ will satisfy $(h_{i,0})_k \ge \frac{1}{10\sqrt{m}}$ and $(h_{j,0})_k \ge \frac{1}{10\sqrt{m}}$. Denote $M_0 \subseteq [m]$ as the set of coordinates $k$ satisfying both properties. We have $|M_0| \ge \frac{m}{2}$ and $\sum_{k \in M_0} |(h_{i,0} - h_{j,0})_k| \ge \frac{\delta}{20} \sqrt{m}$.
- In the following layer $\ell = 1$, we have

$$(h_{i,\ell} - h_{j,\ell})_k = \phi((h_{i,\ell-1})_k + \tau (\mathbf{W}_\ell h_{i,\ell-1})_k) - \phi((h_{j,\ell-1})_k + \tau (\mathbf{W}_\ell h_{j,\ell-1})_k)$$

[23] For simplicity, we only show how to modify Lemma 7.5 with success probability $1 - e^{-\Omega(\log^2 m)}$ because that is all we need to the downstream application of Lemma 7.5. If one is willing to be more careful, the success probability can be much higher.

Using $\|\mathbf{W}_\ell h_{i,\ell}\|_\infty \le O(\frac{\log m}{\sqrt{m}})$ and our choice of $\tau$, we know for every $k \in M_0$, it satisfies $(h_{i,\ell})_k \ge \frac{1}{10\sqrt{m}} - \frac{1}{100L\sqrt{m}}$ and $(h_{j,\ell})_k \ge \frac{1}{10\sqrt{m}} - \frac{1}{100L\sqrt{m}}$. Therefore, the ReLU activation becomes identity for such coordinates $k \in M_0$ and

$$\Delta_k \stackrel{\text{def}}{=} (h_{i,\ell} - h_{j,\ell})_k = (h_{i,\ell-1} - h_{j,\ell-1})_k + \tau(\mathbf{W}_\ell(h_{i,\ell-1} - h_{j,\ell-1}))_k \ .$$

Let $s_k = 1$ if $(h_{i,\ell-1} - h_{j,\ell-1})_k \ge 0$ and $s_k = -1$ otherwise. Then,

$$\sum_{k\in M_0} |\Delta_k| \ge \sum_{k\in M_0} s_k \cdot \Delta_k = \sum_{k\in M_0} |(h_{i,\ell-1} - h_{j,\ell-1})_k| + \tau \cdot s_k(\mathbf{W}_\ell(h_{i,\ell-1} - h_{j,\ell-1}))_k$$

Note that when $h_{i,\ell-1}$ and $h_{j,\ell-1}$ are fixed, the values $s_k(\mathbf{W}_\ell(h_{i,\ell-1} - h_{j,\ell-1}))_k$ are independent Gaussian with mean zero. This means, with probability at least $1 - e^{-\Omega(\log^2 m)}$, the summation $\sum_{k\in M_0} s_k(\mathbf{W}_\ell(h_{i,\ell-1} - h_{j,\ell-1}))_k$ is at most $O(\log m)$ in absolute value. Putting this into the above equation, we have

$$\sum_{k\in M_0} |\Delta_k| \ge \sum_{k\in M_0} |(h_{i,\ell-1} - h_{j,\ell-1})_k| - O(\tau \log m) \ge \frac{\delta}{20}\sqrt{m} - O(\tau \log m) \ .$$

- Continuing this process for $\ell = 2, 3, \dots, L-1$, we can conclude that $\sum_{k\in M_0} |(h_{i,L-1} - h_{j,L-1})_k| \ge \frac{\delta}{30}\sqrt{m}$ and therefore $\|h_{i,L-1} - h_{j,L-1}\| \ge \Omega(\delta^2)$. This is the same statement as before that we shall need for the downstream application of Lemma 7.5.

**Changes to Section 8.** Lemma 8.2 becomes easy to prove with all the $L$ factors disappear for the following reason. Fixing $i$ and ignoring the subscript in $i$, we have for $\ell = 1, 2, \dots, L-1$:

$$\begin{aligned} h'_\ell &= \mathbf{D}''_\ell\big((\mathbf{I} + \tau\mathbf{W}_\ell + \tau\mathbf{W}'_\ell)h_{\ell-1} - (\mathbf{I} + \tau\mathbf{W}_\ell)h^{(0)}_{\ell-1}\big) \\ &= \mathbf{D}''_\ell\big((\mathbf{I} + \tau\mathbf{W}_\ell)h'_{\ell-1} + \tau\mathbf{W}'_\ell h_{\ell-1}\big) \end{aligned}$$

For some diagonal matrix $\mathbf{D}''_\ell \in \mathbb{R}^{m\times m}$ with diagonal entries in $[-1, 1]$ (see Proposition 11.3). By simple spectral norm of matrices bound we have

$$\|h'_\ell\| \le (1+\tau\|\mathbf{W}_\ell\|_2+\tau\|\mathbf{W}'_\ell\|_2)\|h'_{\ell-1}\|+\tau\|\mathbf{W}'_\ell\|_2\|h^{(0)}_{\ell-1}\| \le (1+\frac{1}{10L})\|h'_{\ell-1}\|+O(\tau\omega) \le \cdots \le O(\tau\omega)$$

This implies $\|h'_\ell\|, \|g'_\ell\| \le O(\tau\omega)$ for all $\ell \in [L-1]$, and combining with the old proof we have $\|h'_L\|, \|g'_L\| \le O(\omega)$.

As for the sparsity $\|\mathbf{D}'_\ell\|_0$, because $g^{(0)}_\ell = h^{(0)}_{\ell-1} + \tau\mathbf{W}^{(0)}_\ell h^{(0)}_{\ell-1} \sim \mathcal{N}\big(h^{(0)}_{\ell-1}, \frac{2\tau^2\|h^{(0)}_{\ell-1}\|^2}{m}\big)$ and $\|g'_\ell\| \le O(\tau\omega)$, applying essentially the same Claim 8.3, we have $\|\mathbf{D}'_\ell\|_0 \le O(m\omega^{2/3})$ for every $\ell = 1, 2, \dots, L-1$. One can similarly argue that $\|\mathbf{D}'_L\|_0 \le O(m\omega^{2/3})$.

Next, Lemma 8.6 and Lemma 8.7 become trivial to prove (recall we have to change $\mathbf{W}^{(0)}_\ell$ with $\mathbf{I} + \tau\mathbf{W}^{(0)}_\ell$ for $\ell < L$) and the $L$ factor also gets improved.

**Changes to Section 9.** The proofs of this section require only notational changes.

**Final Theorem.** Since Section 10 and 11 rely on previous sections, they do not need to be changed (besides improving polynomial factors in $L$). Our final theorem becomes

**Theorem 8** (ResNet). *Let $m \ge \widetilde{\Omega}\big(\mathsf{poly}(n, L, \delta^{-1}) \cdot d\big)$. For the residual neural network defined in this section, with probability at least $1 - e^{-\Omega(\log^2 m)}$ over the random initialization, GD needs at most $T = O\big(\frac{n^6 L^2}{\delta^2}\log\frac{1}{\varepsilon}\big)$ iterations and SGD needs at most $T = O\big(\frac{n^7 L^2 \log^2 m}{b\delta^2}\log\frac{1}{\varepsilon}\big)$ iterations to find a point $F(\overrightarrow{\mathbf{W}}) \le \varepsilon$.*

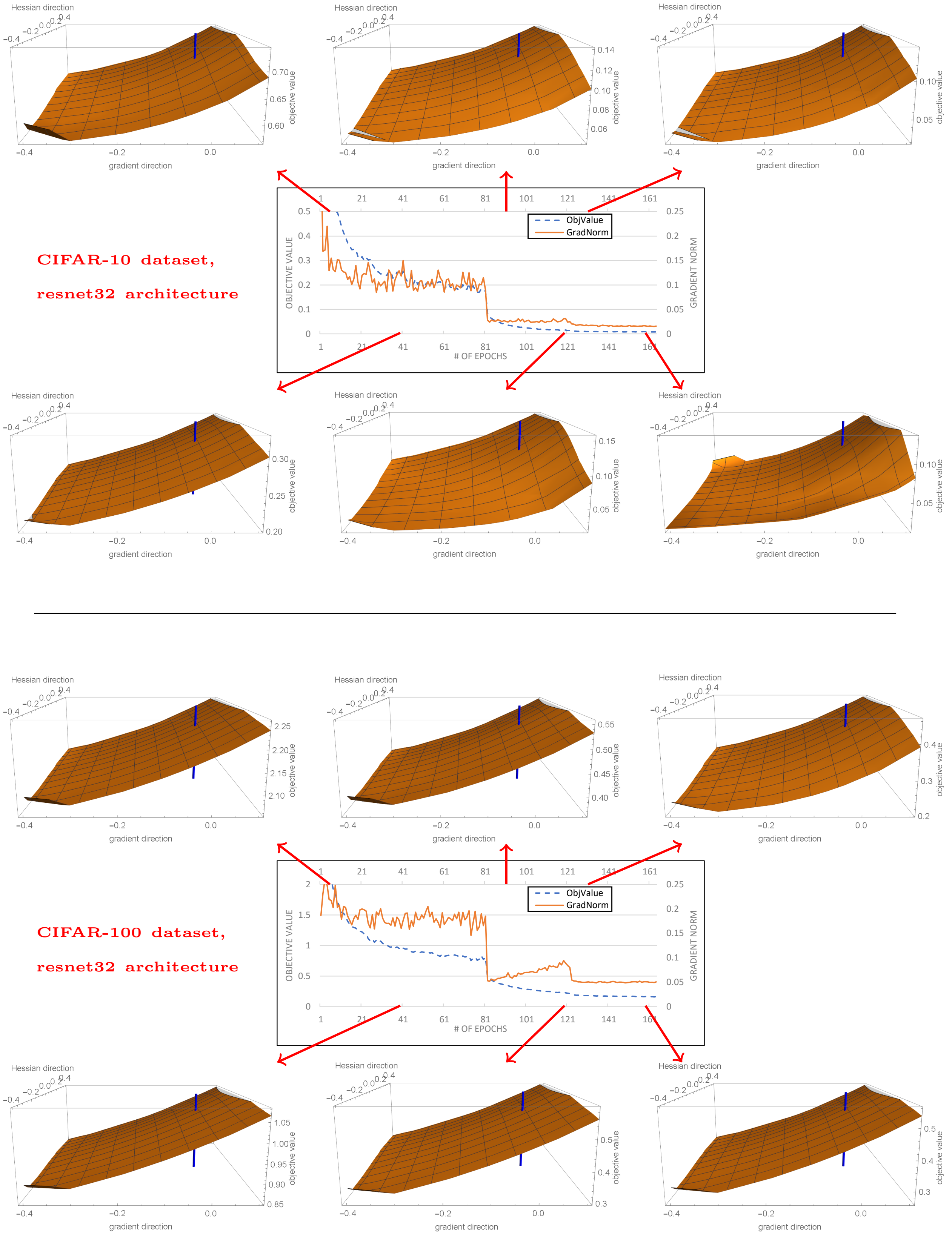

Figure 2: ResNet-32 architecture [58] landscape on CIFAR10 vs CIFAR100.

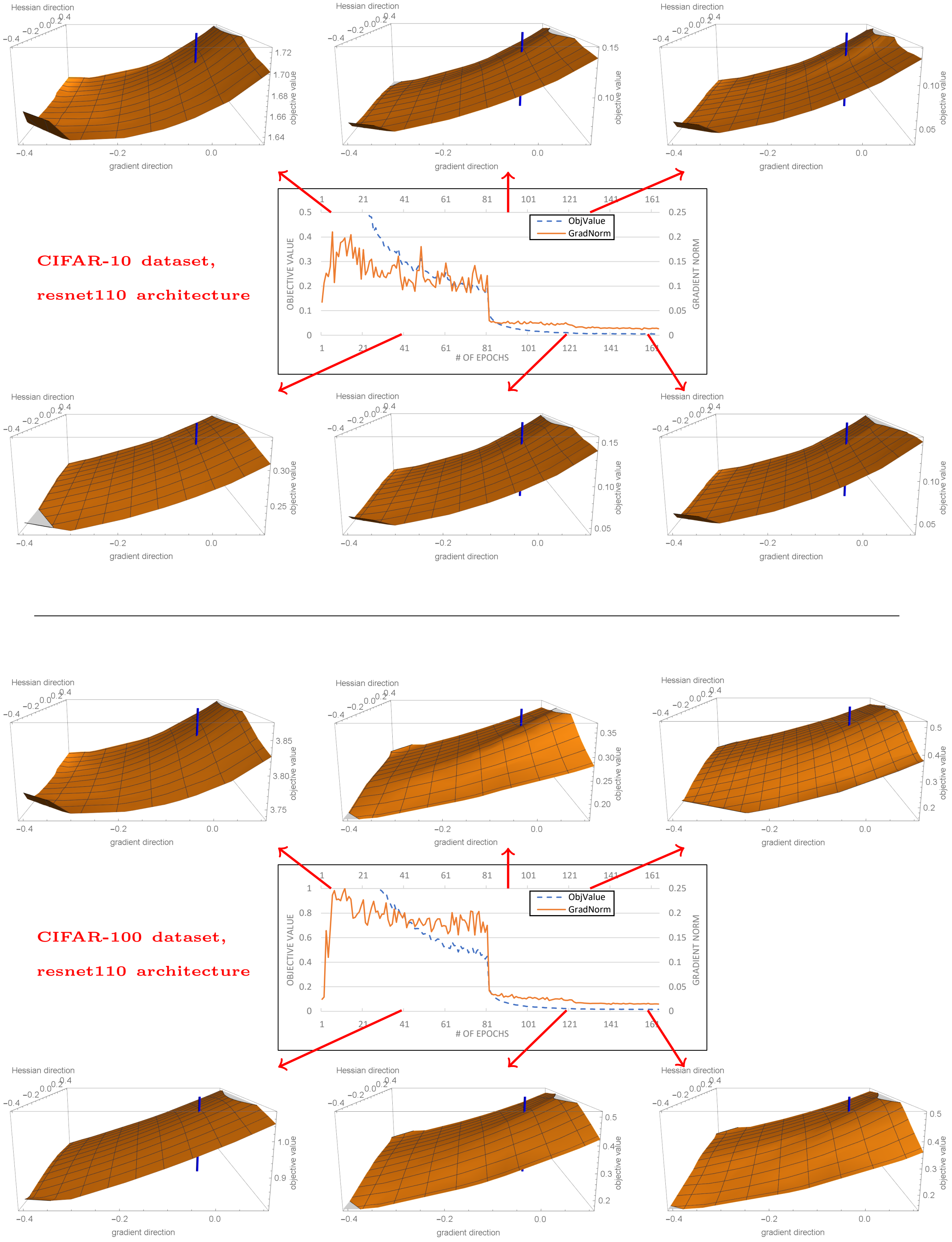

Figure 3: ResNet-110 architecture [58] landscape on CIFAR10 vs CIFAR100.

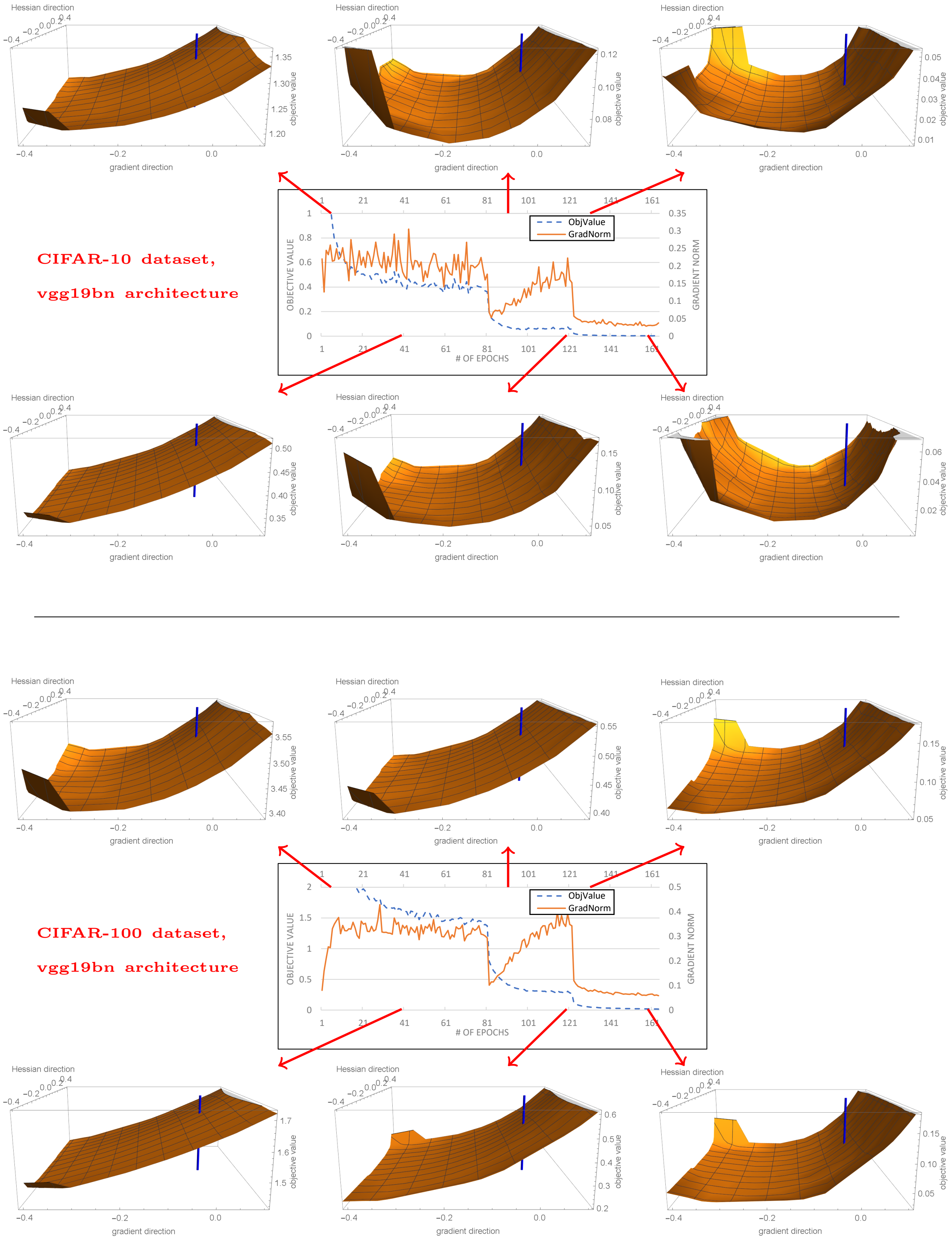

Figure 4: VGG19 architecture (with BN) [58] landscape on CIFAR10 vs CIFAR100.

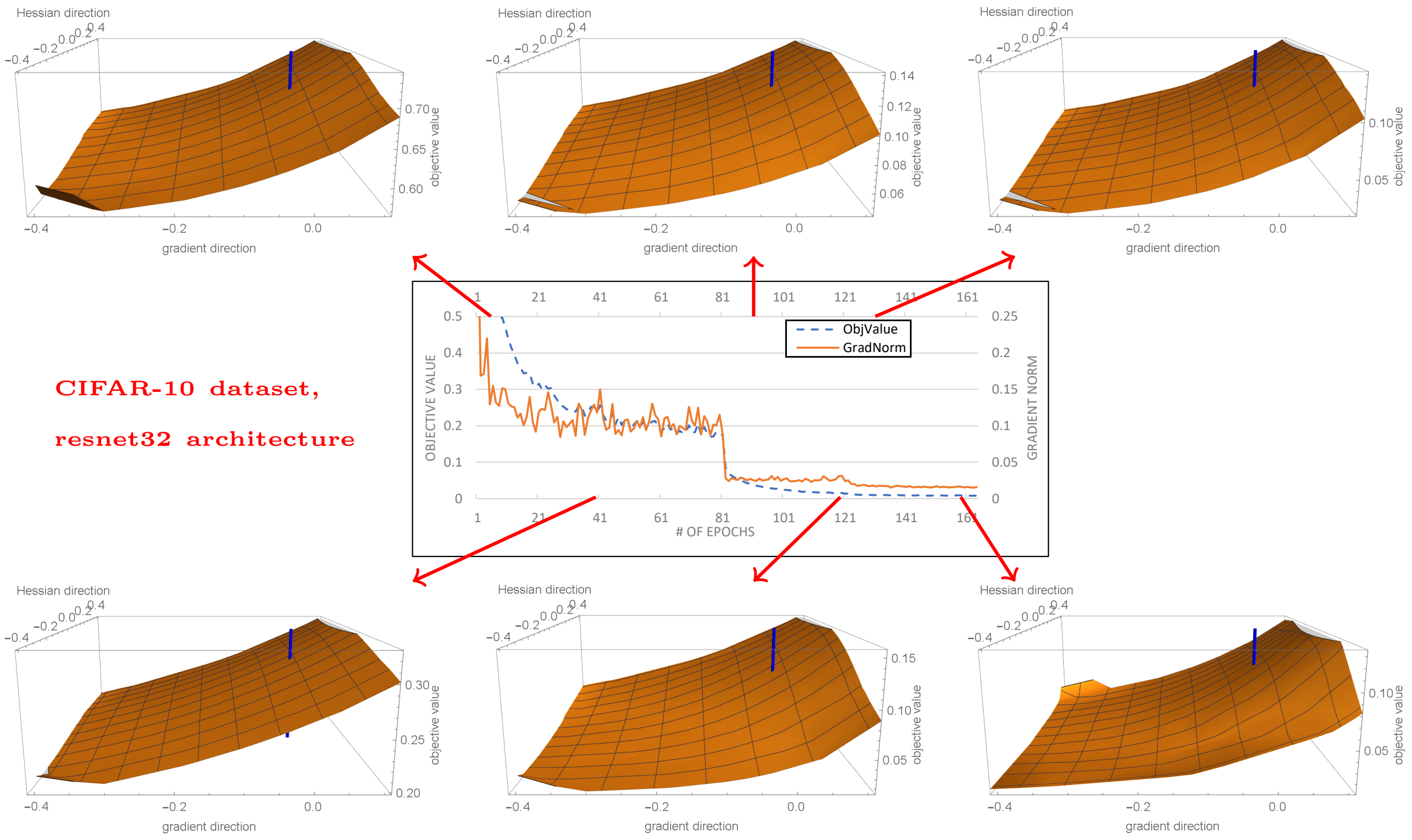

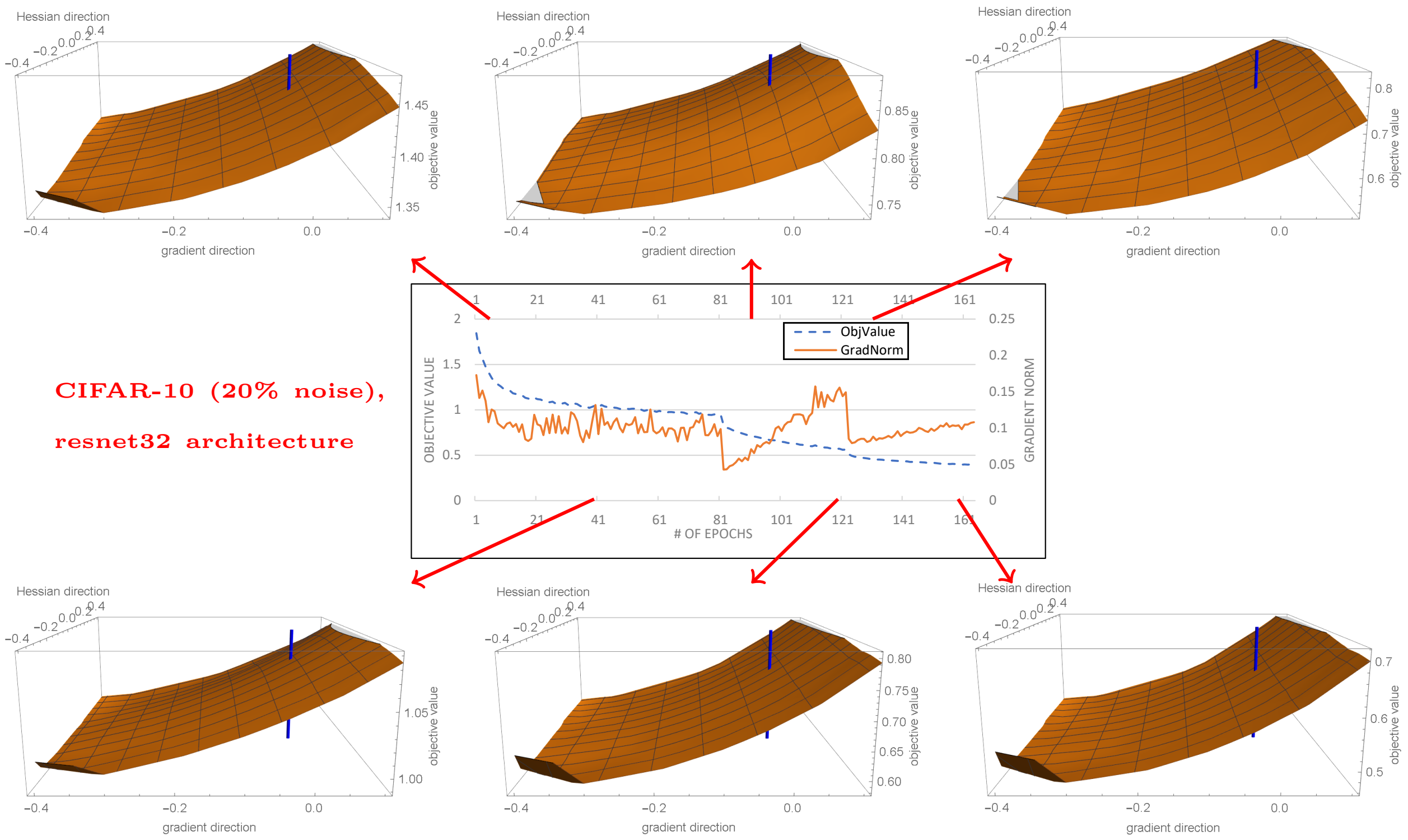

Figure 5: ResNet-32 architecture [58] landscape on CIFAR10 vs CIFAR10 (20% noise), means we have randomly perturbed 20% of the true labels in the training set.

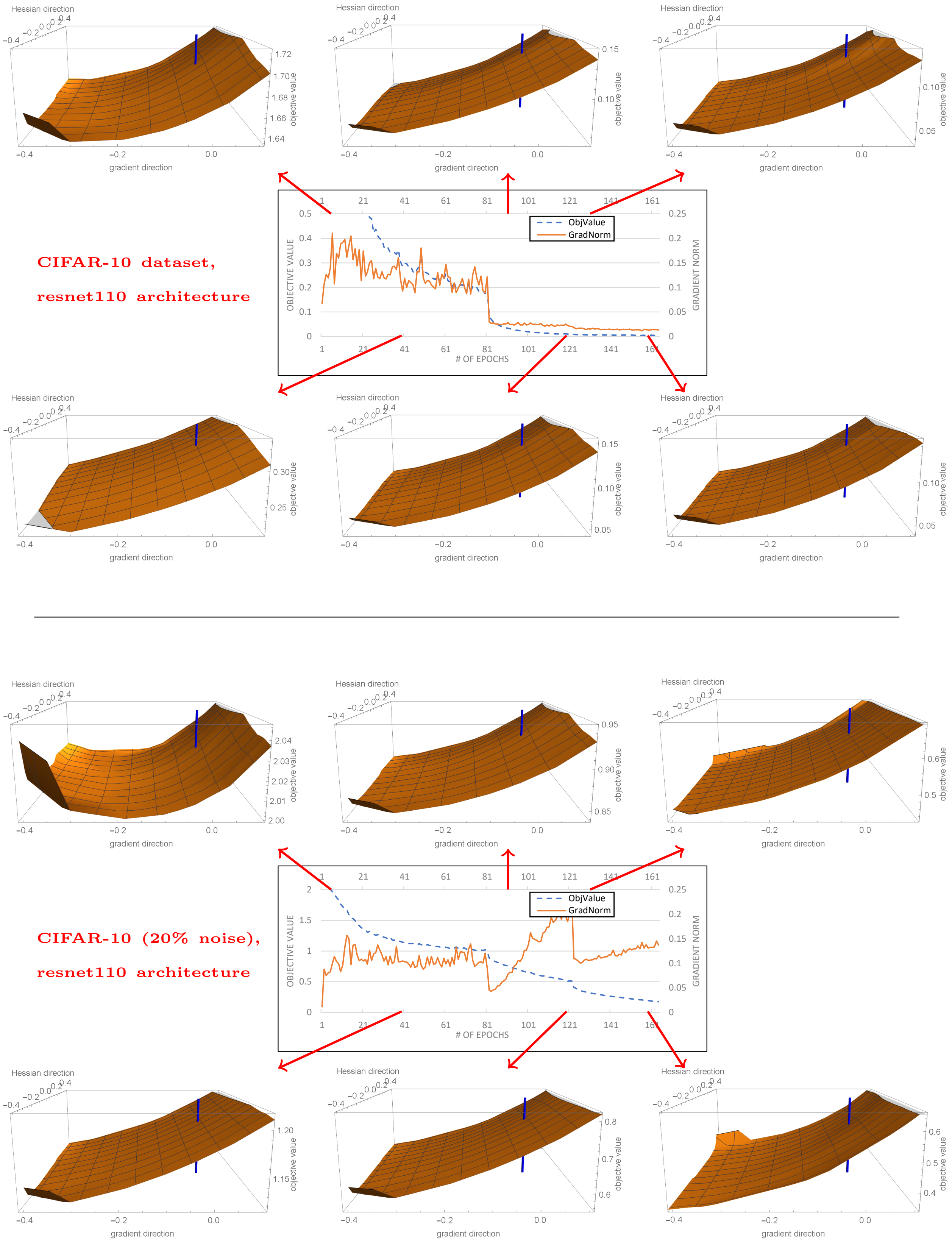

Figure 6: ResNet-110 architecture [58] landscape on CIFAR10 vs CIFAR10 (20% noise), means we have randomly perturbed 20% of the true labels in the training set.

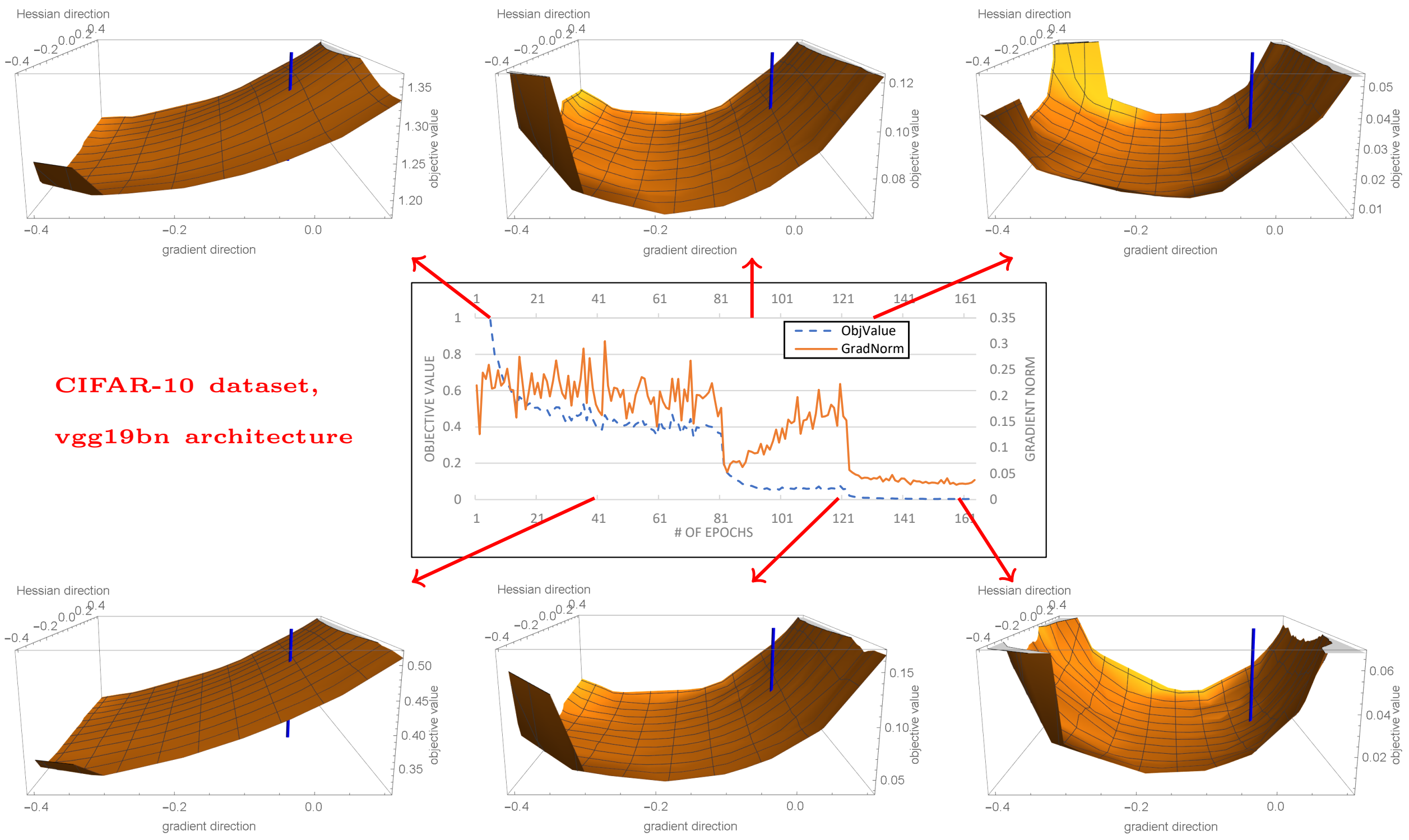

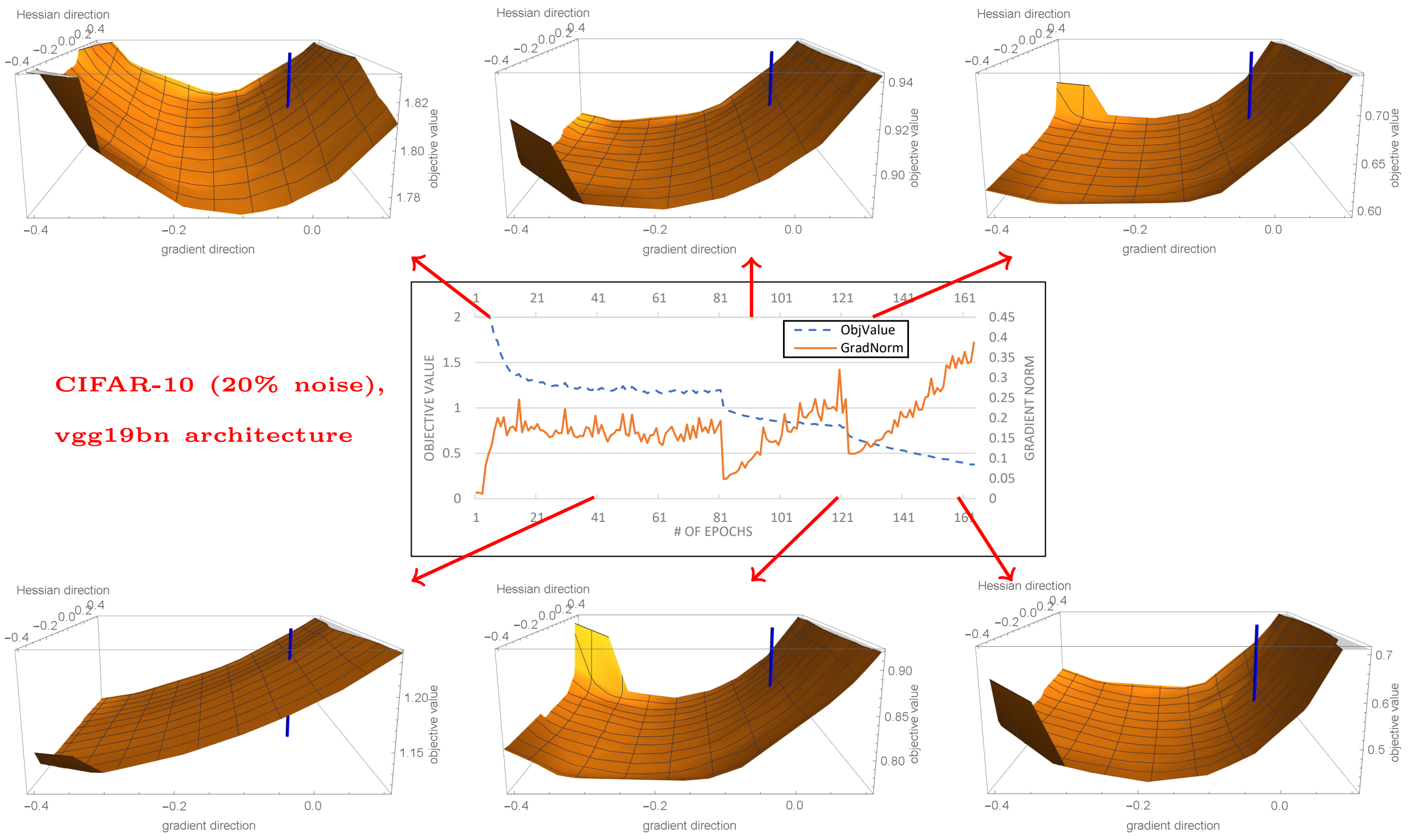

Figure 7: VGG19 architecture (with BN) [58] landscape on CIFAR10 vs CIFAR10 (20% noise), means we have randomly perturbed 20% of the true labels in the training set.